\documentclass[runningheads]{llncs}

\usepackage[mobile]{eccv}

\usepackage{eccvabbrv}

\usepackage{graphicx}
\usepackage{booktabs}
\usepackage{multirow}
\usepackage{float}
\usepackage{makecell}

\usepackage[accsupp]{axessibility}  

\usepackage{hyperref}

\usepackage{orcidlink}

\newcommand{\STAB}[1]{\begin{tabular}{@{}c@{}}#1\end{tabular}}

\begin{document}

\title{ControlNet-XS: Rethinking the Control of Text-to-Image Diffusion Models as Feedback-Control Systems}

\titlerunning{ControlNet-XS}

\author{Denis Zavadski\inst{} \and 
Johann-Friedrich Feiden\inst{} \and 
Carsten Rother\inst{}} 

\authorrunning{D.~Zavadski et al.}

\institute{Computer Vision and Learning Lab, IWR, Heidelberg University, Germany\\
\email{\{name.surname\}@iwr.uni-heidelberg.de}}

\maketitle

\begin{abstract}
The field of image synthesis has made tremendous strides forward in the last years. Besides defining the desired output image with text-prompts, an intuitive approach is to additionally use spatial guidance in form of an image, such as a depth map. In state-of-the-art approaches, this guidance is realized by a separate controlling model that controls a pre-trained image generation network, such as a latent diffusion model \cite{Rombach2022_LDM}. Understanding this process from a control system perspective shows that it forms a feedback-control system, where the control module receives a feedback signal from the generation process and sends a corrective signal back. When analysing existing systems, we observe that the feedback signals are timely sparse and have a small number of bits. As a consequence, there can be long delays between newly generated features and the respective corrective signals for these features. It is known that this delay is the most unwanted aspect of any control system. In this work, we take an existing controlling network (ControlNet~\cite{Zhang2023_ControlNet}) and change the communication between the controlling network and the generation process to be of high-frequency and with large-bandwidth. By doing so, we are able to considerably improve the quality of the generated images, as well as the fidelity of the control. Also, the controlling network needs noticeably fewer parameters and hence is about twice as fast during inference and training time. Another benefit of small-sized models is that they help to democratise our field and are likely easier to understand. We call our proposed network ControlNet-XS. When comparing with the state-of-the-art approaches, we outperform them for pixel-level guidance, such as depth, canny-edges, and semantic segmentation, and are on a par for loose keypoint-guidance of human poses. All code and pre-trained models will be made publicly available.
\end{abstract}

\vspace{-0.8cm}
\section{Introduction}
\label{sec:intro}

\begin{figure}[htp!]
    \centering
    \captionsetup{type=figure}
    \includegraphics[width=0.24\textwidth]{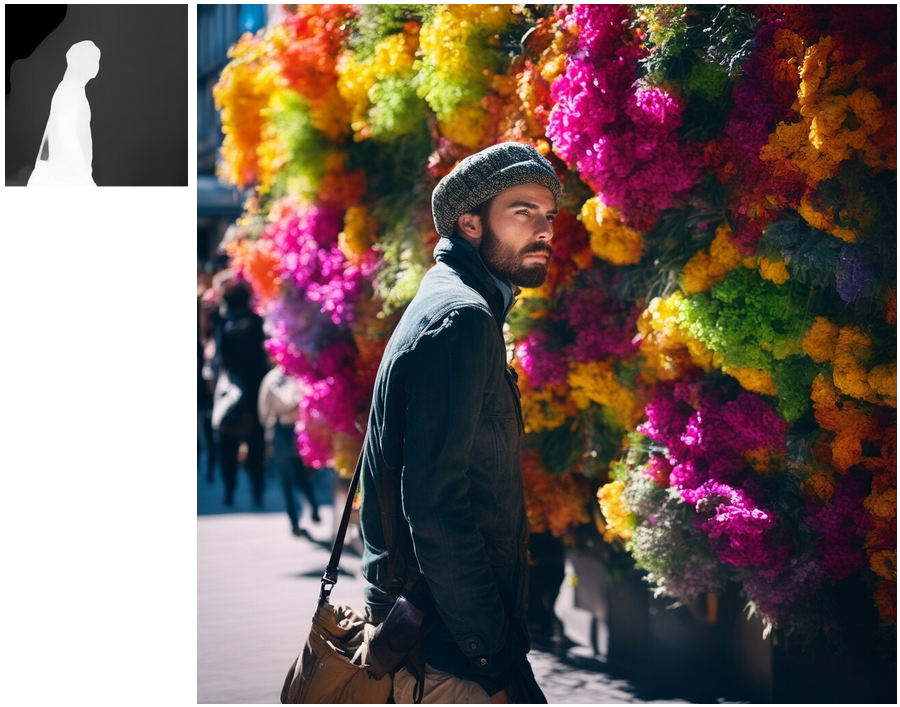}
    \includegraphics[width=0.24\textwidth]{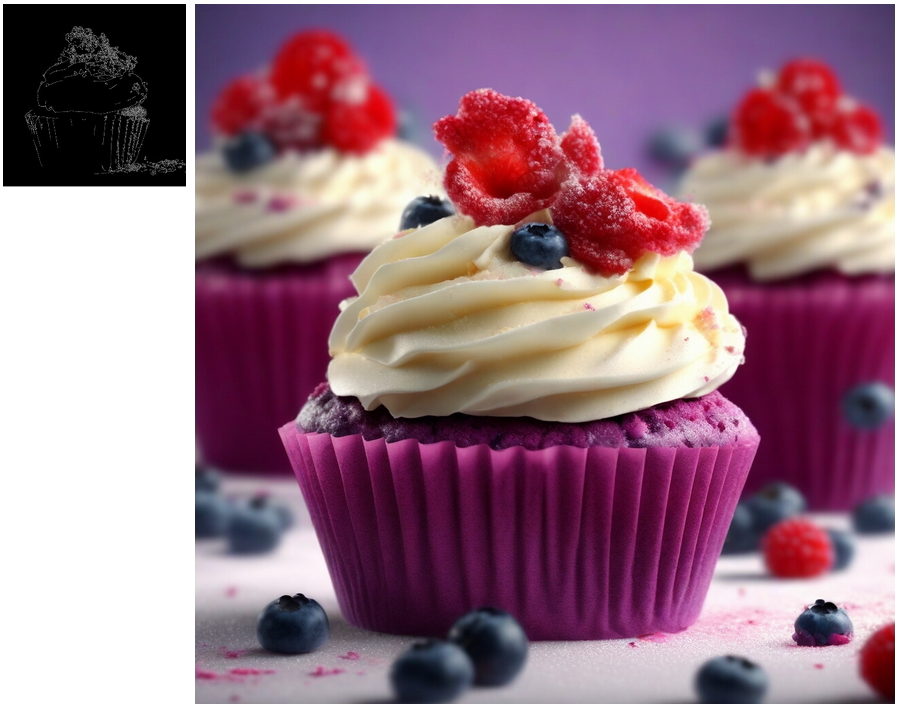}
    \includegraphics[width=0.24\textwidth]{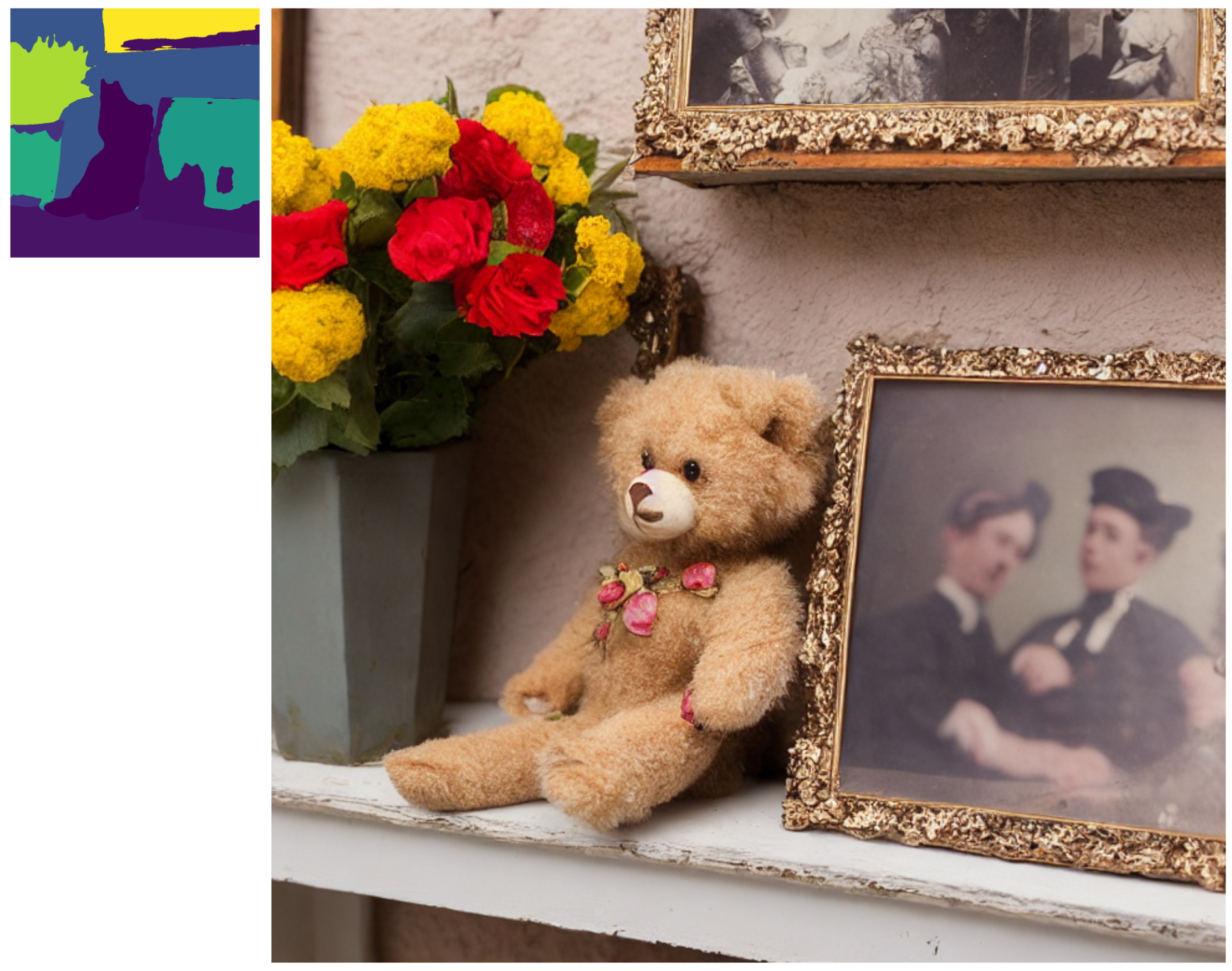}
    \includegraphics[width=0.24\textwidth]{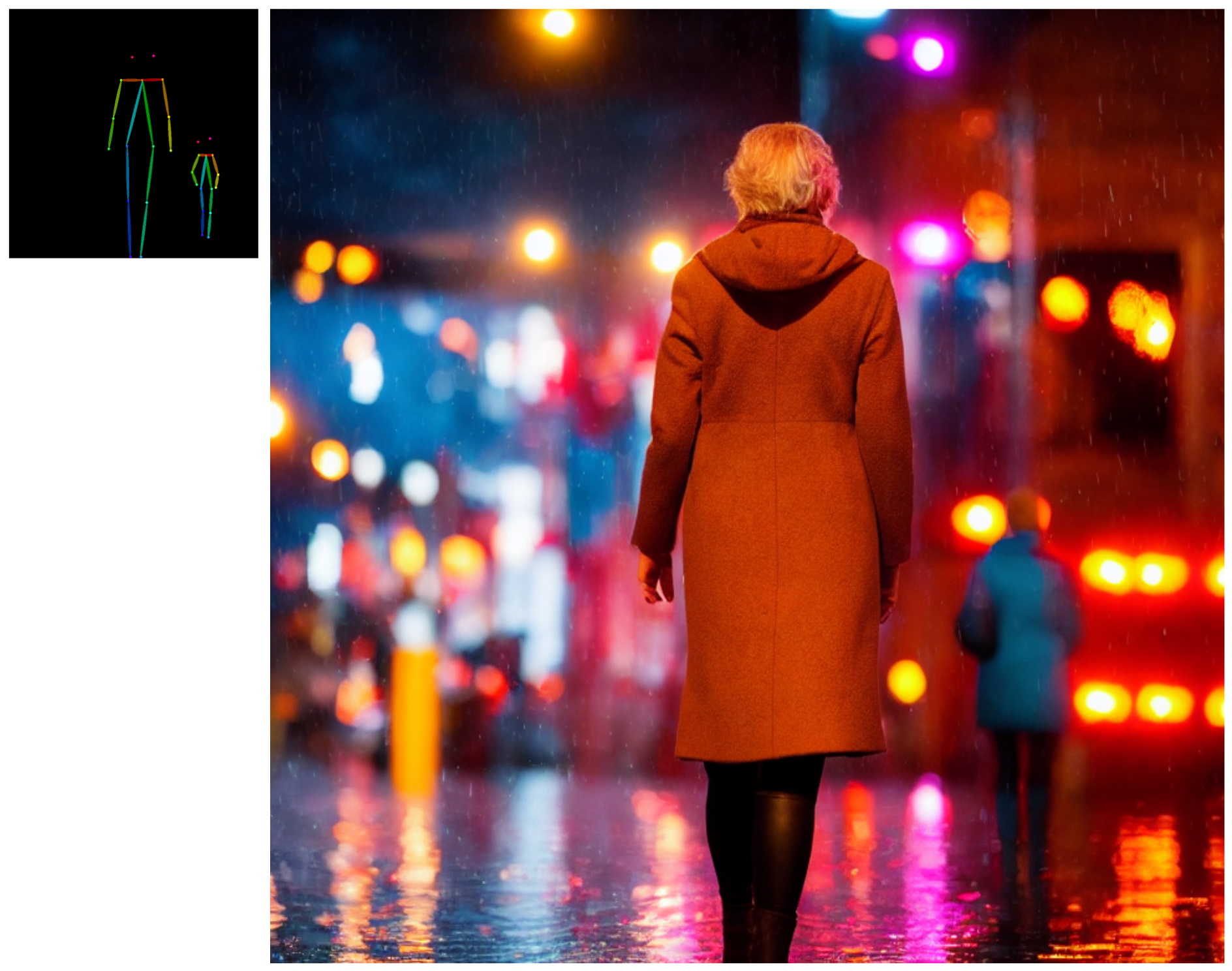}
    \vspace{-0.3cm} %
    \captionof{figure}{Image synthesis using our approach with text-prompts, as well as, a guidance image in form of a depth map, canny-edges image, semantic map, and human pose. The two results on the left-hand side were generated by the production-quality model of Stable Diffusion XL ~\cite{Podell2023_SDXL}, and the remaining by Stable Diffusion Version 1.5.
    \vspace{-0.7cm} %
    } 
\label{fig:teaser}
\end{figure}

Using Generative Artificial Intelligence to synthesize new images is a topic that has received large attention in social media, press, research and industry. It started off in $2014$ with the introduction of Generative Adversarial Networks (GAN) ~\cite{GAN} that were able to synthesize small-sized images of a given class~\cite{radford2016unsupervised}, \eg celebrity faces. Today, we have commercial and non-commercial products, such as Midjourney ~\cite{Midjourney2023} and Stable Diffusion XL ~\cite{Podell2023_SDXL}, which are able to generate large-sized images (up to $1024\times1024$) with almost arbitrary content, \eg ranging from professional photographs to Manga Art. This can be considered as a truly disruptive technology, for which the development is still far from being completed. One ongoing development is to create control tools, with which users can steer the image generation process towards their desired output. A common control mechanism is text-prompts. Another choice is to use a {\it guidance image}, which defines the desired output image in an abstract form such as a sketch or a depth map. 
This control mechanism is known as image-to-image translation, \eg ~\cite{ImageToImage}. 
The control mechanisms can also be combined by adding one or many guiding images to a text-to-image model, \eg ~\cite{Zhang2023_ControlNet}. 
Our work falls into this class of methods.   
There are in general two different choices for implementing the image guidance in a text-to-image model.

On one hand, there is the approach of {\it fine-tuning} a generative model with a new control mechanism at hand, \eg ~\cite{Wang2022}, or training it from scratch \eg ~\cite{Rombach2022_LDM}. Such methods use guidance images as additional input. 
However, such an end-to-end learning approach is challenging since oftentimes there is a large imbalance between the original training data for the generative process, \eg $\sim3$B images for training Stable Diffusion ~\cite{Rombach2022_LDM}, in contrast to only $1$M images with known control, as in~\cite{Zhang2023_ControlNet}. Such an imbalance can lead to effects like ``catastrophic forgetting''~\cite{McCloskey1989}, which means that known properties of the generative model disappear after fine-tuning. Additionally, fine-tuning often requires access to a large computing cluster.

On the other hand, there are approaches that lock the parameters of the generative network and then train a {\it separate controlling network}. 
In general, the idea of two networks communicating with each other has been shown to be beneficial for various computer vision tasks~\cite{tang2020xinggan, wang2019cfsnet, lu2022dual, liu2021cross}.
In our context, this particular design choice~\cite{Mou2023_T2I} is currently most popular and also utilised in this work. 

When analysing this approach from a control system perspective, we see that nearly all existing approaches are feedback-control systems, also known as closed-loop control systems. The controlling network is the {\it controller} and the generation processes is a dynamically changing system. The only exception is the T2I-Adapter approach \cite{Mou2023_T2I} since it forms an open-loop control system where the controlling network does not receive any feedback from the generating network. 
From a control system perspective, it is of paramount importance that the controller receives feedback from the generation process as often as possible (\ie high-frequency) and also as much useful information as possible (\ie large-bandwidth). Furthermore, the controller should send as often as possible a control signal back to the generation process. In a typical hardware control system, such as a manufacturing plant, there are physical limitations to achieve this. However, with software, we do not have such limitations. Unfortunately, to the best of our knowledge, all existing guided image generation methods which work as feedback-control systems \cite{Zhang2023_ControlNet,UniControl,UniControlNet, Cocktail} fall short in terms of designing a bidirectional, high-frequency and large-bandwidth communication between the controller and the generation process. The main contribution of our work is to construct a control system with these properties. For this, we use ControlNet~\cite{Zhang2023_ControlNet} as our initial controlling network. While improving the communication mechanism, we observe that we can scale down the number of parameters of ControlNet by a factor of 6.5 (or even more), and at the same time improve the quality of the generated images as well as the fidelity of the control. As a result, we are about two times faster than ControlNet with respect to inference and training time. We call our network {\it ControlNet-XS}.
Another benefit of small-sized models is that they help to democratise our field and are likely easier to understand. It is important to note that our main contribution, a bidirectional, high-frequency and large-bandwidth communication, can also be integrated into all other existing guided image generation approaches that are designed as a feedback-control system~\cite{Zhang2023_ControlNet, UniControl, UniControlNet, Cocktail}. Furthermore, also approaches in related fields of generative Artificial Intelligence which utilise a feedback-control system could benefit from our contribution, such as video translation~\cite{Feng2023_ccedit, Zhao2023_MakeAProtagonist} or controlled 3D object generation~\cite{Zhang2023_AvatarVerse, Huang2023_DreamControl}. 

Let us consider an example which illustrates the importance of having a high-frequency communication between the controller and the generation process. Assume that the generation process is a vehicle, which is equipped with a controller that is a satellite navigation system. The controller receives the current position of the vehicle, and gives back control signals such as “turn left at next junction”. Assume that the target or input to the system is to drive to a specific address. A crucial requirement of this control system is, obviously, that the navigator knows always the exact position of the vehicle.  However, if the navigator were to know the position of the vehicle with a delay of ten seconds, then the vehicle might have already passed the junction where it should have turned left.
Hence, the controller has to be smart and predict where the vehicle may be in ten seconds time in order to give commands that are sensible for the vehicle at that point of time in the future. This is exactly what happens in all existing approaches for guided image generation that are designed as a feedback-control system. Due to delayed feedback of the generation process to the controlling network, it has to guess what the generation network is doing until it sends its controlling signal. Hence, for this task, the controlling  network needs generative power.    

In summary, our contributions are as follows: (1) Demonstrating the importance of controlling a text-to-image generation model with a bidirectional, high-frequency and large-bandwidth communication. (2) Training a straight-forward, small-sized controlling network that  outperforms the state-of-the-art for pixel-level image guidance (depth, canny-edges, semantic map) and is on a par for loose keypoint guidance of human poses. (3) Controlling the production-quality Stable Diffusion XL model \cite{Podell2023_SDXL} with 2.6B parameters with a control network that has only $20$M parameters. 

\vspace{-.3cm}w
\section{Related Work}
\label{sec:related}
\vspace{-.1cm}
\subsection{Image Generation and Translation}
Generative Adversarial Networks (GANs)~\cite{GAN} are probably the most established generative models for unconditional~\cite{Karras2019_StyleGAN, StyleGAN2, StyleGAN3}, class-conditional \cite{Sauer2022_StyleGAN-XL} and text-conditional \cite{Sauer2023_StyleGAN-t, Kang2023_GigaGAN, Reed2016a, Reed2016, Tao2022, Xu2018, Zhang2017, Zhu2019} image generation as well as domain adaptation~\cite{Zhu2017, Murez2018, Xia2023}. While achieving state-of-the-art results for particular semantic classes \cite{CelebA-HQ, Wah2011_caltecBirds, Russakovsky2015_ImageNet}, generalizing GANs to synthesize images of arbitrary content remains an active area of research.  With GigaGAN \cite{Kang2023_GigaGAN} demonstrating state-of-the-art performance, generic text-to-image synthesis is still dominated by autoregressive networks 
like DALL-E~\cite{Ramesh2021_dalle1}, Parti \cite{Yu2022}, CogView \cite{Ding2021_CogView}, Make-A-Scene \cite{Gafni2022_MakeAScene} and Diffusion Models in particular.
Since their introduction, \textbf{Image Diffusion Models}~\cite{DM_Origin} rapidly became one of the best performing model-families \cite{Dhariwal2021, Nichol2021b, Ho2022, Ho2020, Kingma2021, Song2020}. Diffusion models learn to transform a point from a simple, high-dimensional distribution, such as a Gaussian, to a complex distribution, like the space of all images. This transformation is done by iteratively applying a network that gradually removes the Gaussian noise in the image.  This process allows to theoretically model arbitrary complex data distributions~\cite{DM_Origin}. 

\textbf{Conditioning Image Synthesis Models.}
To generate a desired output image, one popular choice is to use text-prompts as guidance. This is done by conditioning a generic image synthesis model on a textual-embedding, provided by pre-trained text-encoders like BERT \cite{Devlin2019_BERT}, T5 \cite{Raffel2020} or CLIP \cite{Radford2021_CLIP}. Such conditioning has led to impressive results for complex text-to-image generation task by models like Stable Diffusion~\cite{Rombach2022_LDM}, DALL-E~\cite{Ramesh2022_dalle2, Betker2023_dalle3}, Imagen~\cite{Saharia2022_imagen} and many others \cite{Podell2023_SDXL, Nichol2021_GLIDE, Xu2023}.
However, text-prompts alone provide very little control over the exact details within the generated image. To address this problem, the concept of {\it guidance images} became popular. Guidance images describe the desired output scene in an abstract form. This ranges from very loose guidance, such as bounding boxes and keypoints for human poses, on to slightly more precise controls like semantic maps or sketches, and up to pixel-accurate guidance, such as depth maps, normal maps, or edge maps.

As mentioned in the introduction, one line of work is to utilize guidance images as conditioning before training the model ~\cite{Huang2023_Composer, Li2024_UNIMO-G, TamingTransformers, Saharia2022b_palette, Xu2023}. Another variant is to adapt a pre-trained model by fine-tuning it with new guidance images ~\cite{Hu2024_InstructImagen, Wang2022}. However, the drawbacks of both approaches are that they require substantial computational resources to train, and also that conditional modalities cannot be changed without excessive re-training. The more popular approach is to train a separate controlling network which is combined with a pre-trained generation model, as discussed in the next section. 

\textbf{Image Editing and Subject-Driven Image Generation.}
There have been many works leveraging the rich internal representation of pre-trained text-to-image models to customise the output. This usually involves the editing of an existing image ~\cite{Couairon2022_DiffEdit, Choi2023_Custom-Edit, Li2023_DreamEdit, Bar-Tal2022_Text2Live, Goel2023_PAIR-Diffusion} or the insertion of a specific subject instance to the generated output~\cite{Ruiz2023_dreambooth, Gal2022, Yang2023, Patel2024_lambda-Eclipse}. The approach of DiffEdit~\cite{Couairon2022_DiffEdit} takes an existing image and utilises local masks to manipulate the content. 
InstructPix2Pix~\cite{Brooks2023_InstructPix2Pix} trains an image generation model to edit images according to human instructions. This is done by learned image-editing instructions. 
Dreambooth~\cite{Ruiz2023_dreambooth}, on the other hand, fine-tunes a pre-trained text-to-image generation model to a set of photographs representing a specific subject, binding it to a text-token which can be used to generate the desired subject in a new environment. 
In general, such approaches can be considered as complementary to works that use pixel-accurate image guidance, such as ours, since these approaches do not control the generated, or edited, content on a pixel-level.

\subsection{Controlling Pre-Trained Networks}
With the increasing size of generative models, it has become popular to leave the generative base model unaltered in order to keep its generative capabilities. 
A straight-forward approach is to employ weight-adaptation methods like ``Low-Rank-Adaptation’’ (LoRA) \cite{Hu2021_LoRA} to add a learnable offset to the pre-trained weights, which are approximated by the multiplication of two low-rank matrices. After training, the weights can be merged without the need to add new parameters to the network.
Beyond this simple approach, there are in general three concepts for controlling image generation models by the addition of new network components, although these concepts cannot always be clearly distinguished. 

\textbf{Adapters.}
One control mechanism is to use so-called adapters, which insert new trainable modules, \eg neural blocks, to the pre-trained generative network. However, in contrast to LoRA the inference time increases. Adapters are popular in natural language processing \cite{Stickland2019, Pfeiffer2021_AdapterFusion, Mao12022_UNIPELT} and have been transferred to image generation models \cite{Rosenfeld2020, Rebuffi2018}, vision transformers (ViT) \cite{Li2022}, and are also used for dense predictions in the form of ViT-Adapters \cite{chen2022}. In the context of our work, there is one adapter-based approach, T2I-Adapter \cite{Mou2023_T2I}, which uses a guidance image to control a pre-trained text-to-image Stable Diffusion model. It is an open-loop control system where features, derived from the guiding image, are added to the generation model. There is no feedback signal from the generation process back to the controlling network. We validate experimentally that it is on average inferior to feedback-control systems.      

\textbf{Image Control with Attention Maps.}
Another popular control mechanism is to manipulate the attention maps of the diffusion model
~\cite{Li2023_GLIGEN, Lukovnikov2024_Layout2Image, Chen2024, Zhao2023_lcc, He2023_locGuidance}. 
One example of such an approach is GLIGEN~\cite{Li2023_GLIGEN}, to which we compare experimentally. GLIGEN introduces new learnable gated attention layers to incorporate the guidance. While this gives impressive results for loose guidance, such as bounding boxes, we validate experimentally that for pixel-accurate guidance, like depth maps, it performs rather poorly compared to methods discussed next.

\textbf{Image Control with a Separate Controlling Network.}
The final approach for controlling a pre-trained image generation model is to train a separate controlling network, which generates features that are combined with the generation model. This is also our approach. The first work in the context of diffusion models was ControlNet~\cite{Zhang2023_ControlNet} which trains a separate controlling network for each kind of guidance image. Building on the design of ControlNet, three recent works appeared that allow to use multiple guidance images jointly within one controlling model. These are UniControl~\cite{UniControl}, Uni-ControlNet~\cite{UniControlNet} and Cocktail~\cite{Cocktail}. The main focus of these works is to design  neural networks that merge multiple control signals. For instance, UniControl uses a Mixture-of-Experts (MOE) Adapter in combination with a task-aware network. By doing so, each of these works add additional ``concepts'' on top of the initial ControlNet model: i) an additional text-embedding in the control signal \cite{UniControl}, ii) a global control adapter~\cite{UniControlNet}, iii) additional manipulation of attention maps of the encoder of the generation model~\cite{Cocktail}. 
In contrast to these works, we focus on single image guidance. We also use ControlNet as initial model, but then only reduce its size without adding any other ``concept''. Our key contribution is a new communication mechanism between the controlling and generating networks, which is different to all four works \cite{Zhang2023_ControlNet,UniControl,UniControlNet,Cocktail}. By doing so, we are able to improve over the state-of-the-art. It is important to note that our improved communication mechanism can also be integrated into Uni-ControlNet, UniControl and Cocktail.

\vspace{-0.1cm}
\section{Method}
\label{sec:method}
We start with a brief introduction to the Stable Diffusion~\cite{Rombach2022_LDM} architecture, which serves as our generative model (\Cref{sec:preliminary}). The pre-trained generative model is controlled by a controlling network. In \Cref{subsec:controlsystem} we analyse the design of existing controlling network architectures from a feedback-control system perspective. In this work, we build upon ControlNet~\cite{Zhang2023_ControlNet} as a controlling network which we describe in \Cref{subsec:controlNet}.  Lastly, in \Cref{subsec:controlNet-xs}, we introduce our ControlNet-XS network and its training procedure in \Cref{subsec:controlNet-xs-train}.

\begin{figure}[htp!]
    \centering
    \captionsetup{type=figure}
    \includegraphics[width=0.95\textwidth]{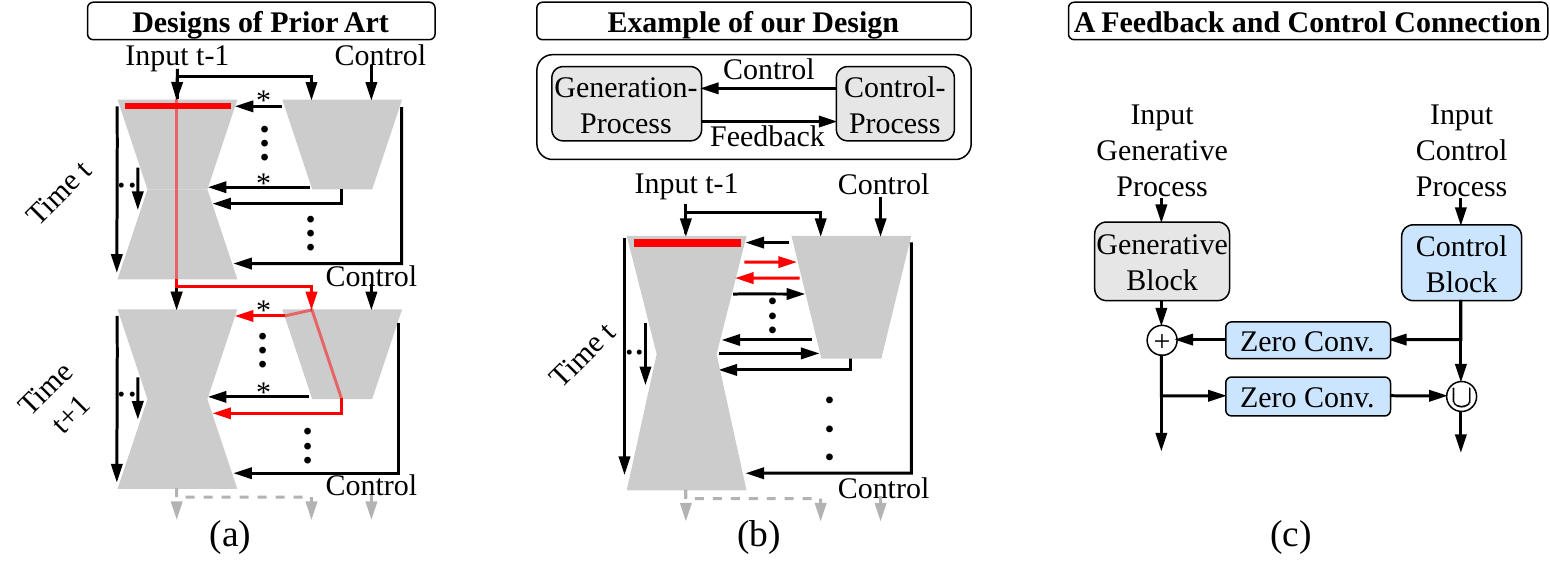}
    \label{fig:feedbackcontrol}
    \vspace{-0.4cm}
    \caption{{\bf Feedback-Control System Perspective}. In each figure (a-c) the generation process is on the left-hand side, and the control process on the right-hand side. The focus of this illustration is on the communication (directed arrows) between the generation and controlling process. (a) Feedback-control system for approaches \cite{UniControlNet,UniControl,Zhang2023_ControlNet,Cocktail}, where links denoted by * are only present in \cite{Cocktail}. (b) An example of our communication design. (c) Zoom into the connections between a generative encoder block and a ControlNet-XS block. Please find the explanation for this figure in \Cref{subsec:controlsystem} 
    \vspace{-0.3cm}
    }
    \label{fig:FeedbackControl}
\end{figure}

\vspace{-0.2cm}
\subsection{Stable Diffusion}
\label{sec:preliminary}
Stable Diffusion~\cite{Rombach2022_LDM} is a U-Net based diffusion model for text-to-image generation. As a conditional diffusion model, it receives a text embedding from a separate text encoder, as well as a learned time embedding. The output image is reconstructed from noise by iteratively running the U-Net over, for example, $50$ time-steps. The U-Net generator is composed of a sequence of neural blocks involving cross-attention mechanisms for text conditioning. The image signal is processed by the encoder in four layers with diminishing resolution and three neural blocks per layer. Through the mirrored structure of the decoder and one middle block in between, the U-Net has a total of 25 neural blocks. The output of each neural block can be influenced individually by a controlling network.

\vspace{-0.2cm}
\subsection{Feedback-Control System Perspective}
\label{subsec:controlsystem} 
In \Cref{fig:FeedbackControl}a-b we analyse two different design choices for controlling a generation process.  Feedback-control systems for approaches \cite{UniControlNet,UniControl,Zhang2023_ControlNet, Cocktail} are shown in (a), where links denoted by * are only present in \cite{Cocktail}.\footnote{The * links adapt the attention maps of the generative encoder.} Note that we only illustrate the generation and controlling networks and their respective communication links. The approaches \cite{UniControlNet,UniControl,Cocktail} have additional networks and communication links (see \Cref{sec:related}) which are omitted here, since it is not relevant to our analysis. 
The main drawback of design (a) is that there can be generated features, illustrated by the red rectangle in the encoder, at time $t$ of the generation process, that evolve in the generative U-Net and receive a control signal only at time $t+1$ (two red arrows pointing towards U-Net). However, by that time, they have travelled through $35$ generative blocks (or $25$ blocks if * links are present). The two red paths show possible flows through the network which start at the generated features (red rectangle) and pass through the controlling network until they form the control signals for the generative process at time $t+1$ (red arrow with * or without *). There is no earlier control signal that is aware of the generated features (red rectangle).
In our design (b), this drawback is eliminated and such features (red rectangle) receive a control signal after one generative block. 

Besides implementing a high-frequency communication, the bandwidth of the connection may also play a crucial role. When measuring the bits that travel through the networks, we notice that rather few bits go from one time-step to the next time-step of the generation process (precisely $524$K bits for a latent image of size 64×64×4). This is the only feedback to the controlling network in design (a). In contrast, the total number of bits entering the controlling network in (b) is 212M bits, which is over 404 times more.\footnote{The size of the features is measured where they leave the generative model.} The conclusion of this analysis is that in (a) the controlling model may face an even more challenging task, compared to design (b), for computing the appropriate control signal since it receives far less input from the generative process.

\begin{figure*}[t!]
  \centering
  \begin{subfigure}{0.2\linewidth}
	\includegraphics[width=\linewidth]{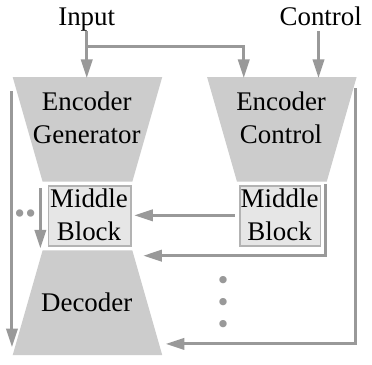}
    \caption{ControlNet~\cite{Zhang2023_ControlNet}}
    \label{fig:arch_cn}
  \end{subfigure}
  \hspace{1em}
  \begin{subfigure}{0.2\linewidth}
	\includegraphics[width=\linewidth]{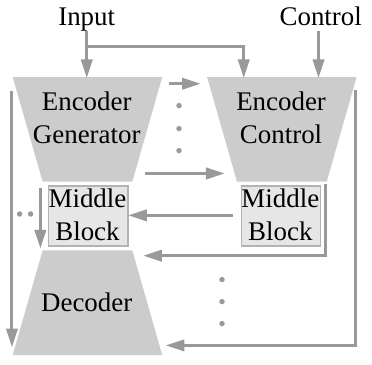}
    \caption{Our Type A}
    \label{fig:arch_a}
  \end{subfigure}
  \hspace{1em}
  \begin{subfigure}{0.2\textwidth}
	\includegraphics[width=\textwidth]{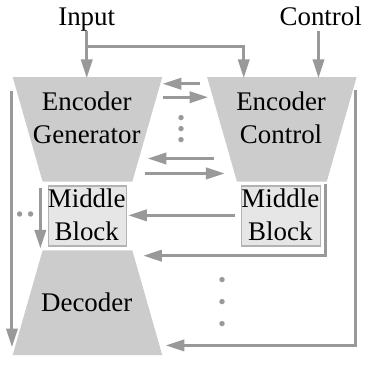}
    \caption{Our Type B}
    \label{fig:arch_b}
  \end{subfigure}
  \hspace{1em}
  \begin{subfigure}{0.2\textwidth}
	\includegraphics[width=\textwidth]{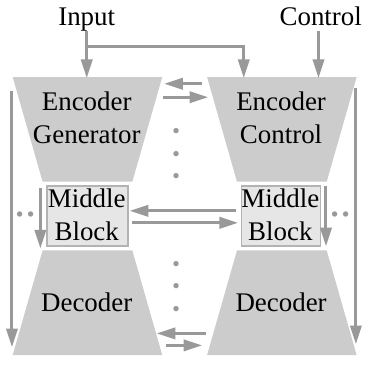}
    \caption{Our Type C}
    \label{fig:arch_c}
  \end{subfigure}
  \hfill
  \vspace{-0.3cm} %
  \caption{{\bf Architectural choices.} Different design-sketches for controlling a U-Net based generation process with a controlling network. The generation process is in each example on the left-hand side and the control process on the right-hand side. (a) The architecture of ControlNet~\cite{Zhang2023_ControlNet}. (b-c) Three new architectures (Type A-C) proposed in this work. 
  We verify experimentally that model Type B performs better than Type A, and is on a par with Type C. We choose Type B as our final architecture, and call it ControlNet-XS, since it has fewer parameters than Type C.
  \vspace{-0.5cm}
  }
  \label{fig:architectures}
\end{figure*}

\vspace{-0.4cm}
\subsection{ControlNet}
\label{subsec:controlNet}
The ControlNet~\cite{Zhang2023_ControlNet} architecture is sketched in \cref{fig:arch_cn}. It 
starts with a pre-trained generative model, here the U-Net of Stable Diffusion~\cite{Rombach2022_LDM}. The control model copies the encoder of the U-Net and hence has a representation that is capable of generating images by itself. The control encoder receives the control signal, \eg in form of a depth map, as well as the intermediate, noisy generated image. It outputs control signals that are fed into the different decoder blocks of the generative process. The connections from the control model to the generation model are initialized by so-called zero-convolutions, which have the effect that the generative capabilities of the controlled U-Net are not diminished at the beginning of training. During training, the encoder can learn to provide useful control signals to the generative process. The training objective is the same as for Stable Diffusion, \ie image denoising (see \Cref{subsec:controlNet-xs-train}).
While these may seem like reasonable design choices at first glance, they are sub-optimal from the perspective of a controlling system, as illustrated in \Cref{fig:FeedbackControl}a and discussed in \Cref{subsec:controlsystem}. In brief, the control model has two jobs at once: i) It has to process the feedback signal in order to make it useful for the generation process; ii) It has to anticipate what the generation process is going to do until the control signal is received by the generation model. We remedy the second job, and hence the control model can focus on the first one.

\vspace{-0.3cm}
\subsection{ControlNet-XS}
\label{subsec:controlNet-xs}
In \Cref{subsec:controlsystem} we have motivated our communication mechanism between controlling and generating network. The key idea is that the two encoders have an interaction with high-frequency. 
Based on this, we design three variants \cref{fig:arch_a}-\cref{fig:arch_c}. They vary in terms of connectivity between the two encoders and the two decoders, respectively. From a feedback-control system perspective, designs (c) and (d) are good, since they have a high-frequency communication between the two encoder networks. In design (b) there are also many so-called control-loops, however, for each loop the generative network does still uncontrolled processing within the loop. We validate experimentally that
model Type B is superior to A and on a par with Type C. Hence, we choose Type B as our final architecture since it has fewer parameters than Type C. We call Type B architecture {\it ControlNet-XS}. A detailed illustration of the architecture of Type B is in the supplement. The key building block for connecting the generation network and ControlNet-XS is shown in \Cref{fig:FeedbackControl}c. The calculated features are processed by zero-convolutions and added to the calculated features of the counterpart. Features coming from the generative block can be either added or concatenated to the features of the control block. Because we train the controlling network from scratch we utilise concatenation.  
Our new design allows to drastically reduce the size of the controlling network, by consistently changing the number of
channels in each control layer. We validate experimentally that even a model with as little as $1.7$M parameters performs on a par with ControlNet~\cite{Zhang2023_ControlNet} with $361$M parameters. Note that in ControlNet, a version of ControlNet with fewer parameters, called ControlNet-light, was evaluated but found to perform inferior. 

\vspace{-0.3cm}
\subsection{Training}
\label{subsec:controlNet-xs-train}
\vspace{-0.1cm}
As in related works, all weights of the generative model are frozen during training and we only learn the weights of the controlling network. Due to our improved design (Type A-C), we do not need the generative power of the controlling network, and hence all parameters are initialized randomly. We observe that the zero convolutions (see \Cref{fig:FeedbackControl}c) help to stabilize training. We train a separate controlling network for each kind of guidance. As training data, we use one million images from the Laion-Aesthetics dataset~\cite{Schuhmann2022_LaionAE}. For getting guidance images, we follow ControlNet~\cite{Zhang2023_ControlNet} and extract canny-edges, use ground truth segmentation maps, predict the depth maps with  MiDaS~\cite{Ranftl2020_Midas}, or predict human keypoints with OpenPose~\cite{Cao2017_OpenPose}. The standard diffusion model objective remains unchanged:
\vspace{-0.4cm}
\begin{alignat}{3}
	\mathcal{L} = \mathbb{E}_{z_0, t, c_t, c_f, \epsilon \sim \mathcal{N}(0, 1)}\bigl[ \Vert \epsilon - \epsilon_\theta (z_t, t, c_t, c_c)\Vert_2^2 \bigr],
\end{alignat}
with the target image $z_0$, the noisy image $z_t$, the timestep $t$, the text conditioning $c_t$ and the control conditioning $c_c$.

\vspace{-0.2cm}
\section{Experiments}
\label{sec:exp}
We start by defining the evaluation metrics (\Cref{subsec:metrics}), and then analyse variations of our new model in terms of architecture and size (\Cref{subsec:arch_eval}). \Cref{subsec:quantitative} is an in-depth, quantitative comparison to state-of-the-art approaches. The next \Cref{subsec:bias} discusses a semantic bias induced by large control models. Finally, to demonstrate the versatility of our approach, we apply it to a larger generative model, namely Stable Diffusion XL (\Cref{subsec:eval_sdxl}). If not stated differently, we use Stable Diffusion Version 1.5 as the generative model.      

\subsection{Evaluation Metrics}
\label{subsec:metrics}
We evaluate performance by estimating the fidelity of the control and the quality of the generated images, to ensure that the generated quality does not reduce with respect to uncontrolled image generation.
For quality evaluation, we use the CLIP-Score~\cite{Hessel2021_CLIPScore} which approximates the similarity between a given text-prompt and an image, and the CLIP-Aesthetics score~\cite{Schuhmann2022_LaionAE} which approximates the aesthetic appearance of an image as perceived by humans. The fidelity of the control is evaluated implicitly with the Learned Perceptual Image Patch Similarity (LPIPS)~\cite{LPIPS} and explicitly by a distance measure between two images, which is MSE for depth control, denoted by MSE-depth, mIoU for semantic map control, and the Hausdorff distance (HDD) for canny-edges and human poses (\ie keypoints). Here, the first image is the reference control image, \eg the depth map, and the second image is the extracted control, \eg the depth map from the generated image. The extraction algorithm is the same as the one used to generate the training data, \ie MiDaS~\cite{Ranftl2020_Midas} for depth extraction. 
Note that for improved readability, the MSE-depth values are scaled by $10^3$ and the Hausdorff distance is scaled by $10^{-1}$.
We also compute the Fréchet Inception Distance (FID)~\cite{FID}. %
We evaluate the semantic map control with the COCO-Stuff~\cite{Caesar2018_COCO-STUFF} validation set of $5000$ images and all other controls with the COCO~\cite{Lin2014_COCO} validation dataset of $5000$ images. 

Note that for pixel-accurate guidance (\ie depth and edges) the FID score measures both quality and fidelity of control since the control signal comes from a target image of the respective COCO validation set and hence the features of the generated image are expected to be similar. %
However, more loose controls like semantic maps and human poses in particular do not contain precise positional information about the features and hence the FID score only measures quality. The same applies to the similarity metrics LPIPS which is less relevant for guidance with semantic maps and not applicable to human pose guidance.

\vspace{-0.3cm}
\subsection{Ablation Study: Architecture}
\label{subsec:arch_eval}
We conduct an ablation study in \cref{tab:arch_type} for the four architectures shown in \cref{fig:architectures}. The size of our Type A-C are chosen to be about $20\%$ of ControlNet. For two metrics, quality (FID) and control (MSE-depth), all of our architectures are clearly superior to ControlNet. For other metrics, our design is marginally superior or on a par, respectively. Furthermore, Type B performs better than Type A for all measures. This is not surprising, since Type A has no control-signals in the encoder part (see \cref{fig:arch_a}). The performance of Type B and C are on a par. However, Type C has the drawback of effectively doubling the model size. We explain this lack of quantitative improvement of Type C in a sensitivity analysis in the supplement material. We choose type B as our final architecture for ControlNet-XS and use it in all remaining experiments. 

\begin{table}
\centering
\caption{{\bf Ablation study for four different architectures} illustrated in \cref{fig:architectures}. We see that all of our designs (Type A to Type C) outperform ControlNet~\cite{Zhang2023_ControlNet} (CN), both in terms of quality (FID) and control (MSE-depth). We select Type B as our ControlNet-XS architecture for all remaining experiments, since it performs best, on average, and has fewer parameters than Type C.
\vspace{-0.4cm}
}
\scriptsize
\begin{tabular}{clccccc}
    \hline
     && Both & \multicolumn{2}{c}{Control} & \multicolumn{2}{c}{Quality} \\
     &\multicolumn{1}{c}{Method} & FID $\downarrow$ &  MSE-d $\downarrow$ & LPIPS $\downarrow$ & CLIP-Sc $\uparrow$ & CLIP-Ae $\uparrow$  \\
    \hline \hline
    \parbox[t]{.5cm}{\multirow{4}{*}{\STAB{\rotatebox[origin=c]{90}{Depth}}}}
    &CN (361M)       & \small \textcolor{black}{19.01} 
                    & \small \textcolor{black}{29.1} 
                    & \small \textcolor{black}{0.532}
                    &  \small \textcolor{black}{28.96} 
                    &  \small \textcolor{black}{6.08}\\

    &Our Type A (53M)  
                    & \small \textcolor{black}{17.11} 
                    & \small \textcolor{black}{20.9}
                    & \small \textcolor{black}{0.492}
                    &  \small \textcolor{black}{29.00} 
                    &  \small \textcolor{black}{6.02}\\
    
    &Our Type B (55M) 
                    & \small 16.36
                    & \small \textbf{19.6}
                    & \small \textbf{0.468}
                    &  \small \textbf{29.21} 
                    & \small 6.09\\
    
    &Our Type C (117M) 
                    & \small \textbf{16.24} 
                    & \small 20.2
                    & \small 0.476
                    & \small 29.14
                    & \small \textbf{6.10}\\
    \hline
\end{tabular}
\label{tab:arch_type}
\vspace{-0.2cm}
\end{table}

\begin{table}
\centering
\caption{{\bf Ablation study of ControlNet-XS (Type B) in terms of size}. On average, we see that the performance increases slightly from $491$M to $55$M and decreases afterwards for smaller model sizes, up to $1.7$M. Please see discussion in \Cref{subsec:arch_eval}.
\vspace{-0.4cm}
}
\scriptsize
\begin{tabular}{clccccc}
   \hline
    && Both & \multicolumn{2}{c}{Control} & \multicolumn{2}{c}{Quality} \\
   &\multicolumn{1}{c}{Method} & FID  $\downarrow$ & MSE-d $\downarrow$ & LPIPS $\downarrow$ 
    & CLIP-Sc $\uparrow$ & CLIP-Ae $\uparrow$ \\
   \hline \hline
   &Stable Diffusion & \small \textcolor{gray}{22.69}    
                    & \small \textcolor{gray}{(69.7)}   
                    & \small \textcolor{gray}{(0.618)}  
                    & \small \textcolor{gray}{28.40}    
                    & \small \textcolor{gray}{6.16}     
                    \\
   \hline
   \parbox[t]{.5cm}{\multirow{4}{*}{\STAB{\rotatebox[origin=c]{90}{Depth}}}}
   &CN-XS (491M) & \small 16.91                     
                & \small 21.4                      
                & \small \textcolor{black}{0.487}  
                & \small 29.09                     
                & \small \textcolor{black}{6.07}   
                \\
                
   &CN-XS (55M)  & \small \textbf{16.36}            
                & \small \textbf{19.6}             
                & \small \textbf{0.468}            
                & \small \textbf{29.21}            
                & \small \textcolor{black}{6.09}   
                \\
                
   &CN-XS (11.7M)& \small \textcolor{black}{17.90}  
                & \small \textcolor{black}{28.6}   
                & \small 0.525                     
                & \small \textcolor{black}{28.83}  
                &\small  6.10                      
                \\
                
   &CN-XS (1.7M)& \small \textcolor{black}{18.45}   
                & \small \textcolor{black}{29.9}   
                & \small \textcolor{black}{0.526}  
                & \small \textcolor{black}{28.73}  
                & \small \textbf{6.12 }            
                \\
   \hline
\end{tabular}
\vspace{-0.6cm}
\label{tab:arch_size}
\end{table}

\begin{figure*}
\vspace{-0.5cm}
    \centering
    \begin{subfigure}[t]{0.19\textwidth}
        \centering
        \includegraphics[width=.95\textwidth]{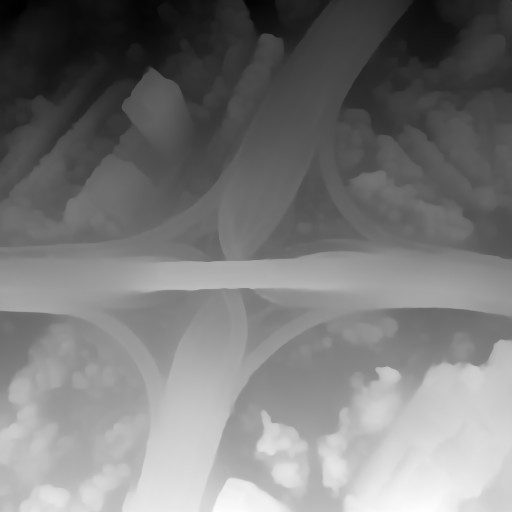} 
        \caption{Control}
    \end{subfigure}
    \hfill
    \begin{subfigure}[t]{0.19\textwidth}
        \centering
        \includegraphics[width=.95\textwidth]{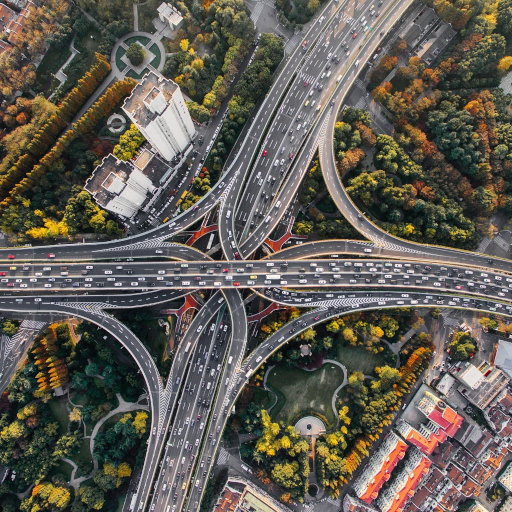}
        \caption{Original Image}
    \end{subfigure}
    \hfill
    \begin{subfigure}[t]{0.19\textwidth}
        \centering
        \includegraphics[width=.95\textwidth]{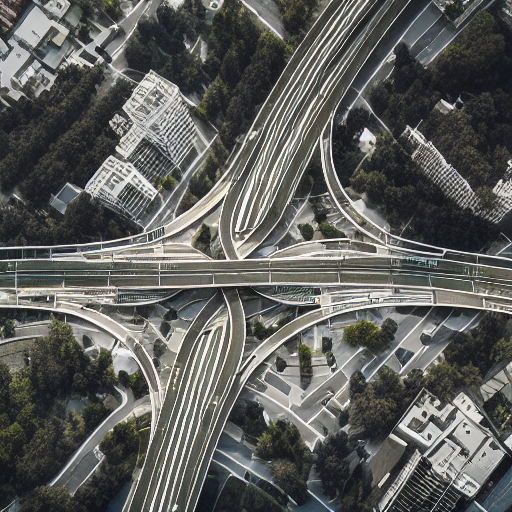}
        \caption{\scriptsize{CN-XS} ($55$M)}
    \end{subfigure}
    \hfill
    \begin{subfigure}[t]{0.19\textwidth}
        \centering
        \includegraphics[width=.95\textwidth]{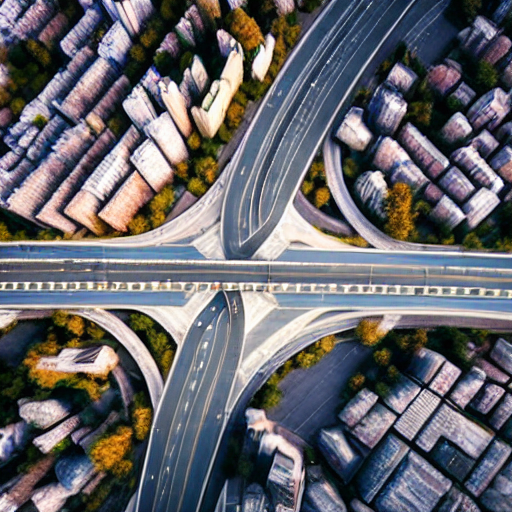}
        \caption{\!\!CN-XS\! (11.7M)}
    \end{subfigure}
    \hfill
    \begin{subfigure}[t]{0.19\textwidth}
        \centering
        \includegraphics[width=.95\textwidth]{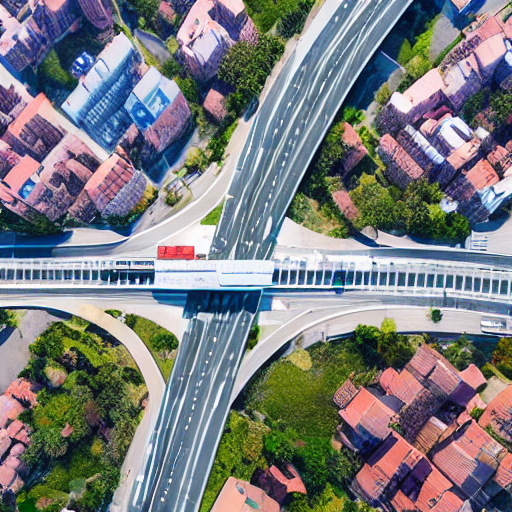}
        \caption{CN-XS (1.7M)}
    \end{subfigure}
    \hfill
    \vspace{-0.3cm}
    \caption{{\bf The fidelity of the control} reduces with smaller model sizes of ControlNet-XS. In the $55$M parameter model the complex structure of the street junction is identical to the one in the original image, as well as the skyscrapers in the upper-left corner. Smaller models with $11.7$M and $1.7$M parameters, respectively, are still guided by the control but less rigorously.
    \vspace{-0.9cm}
    }
    \label{fig:Control_sizes}
\end{figure*}

In the next experiment we evaluate whether changing the parameter size of ControlNet-XS influences the performance, see \Cref{tab:arch_size}. We examined ControlNet-XS (\ie Type B) with $491$M, $55$M, $11.7$M and $1.7$M parameters, respectively. We roughly see the same trend for all metrics that when varying the sizes of ControlNet-XS, the performance increases slightly from $491$M to $55$M and decreases afterwards for smaller model sizes, up to $1.7$M. Hence, we choose the $55$M model as our best model and show qualitative results in \cref{fig:teaser}. In terms of control, it means that smaller models have reduced fidelity of the control. We show qualitative results of this effect in \cref{fig:Control_sizes}. 
The general decrease in performance for smaller model sizes can be explained as follows. Control models with fewer parameters have less power and hence perform more similarly to the uncontrolled generative model, \ie Stable Diffusion, which performs worse in general. Note that the CLIP-Aesthetic score is highest for Stable Diffusion. Hence, our 1.7M model performs best for this score. 

\begin{table}[htb!]
\centering
\caption{{\bf Quantitative comparison of seven different approaches.} For each guidance type the best method is marked {\bf bold} and the two worst are marked \textcolor{BrickRed}{in red}. The results for ControlNet\cite{Zhang2023_ControlNet} with a semantic map guidance are shown in grey brackets since the authors state that the test images were part of the training set. 
\vspace{-0.3cm}
}
\scriptsize
\begin{tabular}{clccccc||c}
    \hline
    && Quality + Control & \multicolumn{2}{c}{Control} & \multicolumn{2}{c}{Quality} & \\
    &\multicolumn{1}{c}{Method} & FID $\downarrow$ & MSE-d $\downarrow$&  LPIPS $\downarrow$ &  CLIP-Sc $\uparrow$ & CLIP-Ae $\uparrow$ & SD    \\
       \hline \hline
   &Stable Diffusion    & \small 22.69      
                        & \small -     
                        & \small (0.618)    
                        & \small 28.40      
                        & \small 6.16     
                        &v1.5\\

   \hline
    \hline
    \parbox[t]{.5cm}{\multirow{7}{*}{\STAB{\rotatebox[origin=c]{90}{Depth}}}}
    &ControlNet (361M)      & \small \textcolor{black}{19.01}         
                            & \small \textcolor{BrickRed}{29.1}     
                            & \small \textcolor{BrickRed}{0.532}       
                            & \small \textcolor{black}{28.96}         
                            & \small {6.08}            
                        &v1.5\\

    &GLIGEN (231M)  & \small \textcolor{black}{19.15}         
                    & \small \textcolor{black}{21.2}     
                    & \small \textcolor{black}{0.490}      
                    & \small {29.03}         
                    & \small \textcolor{BrickRed}{5.81}             
                        &v1.4\\

    &T2I-Adapter[32] (77M)        & \small \textcolor{BrickRed}{20.29}         
                        & \small \textcolor{BrickRed}{31.4}     
                        & \small \textcolor{BrickRed}{0.526}      
                        & \small \textcolor{BrickRed}{28.80}         
                        & \small \textcolor{black}{5.98}           
                        &v1.5\\
    
    &UniControl (374M) & \small \textcolor{BrickRed}{26.80}         
                        & \small {19.9}     
                        & \small {0.487}      
                        & \small \textcolor{BrickRed}{28.04}         
                        & \small \textcolor{black}{5.97}             
                        &v1.5\\

    &Cocktail (378M) & \small /         
                        & \small /          
                        & \small /         
                        & \small /       
                        & \small /       
                        & / \\
    
    &Uni-ControlNet (459M) & \small {18.38}         
                        & \small \textcolor{black}{26.1}     
                        & \small \textcolor{black}{0.524}      
                        & \small \textcolor{black}{29.00}         
                        & \small \textcolor{BrickRed}{5.90}             
                        &v1.5\\
    
    &ControlNet-XS (55M)     & \small \textbf{16.36}         
                        & \small \textbf{19.6}     
                        & \small \textbf{0.468}       
                        & \small \textbf{29.21}         
                        & \small \textbf{6.09}            
                        &v1.5\\
    \hline
    &&&  HDD $\downarrow$  \\
    \hline
    
    \parbox[t]{.5cm}{\multirow{7}{*}{\STAB{\rotatebox[origin=c]{90}{Canny Edges}}}}&
    ControlNet (361M)
                    & \small \textcolor{black}{21.18}              
                    & \small \textcolor{BrickRed}{18.52}              
                    & \small \textcolor{BrickRed}{0.544}              
                    & \small \textcolor{BrickRed}{29.01}              
                    & \small \textbf{6.17}      
                        &v1.5\\

&GLIGEN (231M)      & \small \textcolor{BrickRed}{27.24}              
                    & \small \textbf{15.09}     
                    & \small \textcolor{black}{0.446}              
                    & \small \textcolor{black}{29.20}              
                    & \small \textcolor{BrickRed}{5.76}               
                        &v1.4\\

&T2I-Adapter (77M)          & \small \textcolor{black}{18.34}         
                    & \small \textcolor{BrickRed}{16.66}         
                    & \small \textcolor{black}{0.459}         
                    & \small \textcolor{black}{29.14}         
                    & \small \textcolor{BrickRed}{5.66}            
                        &v1.5\\

   &UniControl (374M) & \small \textcolor{BrickRed}{33.12}            
                    & \small \textcolor{black}{16.02}              
                    & \small \textbf{0.416}     
                    & \small \textcolor{BrickRed}{28.13}              
                    & \small \textcolor{black}{5.79}               
                        &v1.5\\

   &Cocktail (378M) & \small /         
                    & \small /          
                    & \small /         
                    & \small /       
                    & \small /       
                        & / \\

    &Uni-ControlNet (459M) & \small {17.37}         
                     & \small \textcolor{black}{15.94}     
                     & \small \textcolor{BrickRed}{0.460}       
                    & \small {29.28}         
                     & \small \textcolor{black}{5.87}            
                        &v1.5\\

    &ControlNet-XS (55M)     & \small \textbf{15.13}         
                     & \small {15.22}     
                     & \small {0.417}       
                     & \small \textbf{29.61}         
                     & \small {5.98}            
                        &v1.5\\

    \hline
    &&&    mIoU $\uparrow$\\
    \hline
    \parbox[t]{.5cm}{\multirow{7}{*}{\STAB{\rotatebox[origin=c]{90}{Semantic Map}}}}
    
    &ControlNet (361M)$^\ast$    & \small    (\textcolor{gray}{35.35})      
                        & \small (\textcolor{gray}{0.32})     
                        & \small (\textcolor{gray}{0.590})      
                        & \small   (\textcolor{gray}{30.22})       
                        & \small    (\textcolor{gray}{5.85})          
                        &v1.5\\
    
    &GLIGEN (231M)   & \small \textcolor{BrickRed}{29.83}       
                        & \small \textcolor{black}{0.25}     
                        & \small \textcolor{BrickRed}{0.608}         
                        & \small \textcolor{BrickRed}{29.82}          
                        & \small \textcolor{BrickRed}{5.84}           
                        &v1.4\\
                        
    &T2I-Adapter (77M)        & \small \textcolor{black}{23.76}         
                        & \small \textcolor{BrickRed}{0.22}     
                        & \small \textcolor{BrickRed}{0.613}      
                        & \small \textcolor{black}{30.31}         
                        & \small \textcolor{BrickRed}{5.69}             
                        &v1.4\\
    
    &UniControl (374M) & \small \textcolor{BrickRed}{39.26}         
                        & \small \textbf{0.31}     
                        & \small 0.552      
                        & \small \textcolor{BrickRed}{27.85}         
                        & \small {6.02}             
                        &v1.5\\
    
    &Cocktail[19] (378M) & \small \textcolor{black}{26.07}         
                        & \small \textcolor{BrickRed}{0.19}     
                        & \small {0.604}       
                        & \small \textcolor{black}{31.25}         
                        & \small \textbf{6.11}            
                        &v2.1\\
    
    &Uni-ControlNet (459M)    & \small {22.26}         
                        & \small \textcolor{black}{0.25}     
                        & \small \textcolor{black}{0.588}     
                        & \small \textcolor{black}{31.16}         
                        & \small \textcolor{black}{5.88}              
                        &v1.5\\

    &ControlNet-XS (55M)    & \small \textbf{17.70}         
                        & \small \textbf{0.31}     
                        & \small \textcolor{black}{\textbf{0.519}}      
                        & \small \textcolor{black}{\textbf{31.26}}         
                        & \small \textcolor{black}{5.95}             
                        &v1.5\\
    \hline
    &&&  HDD $\downarrow$   \\
    \hline
    \parbox[t]{.5cm}{\multirow{7}{*}{\STAB{\rotatebox[origin=c]{90}{Hum-Poses}}}}
    &ControlNet (361M)
                        & \small \textcolor{black}{23.82}         
                        & \small {8.58}     
                        & \small - 
                        & \small \textcolor{black}{28.14}         
                        & \small {6.18}             
                        &v1.5\\
    
    &GLIGEN (231M) & \small /         
                        & \small /     
                        & \small - 
                        & \small /         
                        & \small /         
                        & / \\
    
    &T2I-Adapter (77M)        & \small {23.16}         
                        & \small \textcolor{black}{8.81}     
                        & \small - 
                        & \small {28.21}         
                        & \small \textcolor{black}{6.05}              
                        &v1.5\\
    
    &UniControl (374M) & \small \textcolor{BrickRed}{54.63}         
                        & \small \textbf{8.53}     
                        & \small - 
                        & \small \textcolor{BrickRed}{24.72}         
                        & \small \textcolor{BrickRed}{5.87}            
                        &v1.5\\
    
    &Cocktail[19] (378M) & \small \textcolor{BrickRed}{26.44}         
                        & \small \textcolor{BrickRed}{9.87}     
                        & \small - 
                        & \small \textbf{28.28}         
                        & \small \textcolor{black}{6.15}             
                        &v2.1\\
    
    &Uni-ControlNet (459M) & \small \textbf{22.80}         
                        & \small \textcolor{BrickRed}{8.98}     
                        & \small - 
                        & \small \textcolor{BrickRed}{28.06}         
                        & \small \textcolor{BrickRed}{5.84}            
                        &v1.5\\
    
    &ControlNet-XS (55M)     & \small \textcolor{black}{23.58}         
                        & \small \textcolor{black}{8.67}     
                        & \small - 
                        & \small \textcolor{black}{28.15}        
                        & \small \textbf{6.21}            
                        &v1.5\\

    \hline
    
\end{tabular}
\vspace{-0.2cm}
\label{tab:quantative_analysis}
\end{table}

\subsection{Quantitative Comparison}
\label{subsec:quantitative}
\vspace{-0.2cm}
\cref{tab:quantative_analysis} compares our ControlNet-XS with six state-of-the-art control methods. We only use the officially published weights to not bias the comparison. We examined four different guidance types: pixel-accurate depth and canny-edge guidance, more loose semantic map guidance, and very loose human pose guidance.
Among all competitors, Uni-ControlNet is on average the best performing method, since it rarely has negative outliers and scores mostly among the top methods.
For pixel-accurate depth guidance  
our ControlNet-XS outperforms all other approaches for all metrics. This includes our baseline model ControlNet. For pixel-accurate canny-edge guidance, ControlNet-XS is either the best or second best performing. It is important to note that there is no clear runner-up method. For instance GLIGEN, which is second best for the HDD score, performs very poorly with respect to quality, \eg FID score. In general, GLIGEN and UniControl exhibit a trade-off between control (HDD/MSE-d, LPIPS) and quality (FID score) since they are not able to consistently exert proper control without diminishing image quality.        
For guidance with a semantic map, which is a more loose control, ControlNet-XS clearly outperforms all competitors, apart from the CLIP-Aesthetics score, where it ranks third. Especially noticeable is the tremendous gain in FID. 
Runner-ups are Uni-ControlNet and UniControl, although UniControl performs very poorly with respect to FID score.    
For the most loose control, \ie keypoint guidance for human poses, there is in general not much control that has to be enforced by the controlling model. All compared models perform similarly in terms of HDD score and keep the FID score in the vicinity of the score of the unguided base model. The two exceptions appear to be UniControl and Cocktail, which show a major drop in performance for the FID score. In general, the quantitative results for human poses have to be taken with a large grain of salt, given considerably less training data.

\begin{table}[ht!]
\vspace{-0.4cm}
    \centering
    \caption{\textbf{Comparison of inference and training times} of our ControlNet-XS and ControlNet~\cite{Zhang2023_ControlNet}, trained to control depth. Inference times are averaged over seven runs and we evaluate for 50 DDIM steps with a batch size of 10. The training time is given in NVIDIA A$100$ GPU hours.
    \vspace{-0.4cm}
    }
    \begin{tabular}{lcc}
    \hline
        \multicolumn{1}{c}{Method} & Inference $\downarrow$ & Training $\downarrow$\\
        \hline \hline
        ControlNet ($361$M)& 1min 11sec &  $\sim$ $500$h (A100) \\
        ControlNet-XS ($55$M) & 38sec & $\sim$ $200$h (A100)\\
        \hline
    \end{tabular}
\vspace{-0.4cm}
\label{tab:TrinRunTime}
\end{table}

In summary, we see that our improved communication mechanism plays an important role when it comes to pixel-accurate image guidance (depth and canny-edges) as well as more loose guidance (semantic map). For very loose guidance, such as human pose, the communication mechanism may play a less important role. However, even then our approach performs well across all metrics without any negative outlier. In general, only Uni-ControlNet and our approach show the behaviour of no negative outliers, while all other approaches seem to sometimes trade-off image quality for more accurate control.                

\cref{tab:TrinRunTime} compares inference and training times of ControlNet-XS and ControlNet. For both, we increase the speed by about a factor of $2$.

\begin{figure*}[htp!]
\vspace{-0.4cm}
    \centering
    \includegraphics[width=0.98\textwidth]{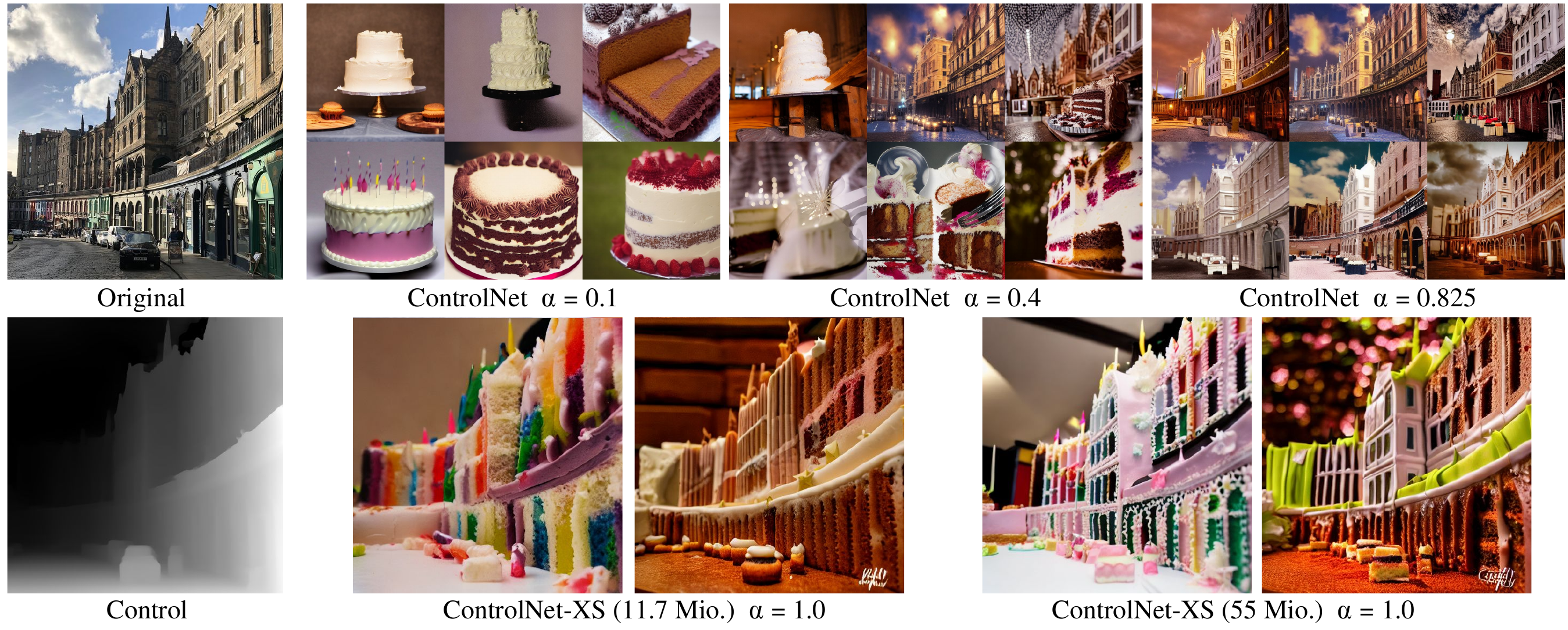}
    \vspace{-0.3cm}
    \caption{{\bf Semantic bias for depth control.} Given the control depth map of a street scene and an unrelated text-prompt: ``high quality photo of a delicious cake, 4k image''. The large-sized ($361$M) ControlNet~\cite{Zhang2023_ControlNet} has a semantic bias and is unable to produce a cake scene with the given depth, independent of control strength $\alpha$. Our small-sized models with $11.7$M and $55$M respectively mitigate this bias.  
    } 
    \vspace{-0.8cm}             
    \label{fig:Semantic bias}
\end{figure*}

\subsection{Semantic Bias of Large Control Models}
\label{subsec:bias}

We have seen already that the large-sized ControlNet~\cite{Zhang2023_ControlNet} needs generative power to produce good results. However, this can induce a semantic bias as shown in \cref{fig:Semantic bias}.
The images are generated with a control depth map of a street scene and an unrelated text-prompt: ``high quality photo of a delicious cake, 4k image''. Note that these are not contradicting control inputs but the inputs rather challenge the generative process to produce a creative solution with a cake in form of a street scene. We see that ControlNet-XS with $11.7$M parameters is able to produce impressive results, followed by results of the $55$M model. In contrast, ControlNet~\cite{Zhang2023_ControlNet} is not able to produce satisfying results, even when adjusting the control strength $\alpha$.\footnote{The output signals of the controlling network are added with a global weighting $\alpha$ to the output signals of the generation network at the respective neural blocks. This weighting can be adjusted at test time.} Note that $\alpha=0.825$ is the default for ControlNet. With this default value, ControlNet shows proper house facade textures, while ControlNet-XS shows typical cake textures such as ``sponge'', ``marzipan'' or ``icing''. We conjecture that the reason is a semantic bias induced by large control models. A large control model can use its power to add semantic meaning to input depth maps. This semantic bias cannot be removed by adapting $\alpha$. Here $\alpha=0.4$ was the ``sweet spot'' where ControlNet suddenly transitions from producing images of a cake to images of a street scene. In the supplement, we show that a large-sized ControlNet-XS also has this semantic bias.

\vspace{-0.2cm}
\subsection{Evaluation with Stable Diffusion XL}
\label{subsec:eval_sdxl}
\vspace{-0.1cm}
We evaluate our ControlNet-XS model with Stable Diffusion XL~\cite{Podell2023_SDXL} as generative model. Stable Diffusion XL has about $2.6$B parameters and hence is over three times larger than its predecessor Stable Diffusion. We are able to train a ControlNet-XS for depth control which has only $20$M parameters, \ie less than $1$\% of parameters of the generative model. Our model provides good control, \ie MSE-d score is $22.6$ in contrast to $123.2$ of the uncontrolled Stable Diffusion XL. Furthermore, we achieve high quality results with a low FID score of $18.75$. ControlNet-XS is also considerably superior to the T2I-Adapter~\cite{Mou2023_T2I} with an FID score of $61.03$ and MSE-d score of $49$. A qualitative result is shown in \cref{fig:teaser}. We refer to the supplement for more results and a discussion. 

\vspace{-0.4cm}
\section{Conclusion and Limitation}
\label{sec:conclusion}
\vspace{-0.1cm}
We have analyzed existing approaches for controlling pre-trained text-to-image Diffusion Models with respect to their communication mechanism with the generative model. We proposed a new bidirectional, high-frequency and large-bandwidth communication. This led to the development of ControlNet-XS, a small-sized controlling network that outperforms the state-of-the-art for pixel-level image guidance. One major limitation in this field is a missing unifying benchmark with consistent evaluation protocols and ideally a metric that truly represents human judgment.  
There are many exciting directions for future work. One next step is to integrate our approach into the multi-image guidance approaches, which are based on a feedback-control system, but also approaches in related fields such as video translation~\cite{Feng2023_ccedit, Zhao2023_MakeAProtagonist} and controlled 3D object generation~\cite{Zhang2023_AvatarVerse, Huang2023_DreamControl} should benefit from our method.          

\section*{Acknowledgements}
We thank Nicolas Bender for his help in conducting experiments. The project has been supported by the Konrad Zuse School of Excellence in Learning and Intelligent Systems (ELIZA) funded by the German Academic Exchange Service (DAAD). The project has also been supported by the Trilateral DFG Research Program (Germany-France-Japan). The project was also  support by the state of Baden-Württemberg through bwHPC and the German Research  Foundation (DFG) through grant INST 35/1597-1 FUGG.


\bibliographystyle{splncs04}
\bibliography{egbib}

\begin{thebibliography}{10}
\providecommand{\url}[1]{\texttt{#1}}
\providecommand{\urlprefix}{URL }
\providecommand{\doi}[1]{https://doi.org/#1}

\bibitem{Midjourney2023}
Midjourney (2023), \url{https://www.midjourney.com/}

\bibitem{Bar-Tal2022_Text2Live}
Bar-Tal, O., Ofri-Amar, D., Fridman, R., Kasten, Y., Dekel, T.: {Text2live: Text-driven layered image and video editing}. In: European Conference on Computer Vision. pp. 707--723 (2022)

\bibitem{Betker2023_dalle3}
Betker, J., Goh, G., Jing, L., Brooks, T., Wang, J., Li, L., Ouyang, L., Zhuang, J., Lee, J., Guo, Y., Manassra, W., Dhariwal, P., Chu, C., Jiao, Y., Ramesh, A.: {Improving Image Generation with Better Captions}  (2023)

\bibitem{Brooks2023_InstructPix2Pix}
Brooks, T., Holynski, A., Efros, A.A.: {Instructpix2pix: Learning to follow image editing instructions}. In: Proceedings of the IEEE/CVF Conference on Computer Vision and Pattern Recognition. pp. 18392--18402 (2023)

\bibitem{Caesar2018_COCO-STUFF}
Caesar, H., Uijlings, J., Ferrari, V.: {Coco-stuff: Thing and stuff classes in context}. In: Proceedings of the IEEE conference on computer vision and pattern recognition. pp. 1209--1218 (2018)

\bibitem{canny1986computational}
Canny, J.: A computational approach to edge detection. IEEE Transactions on pattern analysis and machine intelligence (6),  679--698 (1986)

\bibitem{Cao2017_OpenPose}
Cao, Z., Simon, T., Wei, S.E., Sheikh, Y.: {Realtime multi-person 2d pose estimation using part affinity fields}. In: Proceedings of the IEEE conference on computer vision and pattern recognition. pp. 7291--7299 (2017)

\bibitem{Chen2024}
Chen, M., Laina, I., Vedaldi, A.: {Training-free layout control with cross-attention guidance}. In: Proceedings of the IEEE/CVF Winter Conference on Applications of Computer Vision. pp. 5343--5353 (2024)

\bibitem{chen2022}
Chen, Z., Duan, Y., Wang, W., He, J., Lu, T., Dai, J., Qiao, Y.: Vision transformer adapter for dense predictions. arXiv preprint arXiv:2205.08534  (2022)

\bibitem{Choi2023_Custom-Edit}
Choi, J., Choi, Y., Kim, Y., Kim, J., Yoon, S.: {Custom-edit: Text-guided image editing with customized diffusion models}. arXiv preprint arXiv:2305.15779  (2023)

\bibitem{Couairon2022_DiffEdit}
Couairon, G., Verbeek, J., Schwenk, H., Cord, M.: {Diffedit: Diffusion-based semantic image editing with mask guidance}. arXiv preprint arXiv:2210.11427  (2022)

\bibitem{Devlin2019_BERT}
Devlin, J., Chang, M.W., Lee, K., Toutanova, K.: {BERT: Pre-training of Deep Bidirectional Transformers for Language Understanding}. In: Burstein, J., Doran, C., Solorio, T. (eds.) Proceedings of the 2019 Conference of the North American Chapter of the Association for Computational Linguistics: Human Language Technologies, Volume 1 (Long and Short Papers). pp. 4171--4186. Association for Computational Linguistics, Minneapolis, Minnesota (2019). \doi{10.18653/v1/N19-1423}, \url{https://aclanthology.org/N19-1423}

\bibitem{Dhariwal2021}
Dhariwal, P., Nichol, A.: {Diffusion models beat gans on image synthesis}. Advances in neural information processing systems  \textbf{34},  8780--8794 (2021)

\bibitem{Ding2021_CogView}
Ding, M., Yang, Z., Hong, W., Zheng, W., Zhou, C., Yin, D., Lin, J., Zou, X., Shao, Z., Yang, H., Tang, J.: {CogView: Mastering Text-to-Image Generation via Transformers}. In: Ranzato, M., Beygelzimer, A., Dauphin, Y., Liang, P.S., Vaughan, J.W. (eds.) Advances in Neural Information Processing Systems. vol.~34, pp. 19822--19835. Curran Associates, Inc. (2021), \url{https://proceedings.neurips.cc/paper_files/paper/2021/file/a4d92e2cd541fca87e4620aba658316d-Paper.pdf}

\bibitem{TamingTransformers}
Esser, P., Rombach, R., Ommer, B.: {Taming transformers for high-resolution image synthesis}. In: Proceedings of the IEEE/CVF conference on computer vision and pattern recognition. pp. 12873--12883 (2021)

\bibitem{Feng2023_ccedit}
Feng, R., Weng, W., Wang, Y., Yuan, Y., Bao, J., Luo, C., Chen, Z., Guo, B.: {Ccedit: Creative and controllable video editing via diffusion models}. arXiv preprint arXiv:2309.16496  (2023)

\bibitem{Gafni2022_MakeAScene}
Gafni, O., Polyak, A., Ashual, O., Sheynin, S., Parikh, D., Taigman, Y.: {Make-A-Scene: Scene-Based Text-to-Image Generation with Human Priors}. In: {Avidan Shai and Brostow}, G., Moustapha, C., Maria, F.G., Tal, H. (eds.) Computer Vision – ECCV 2022. pp. 89--106. Springer Nature Switzerland, Cham (2022)

\bibitem{Gal2022}
Gal, R., Alaluf, Y., Atzmon, Y., Patashnik, O., Bermano, A.H., Chechik, G., Cohen-Or, D.: {An image is worth one word: Personalizing text-to-image generation using textual inversion}. arXiv preprint arXiv:2208.01618  (2022)

\bibitem{Goel2023_PAIR-Diffusion}
Goel, V., Peruzzo, E., Jiang, Y., Xu, D., Sebe, N., Darrell, T., Wang, Z., Shi, H.: {PAIR-Diffusion: Object-Level Image Editing with Structure-and-Appearance Paired Diffusion Models} (2023), \url{http://arxiv.org/abs/2303.17546}

\bibitem{GAN}
Goodfellow, I., Pouget-Abadie, J., Mirza, M., Xu, B., Warde-Farley, D., Ozair, S., Courville, A., Bengio, Y.: {Generative adversarial nets}. In: Advances in neural information processing systems. pp. 2672--2680 (2014)

\bibitem{He2023_locGuidance}
He, Y., Salakhutdinov, R., Kolter, J.Z.: {Localized Text-to-Image Generation for Free via Cross Attention Control}. arXiv preprint arXiv:2306.14636  (2023)

\bibitem{Hessel2021_CLIPScore}
Hessel, J., Holtzman, A., Forbes, M., Bras, R.L., Choi, Y.: {Clipscore: A reference-free evaluation metric for image captioning}. arXiv preprint arXiv:2104.08718  (2021)

\bibitem{FID}
Heusel, M., Ramsauer, H., Unterthiner, T., Nessler, B., Hochreiter, S.: {Gans trained by a two time-scale update rule converge to a local nash equilibrium}. Advances in neural information processing systems  \textbf{30} (2017)

\bibitem{Ho2020}
Ho, J., Jain, A., Abbeel, P.: {Denoising diffusion probabilistic models}. Advances in neural information processing systems  \textbf{33},  6840--6851 (2020)

\bibitem{Ho2022}
Ho, J., Saharia, C., Chan, W., Fleet, D.J., Norouzi, M., Salimans, T.: {Cascaded diffusion models for high fidelity image generation}. The Journal of Machine Learning Research  \textbf{23}(1),  2249--2281 (2022)

\bibitem{Hu2021_LoRA}
Hu, E.J., Shen, Y., Wallis, P., Allen-Zhu, Z., Li, Y., Wang, S., Wang, L., Chen, W.: {LoRA: Low-Rank Adaptation of Large Language Models} (2021)

\bibitem{Hu2024_InstructImagen}
Hu, H., Chan, K.C.K., Su, Y.C., Chen, W., Li, Y., Sohn, K., Zhao, Y., Ben, X., Gong, B., Cohen, W.: {Instruct-Imagen: Image generation with multi-modal instruction}. arXiv preprint arXiv:2401.01952  (2024)

\bibitem{Cocktail}
Hu, M., Zheng, J., Liu, D., Zheng, C., Wang, C., Tao, D., Cham, T.J.: {Cocktail: Mixing Multi-Modality Control for Text-Conditional Image Generation}. In: Thirty-seventh Conference on Neural Information Processing Systems (2023)

\bibitem{Huang2023_Composer}
Huang, L., Chen, D., Liu, Y., Shen, Y., Zhao, D., Zhou, J.: {Composer: Creative and controllable image synthesis with composable conditions}. arXiv preprint arXiv:2302.09778  (2023)

\bibitem{Huang2023_DreamControl}
Huang, T., Zeng, Y., Zhang, Z., Xu, W., Xu, H., Xu, S., Lau, R.W.H., Zuo, W.: {Dreamcontrol: Control-based text-to-3d generation with 3d self-prior}. arXiv preprint arXiv:2312.06439  (2023)

\bibitem{ImageToImage}
Isola, P., Zhu, J.Y., Zhou, T., Efros, A.A.: {Image-to-image translation with conditional adversarial networks}. In: Proceedings of the IEEE conference on computer vision and pattern recognition. pp. 1125--1134 (2017)

\bibitem{Kang2023_GigaGAN}
Kang, M., Zhu, J.Y., Zhang, R., Park, J., Shechtman, E., Paris, S., Park, T.: {Scaling up gans for text-to-image synthesis}. In: Proceedings of the IEEE/CVF Conference on Computer Vision and Pattern Recognition. pp. 10124--10134 (2023)

\bibitem{CelebA-HQ}
Karras, T., Aila, T., Laine, S., Lehtinen, J.: {Progressive growing of gans for improved quality, stability, and variation}. arXiv preprint arXiv:1710.10196  (2017)

\bibitem{StyleGAN3}
Karras, T., Aittala, M., Laine, S., H{\"{a}}rk{\"{o}}nen, E., Hellsten, J., Lehtinen, J., Aila, T.: {Alias-free generative adversarial networks}. Advances in Neural Information Processing Systems  \textbf{34} (2021)

\bibitem{Karras2019_StyleGAN}
Karras, T., Laine, S., Aila, T.: {A style-based generator architecture for generative adversarial networks}. In: Proceedings of the IEEE/CVF conference on computer vision and pattern recognition. pp. 4401--4410 (2019)

\bibitem{StyleGAN2}
Karras, T., Laine, S., Aittala, M., Hellsten, J., Lehtinen, J., Aila, T.: {Analyzing and improving the image quality of stylegan}. In: Proceedings of the IEEE/CVF conference on computer vision and pattern recognition. pp. 8110--8119 (2020)

\bibitem{Kingma2021}
Kingma, D., Salimans, T., Poole, B., Ho, J.: {Variational diffusion models}. Advances in neural information processing systems  \textbf{34},  21696--21707 (2021)

\bibitem{Li2023_DreamEdit}
Li, T., Ku, M., Wei, C., Chen, W.: {DreamEdit: Subject-driven Image Editing}. arXiv preprint arXiv:2306.12624  (2023)

\bibitem{Li2024_UNIMO-G}
Li, W., Xu, X., Liu, J., Xiao, X.: {UNIMO-G: Unified Image Generation through Multimodal Conditional Diffusion}. arXiv preprint arXiv:2401.13388  (2024)

\bibitem{Li2022}
Li, Y., Mao, H., Girshick, R., He, K.: {Exploring Plain Vision Transformer Backbones for Object Detection}. In: Computer Vision – ECCV 2022, pp. 280--296. Springer Nature Switzerland (2022). \doi{10.1007/978-3-031-20077-9_17}, \url{http://dx.doi.org/10.1007/978-3-031-20077-9_17}

\bibitem{Li2023_GLIGEN}
Li, Y., Liu, H., Wu, Q., Mu, F., Yang, J., Gao, J., Li, C., Lee, Y.J.: {Gligen: Open-set grounded text-to-image generation}. In: Proceedings of the IEEE/CVF Conference on Computer Vision and Pattern Recognition. pp. 22511--22521 (2023)

\bibitem{Lin2014_COCO}
Lin, T.Y., Maire, M., Belongie, S., Hays, J., Perona, P., Ramanan, D., Doll{\'{a}}r, P., Zitnick, C.L.: {Microsoft coco: Common objects in context}. In: Computer Vision–ECCV 2014: 13th European Conference, Zurich, Switzerland, September 6-12, 2014, Proceedings, Part V 13. pp. 740--755. Springer (2014)

\bibitem{liu2021cross}
Liu, L., Chen, J., Wu, H., Li, G., Li, C., Lin, L.: Cross-modal collaborative representation learning and a large-scale rgbt benchmark for crowd counting. In: Proceedings of the IEEE/CVF conference on computer vision and pattern recognition. pp. 4823--4833 (2021)

\bibitem{lu2022dual}
Lu, C., Xia, M., Qian, M., Chen, B.: Dual-branch network for cloud and cloud shadow segmentation. IEEE Transactions on Geoscience and Remote Sensing  \textbf{60},  1--12 (2022)

\bibitem{Lukovnikov2024_Layout2Image}
Lukovnikov, D., Fischer, A.: {Layout-to-Image Generation with Localized Descriptions using ControlNet with Cross-Attention Control}. arXiv preprint arXiv:2402.13404  (2024)

\bibitem{Mao12022_UNIPELT}
Mao1, Y., Mathias, L., Hou, R., Almahairi, A., Ma, H., Han, J., Yih, W.t., Khabsa, M.: {UNIPELT: A Unified Framework for Parameter-Efficient Language Model Tuning}. In: Proceedings of the 60th Annual Meeting of the Association for Computational Linguistics Volume 1: Long Papers. vol.~1, pp. 6253--6264. ACL (2022)

\bibitem{McCloskey1989}
McCloskey, M., Cohen, N.J.: {Catastrophic Interference in Connectionist Networks: The Sequential Learning Problem}. vol.~24, pp. 109--165. Academic Press (1989). \doi{https://doi.org/10.1016/S0079-7421(08)60536-8}, \url{https://www.sciencedirect.com/science/article/pii/S0079742108605368}

\bibitem{Mou2023_T2I}
Mou, C., Wang, X., Xie, L., Wu, Y., Zhang, J., Qi, Z., Shan, Y., Qie, X.: {T2i-adapter: Learning adapters to dig out more controllable ability for text-to-image diffusion models}. arXiv preprint arXiv:2302.08453  (2023)

\bibitem{Murez2018}
Murez, Z., Kolouri, S., Kriegman, D., Ramamoorthi, R., Kim, K.: {Image to image translation for domain adaptation}. In: Proceedings of the IEEE conference on computer vision and pattern recognition. pp. 4500--4509 (2018)

\bibitem{Nichol2021_GLIDE}
Nichol, A., Dhariwal, P., Ramesh, A., Shyam, P., Mishkin, P., McGrew, B., Sutskever, I., Chen, M.: {Glide: Towards photorealistic image generation and editing with text-guided diffusion models}. arXiv preprint arXiv:2112.10741  (2021)

\bibitem{Nichol2021b}
Nichol, A.Q., Dhariwal, P.: {Improved denoising diffusion probabilistic models}. In: International Conference on Machine Learning. pp. 8162--8171. PMLR (2021)

\bibitem{Patel2024_lambda-Eclipse}
Patel, M., Jung, S., Baral, C., Yang, Y.: {$\lambda $-ECLIPSE: Multi-Concept Personalized Text-to-Image Diffusion Models by Leveraging CLIP Latent Space}. arXiv preprint arXiv:2402.05195  (2024)

\bibitem{Pfeiffer2021_AdapterFusion}
Pfeiffer, J., Kamath, A., Ruckl, A., Cho, K., Gurevych1, I.: {AdapterFusion: Non-Destructive Task Composition for Transfer Learning}. In: Proceedings of the 16th Conference of the European Chapter of the Association for Computational Linguistics. pp. 487--503. ACL (2021)

\bibitem{Podell2023_SDXL}
Podell, D., English, Z., Lacey, K., Blattmann, A., Dockhorn, T., M{\"{u}}ller, J., Penna, J., Rombach, R.: {SDXL: Improving Latent Diffusion Models for High-Resolution Image Synthesis} (2023)

\bibitem{UniControl}
Qin, C., Zhang, S., Yu, N., Feng, Y., Yang, X., Zhou, Y., Wang, H., Niebles, J.C., Xiong, C., Savarese, S.: {UniControl: A Unified Diffusion Model for Controllable Visual Generation In the Wild}. arXiv preprint arXiv:2305.11147  (2023)

\bibitem{Radford2021_CLIP}
Radford, A., Kim, J.W., Hallacy, C., Ramesh, A., Goh, G., Agarwal, S., Sastry, G., Askell, A., Mishkin, P., Clark, J., Others: {Learning transferable visual models from natural language supervision}. In: International conference on machine learning. pp. 8748--8763 (2021)

\bibitem{radford2016unsupervised}
Radford, A., Metz, L., Chintala, S.: Unsupervised representation learning with deep convolutional generative adversarial networks (2016)

\bibitem{Raffel2020}
Raffel, C., Shazeer, N., Roberts, A., Lee, K., Narang, S., Matena, M., Zhou, Y., Li, W., Liu, P.J.: {Exploring the limits of transfer learning with a unified text-to-text transformer}. The Journal of Machine Learning Research  \textbf{21}(1),  5485--5551 (2020)

\bibitem{Ramesh2022_dalle2}
Ramesh, A., Dhariwal, P., Nichol, A., Chu, C., Chen, M.: {Hierarchical Text-Conditional Image Generation with CLIP Latents} (2022)

\bibitem{Ramesh2021_dalle1}
Ramesh, A., Pavlov, M., Goh, G., Gray, S., Voss, C., Radford, A., Chen, M., Sutskever, I.: {Zero-Shot Text-to-Image Generation}. CoRR  \textbf{abs/2102.1} (2021), \url{https://arxiv.org/abs/2102.12092}

\bibitem{Ranftl2020_Midas}
Ranftl, R., Lasinger, K., Hafner, D., Schindler, K., Koltun, V.: {Towards robust monocular depth estimation: Mixing datasets for zero-shot cross-dataset transfer}. IEEE transactions on pattern analysis and machine intelligence  \textbf{44}(3),  1623--1637 (2020)

\bibitem{Rebuffi2018}
Rebuffi, S.A., Bilen, H., Vedaldi, A.: {Efficient Parametrization of Multi-Domain Deep Neural Networks}. In: Proceedings of the IEEE Conference on Computer Vision and Pattern Recognition (CVPR) (2018)

\bibitem{Reed2016a}
Reed, S., Akata, Z., Yan, X., Logeswaran, L., Schiele, B., Lee, H.: {Generative adversarial text to image synthesis}. In: International conference on machine learning. pp. 1060--1069. PMLR (2016)

\bibitem{Reed2016}
Reed, S.E., Akata, Z., Mohan, S., Tenka, S., Schiele, B., Lee, H.: {Learning what and where to draw}. Advances in neural information processing systems  \textbf{29} (2016)

\bibitem{Rombach2022_LDM}
Rombach, R., Blattmann, A., Lorenz, D., Esser, P., Ommer, B.: {High-Resolution Image Synthesis with Latent Diffusion Models}. In: Proceedings of the IEEE Conference on Computer Vision and Pattern Recognition (CVPR) (2022), \url{https://github.com/CompVis/latent-diffusion}

\bibitem{Rosenfeld2020}
Rosenfeld, A., Tsotsos, J.K.: {Incremental Learning Through Deep Adaptation}. IEEE Transactions on Pattern Analysis and Machine Intelligence  \textbf{42}(3),  651--663 (2020). \doi{10.1109/TPAMI.2018.2884462}

\bibitem{Ruiz2023_dreambooth}
Ruiz, N., Li, Y., Jampani, V., Pritch, Y., Rubinstein, M., Aberman, K.: {Dreambooth: Fine tuning text-to-image diffusion models for subject-driven generation}. In: Proceedings of the IEEE/CVF Conference on Computer Vision and Pattern Recognition. pp. 22500--22510 (2023)

\bibitem{Russakovsky2015_ImageNet}
Russakovsky, O., Deng, J., Su, H., Krause, J., Satheesh, S., Ma, S., Huang, Z., Karpathy, A., Khosla, A., Bernstein, M.: {Imagenet large scale visual recognition challenge}. International journal of computer vision  \textbf{115},  211--252 (2015)

\bibitem{Saharia2022b_palette}
Saharia, C., Chan, W., Chang, H., Lee, C., Ho, J., Salimans, T., Fleet, D., Norouzi, M.: {Palette: Image-to-Image Diffusion Models}. In: Special Interest Group on Computer Graphics and Interactive Techniques Conference Proceedings. ACM (2022). \doi{10.1145/3528233.3530757}, \url{http://dx.doi.org/10.1145/3528233.3530757}

\bibitem{Saharia2022_imagen}
Saharia, C., Chan, W., Saxena, S., Li, L., Whang, J., Denton, E.L., Ghasemipour, K., {Gontijo Lopes}, R., {Karagol Ayan}, B., Salimans, T., Ho, J., Fleet, D.J., Norouzi, M.: {Photorealistic Text-to-Image Diffusion Models with Deep Language Understanding}. In: Koyejo, S., Mohamed, S., Agarwal, A., Belgrave, D., Cho, K., Oh, A. (eds.) Advances in Neural Information Processing Systems. vol.~35, pp. 36479--36494. Curran Associates, Inc. (2022), \url{https://proceedings.neurips.cc/paper_files/paper/2022/file/ec795aeadae0b7d230fa35cbaf04c041-Paper-Conference.pdf}

\bibitem{Sauer2023_StyleGAN-t}
Sauer, A., Karras, T., Laine, S., Geiger, A., Aila, T.: {Stylegan-t: Unlocking the power of gans for fast large-scale text-to-image synthesis}. arXiv preprint arXiv:2301.09515  (2023)

\bibitem{Sauer2022_StyleGAN-XL}
Sauer, A., Schwarz, K., Geiger, A.: {StyleGAN-XL: Scaling StyleGAN to Large Diverse Datasets}. In: ACM SIGGRAPH 2022 conference proceedings. pp. 1--10 (2022). \doi{10.1145/3528233.3530738}

\bibitem{Schuhmann2022_LaionAE}
Schuhmann, C., Beaumont, R., Vencu, R., Gordon, C., Wightman, R., Cherti, M., Coombes, T., Katta, A., Mullis, C., Wortsman, M.: {Laion-5b: An open large-scale dataset for training next generation image-text models}. Advances in Neural Information Processing Systems  \textbf{35},  25278--25294 (2022)

\bibitem{DM_Origin}
Sohl-Dickstein, J., Weiss, E., Maheswaranathan, N., Ganguli, S.: {Deep unsupervised learning using nonequilibrium thermodynamics}. In: International Conference on Machine Learning. pp. 2256--2265. PMLR (2015)

\bibitem{Song2020}
Song, J., Meng, C., Ermon, S.: {Denoising diffusion implicit models}. arXiv preprint arXiv:2010.02502  (2020)

\bibitem{Stickland2019}
Stickland, A.C., Murray, I.: {BERT and PALs: Projected Attention Layers for Efficient Adaptation in Multi-Task Learning}. In: Chaudhuri, K., Salakhutdinov, R. (eds.) Proceedings of the 36th International Conference on Machine Learning. vol.~97, pp. 5986--5995. PMLR (2019), \url{https://proceedings.mlr.press/v97/stickland19a.html}

\bibitem{tang2020xinggan}
Tang, H., Bai, S., Zhang, L., Torr, P.H., Sebe, N.: Xinggan for person image generation. In: Computer Vision--ECCV 2020: 16th European Conference, Glasgow, UK, August 23--28, 2020, Proceedings, Part XXV 16. pp. 717--734. Springer (2020)

\bibitem{Tao2022}
Tao, M., Tang, H., Wu, F., Jing, X.Y., Bao, B.K., Xu, C.: {Df-gan: A simple and effective baseline for text-to-image synthesis}. In: Proceedings of the IEEE/CVF Conference on Computer Vision and Pattern Recognition. pp. 16515--16525 (2022)

\bibitem{Wah2011_caltecBirds}
Wah, C., Branson, S., Welinder, P., Perona, P., Belongie, S.: {The caltech-ucsd birds-200-2011 dataset}  (2011)

\bibitem{Wang2022}
Wang, T., Zhang, T., Zhang, B., Ouyang, H., Chen, D., Chen, Q., Wen, F.: {Pretraining is All You Need for Image-to-Image Translation} (2022)

\bibitem{wang2019cfsnet}
Wang, W., Guo, R., Tian, Y., Yang, W.: Cfsnet: Toward a controllable feature space for image restoration. In: Proceedings of the IEEE/CVF international conference on computer vision. pp. 4140--4149 (2019)

\bibitem{Xia2023}
Xia, Y., Monica, J., Chao, W.L., Hariharan, B., Weinberger, K.Q., Campbell, M.: {Image-to-Image Translation for Autonomous Driving from Coarsely-Aligned Image Pairs}. In: 2023 IEEE International Conference on Robotics and Automation (ICRA). pp. 7756--7762. IEEE (2023)

\bibitem{Xu2018}
Xu, T., Zhang, P., Huang, Q., Zhang, H., Gan, Z., Huang, X., He, X.: {Attngan: Fine-grained text to image generation with attentional generative adversarial networks}. In: Proceedings of the IEEE conference on computer vision and pattern recognition. pp. 1316--1324 (2018)

\bibitem{Xu2023}
Xu, X., Wang, Z., Zhang, G., Wang, K., Shi, H.: {Versatile Diffusion: Text, Images and Variations All in One Diffusion Model}. In: Proceedings of the IEEE/CVF International Conference on Computer Vision (ICCV). pp. 7754--7765 (2023)

\bibitem{Yang2023}
Yang, B., Gu, S., Zhang, B., Zhang, T., Chen, X., Sun, X., Chen, D., Wen, F.: {Paint by example: Exemplar-based image editing with diffusion models}. In: Proceedings of the IEEE/CVF Conference on Computer Vision and Pattern Recognition. pp. 18381--18391 (2023)

\bibitem{Yu2022}
Yu, J., Xu, Y., Koh, J.Y., Luong, T., Baid, G., Wang, Z., Vasudevan, V., Ku, A., Yang, Y., Ayan, B.K., Others: {Scaling autoregressive models for content-rich text-to-image generation}. arXiv preprint arXiv:2206.10789  \textbf{2}(3), ~5 (2022)

\bibitem{Zhang2017}
Zhang, H., Xu, T., Li, H., Zhang, S., Wang, X., Huang, X., Metaxas, D.N.: {Stackgan: Text to photo-realistic image synthesis with stacked generative adversarial networks}. In: Proceedings of the IEEE international conference on computer vision. pp. 5907--5915 (2017)

\bibitem{Zhang2023_AvatarVerse}
Zhang, H., Chen, B., Yang, H., Qu, L., Wang, X., Chen, L., Long, C., Zhu, F., Du, K., Zheng, M.: {Avatarverse: High-quality \& stable 3d avatar creation from text and pose}. arXiv preprint arXiv:2308.03610  (2023)

\bibitem{Zhang2023_ControlNet}
Zhang, L., Rao, A., Agrawala, M.: {Adding Conditional Control to Text-to-Image Diffusion Models}. In: Proceedings of the IEEE/CVF International Conference on Computer Vision (ICCV). pp. 3836--3847 (2023)

\bibitem{LPIPS}
Zhang, R., Isola, P., Efros, A.A., Shechtman, E., Wang, O.: {The unreasonable effectiveness of deep features as a perceptual metric}. In: Proceedings of the IEEE conference on computer vision and pattern recognition. pp. 586--595 (2018)

\bibitem{UniControlNet}
Zhao, S., Chen, D., Chen, Y.C., Bao, J., Hao, S., Yuan, L., Wong, K.Y.K.: {Uni-controlnet: All-in-one control to text-to-image diffusion models}. Advances in Neural Information Processing Systems  \textbf{36} (2024)

\bibitem{Zhao2023_lcc}
Zhao, Y., Peng, L., Yang, Y., Luo, Z., Li, H., Chen, Y., Zhao, W., Liu, W., Wu, B.: {Local Conditional Controlling for Text-to-Image Diffusion Models}. arXiv preprint arXiv:2312.08768  (2023)

\bibitem{Zhao2023_MakeAProtagonist}
Zhao, Y., Xie, E., Hong, L., Li, Z., Lee, G.H.: {Make-A-Protagonist: Generic Video Editing with An Ensemble of Experts}. arXiv preprint arXiv:2305.08850  (2023)

\bibitem{Zhu2017}
Zhu, J.Y., Park, T., Isola, P., Efros, A.A.: {Unpaired image-to-image translation using cycle-consistent adversarial networks}. In: Proceedings of the IEEE international conference on computer vision. pp. 2223--2232 (2017)

\bibitem{Zhu2019}
Zhu, M., Pan, P., Chen, W., Yang, Y.: {Dm-gan: Dynamic memory generative adversarial networks for text-to-image synthesis}. In: Proceedings of the IEEE/CVF conference on computer vision and pattern recognition. pp. 5802--5810 (2019)

\end{thebibliography}

\appendix

\section*{Appendix}
In the following, we provide details about the architecture of ControlNet-XS (\cref{sec:supp_architecture}) and the reasoning behind the sufficiency of its encoder (\cref{sec:supp_sensitivity}). In \cref{sec:supp_training_details}, we explain how we trained all our models for different controls. To validate the ablation study from the main article on the architecture type and model size (Sec.4.2), we examine the same setup for the control with Canny-edges in \cref{sec:supp_quantitative_edges}. Additional qualitative results for ControlNet-XS applied to Stable Diffusion~\cite{Rombach2022_LDM} and Stable Diffusion XL~\cite{Podell2023_SDXL} are provided in \cref{sec:supp_qualitative_results}, for all trained controls. In \cref{sec:supp_prompts}, we provide the text-prompts that were used to generate the images from the main article and the supplementary material. \cref{sec:supp_control_fidelity} and \cref{sec:supp_semantic_bias} extend the analysis from the main article of the fidelity of the control (Figure 4, main article) and the semantic bias of large control models (Sec.4.4), respectively, with additional examples.


\section{Architecture}
\label{sec:supp_architecture}
\begin{figure}[htp!]
    \centering
    \captionsetup{type=figure}
    \includegraphics[width=.9\linewidth]{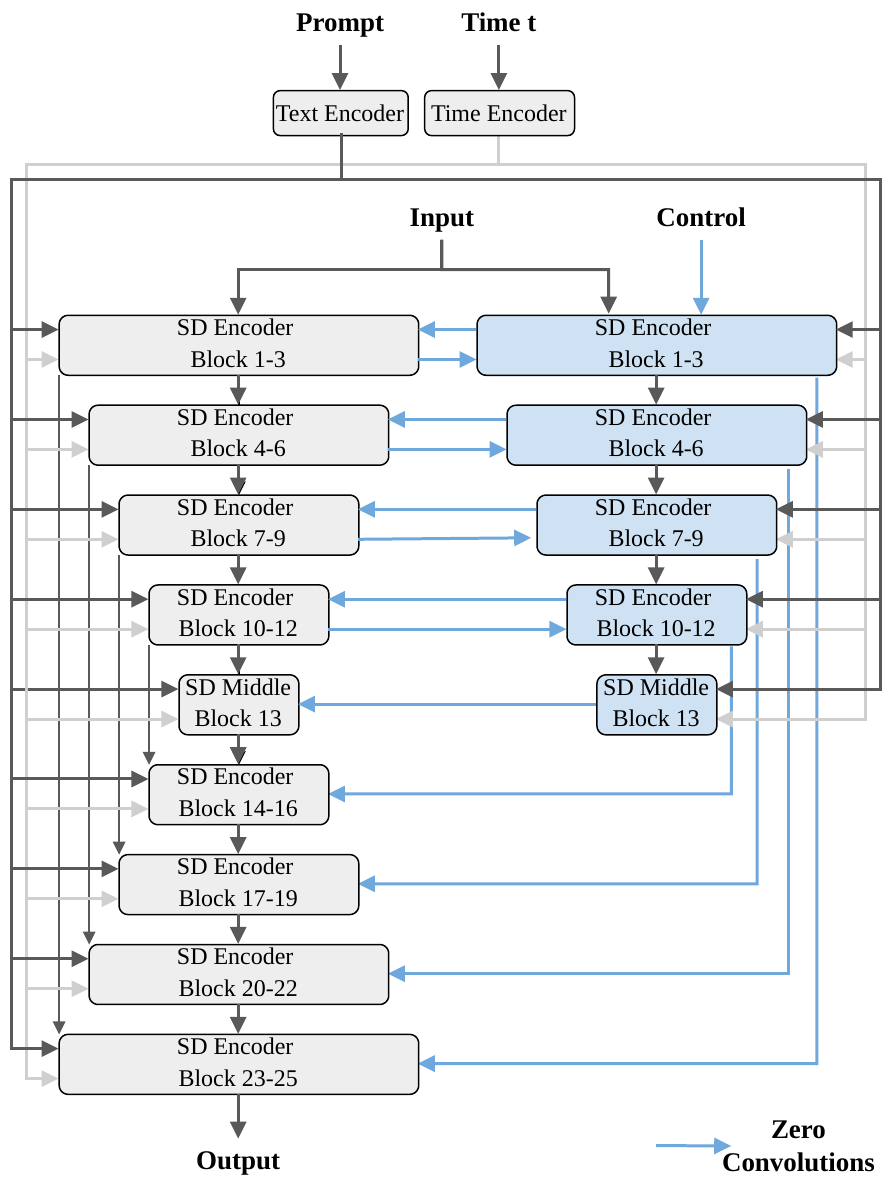}
    \caption{{\bf ControlNet-XS architecture} applied to the Stable Diffusion~\cite{Rombach2022_LDM} U-Net. The U-Net generation process is shown to the left. The U-Net encoder consists of three neural blocks per resolution, followed by a middle block and a decoder with corresponding architecture. The encoder and decoder are connected by the common U-Net skip-connections. ControlNet-XS mirrors the structure of the encoder but with significantly less parameters. Both encoders process the image signal, a text conditioning and a timestep embedding.  ControlNet-XS additionally receives a control signal. The intermediate feature maps of the generative encoder are communicated from each block of the generative encoder to ControlNet-XS. The connections from ControlNet-XS to the generative U-Net provide an additive correction to all intermediate feature maps. All connections between both networks contain zero-convolutions to ease training. Please see details about the connection of a generative block and a control block in the main article Figure 2c.}
    \label{fig:architecture}
\end{figure}

In \cref{fig:architecture}, we illustrate the interaction between the generative model Stable Diffusion (left) and ControlNet-XS (right). As mentioned in the main text, we leave the general structure of the base encoder untouched but decrease the number of input- and output channels in all layers. Our ControlNet-XS has 20\% of the channel sizes of the generative encoder. Additionally, each block of the control model that receives a feedback from the base network has additional input channels to cover the concatenated feedback. More precisely, each neural block is comprised of several modules like convolutions or attentions. The channel sizes of a control module are 
\begin{align}
    C^i_{\text{Con},\text{in}} = \begin{cases}
        \max(\lceil r \cdot C^i_{\text{Gen}, \text{in}}  \rceil, 1) + C^{i-1}_{\text{Gen},\text{out}},  & \text{module receives feedback},  \\
        \max(\lceil r \cdot C^i_{\text{Gen}, \text{in}}  \rceil, 1),                                   & \text{otherwise},
    \end{cases}
\end{align}
with $C_{\text{Con},\text{in}}^i$, $C_{\text{Gen},\text{in}}^i$ being the input channel size of the $i$th network module of the control network (Con) and the generative network (Gen), respectively. $C_{\text{Gen}, \text{out}}$ is the output channel size of a generative module and $r\in[0, 1]$ is the ratio for the channel size. ControlNet-XS uses $r = 0.2$.

\section{Sensitivity Analysis}
\label{sec:supp_sensitivity}

\begin{figure}[htp!]
 \begin{center}
   \centering
   \captionsetup{type=figure}     
   \includegraphics[width=1\linewidth]{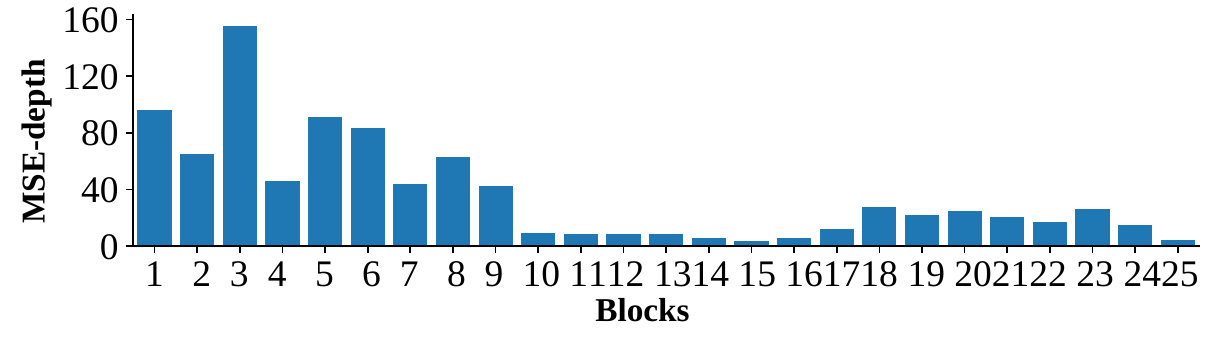}
   \caption{\textbf{Sensitivity analysis.} We show an MSE-depth error with respect to all $25$ blocks (x-axis) of the generative U-Net. The MSE-depth error is computed between the extracted depth maps of two generated images: i) a generated image for which all control connections were active, ii) a generated image were the control for one individual block was turned off. The plot is an average over $500$ images. Note that blocks $1-12$ belong to the encoder, block $13$ is the middle block and blocks $14-25$ belong to the decoder. We see that the 3rd block contributes most to the control, followed by the 1st, 5th and 6th block. In general, the blocks in the encoder appear to be far more essential for controlling than the remaining blocks of the decoder.
   \vspace{-0.5cm} 
   }
   \label{fig:sens_analysis}
\end{center} 
\end{figure}

In the following analysis, we want to understand by how much each individual block of the generative U-Net is affected by the control network. The study is shown in \cref{fig:sens_analysis}. We see that certain blocks are affected more than others. In particular, blocks in the encoder are affected considerably more than blocks in the decoder. We conjecture that this is the reason why our Type C architecture 
(see Figure 3 (d))
with a mirrored generative decoder does not lead to a clear improvement in performance 
(see Table 1, \cref{tab:arch_type_edges}).

\section{Training Details}
\label{sec:supp_training_details}

To control Stable Diffusion~\cite{Rombach2022_LDM}, we have trained four ControlNet-XS models, one for each type of guidance: Canny-edges, depth maps, semantic maps and human poses. We have also trained two ControlNet-XS models for Stable Diffusion XL~\cite{Podell2023_SDXL} with Canny-edge and depth map guidance. As training data, we used one million images from the LAION Aesthetics dataset~\cite{Schuhmann2022_LaionAE} for depth and edge guidance. For human poses, we filter out 160K images which show people and use this subset for training. For semantic maps, we use the COCO-Stuff~\cite{Caesar2018_COCO-STUFF} training dataset with 118K images and the corresponding semantic maps. For edge guidance, we extracted edges using Canny edge detection~\cite{canny1986computational} with random thresholds. For depth control, we approximated the depths using the MiDaS~\cite{Ranftl2020_Midas} approach. Human Pose guidance uses OpenPose~\cite{Cao2017_OpenPose} to extract the human keypoints. In \cref{tab:training_details}, we summarise the training setting for all models.

\begin{table*}
\centering
\small
\caption{\textbf{Training details} for ControlNet-XS, and for different controls applied to Stable Diffusion~\cite{Rombach2022_LDM} (SD) and Stable Diffusion XL~\cite{Podell2023_SDXL} (SD XL). The size of ControlNet-XS (CN-XS) does not have to increase in correspondence with the size of the controlled generative model. 'lr' is the learning rate used during training.}
\begin{tabular}{llcccc}
    \hline
    Condition & Control Model & Generative Model & Training Hours [A100] & lr & Batch Size\\
    \hline
    Edges           & CN-XS (55M) & SD (860M)      & $\sim200$   & 1e-5 & 16\\
    Depth Maps      & CN-XS (55M) & SD (860M)      & $\sim200$   & 1e-5 & 16\\
    Human Poses     & CN-XS (55M) & SD (860M)      & $\sim200$   & 1e-5 & 16\\
    Semantic Maps   & CN-XS (55M) & SD (860M)      & $\sim200$   & 1e-5 & 32\\
    Edges           & CN-XS (20M) & SD XL (2.6B)   & $\sim250$   & 1e-4 & 40\\
    Depth Maps      & CN-XS (20M) & SD XL (2.6B)   & $\sim250$   & 1e-4 & 40\\
    \hline
\end{tabular}
\label{tab:training_details}
\end{table*}

\section{Quantitative Results for Edge Control}
\label{sec:supp_quantitative_edges}
In \cref{tab:arch_type_edges}, we conduct an ablation study for three types of architectures of with our proposed modification 
(Figure 3 (b-d))
 and ControlNet for edge control. Please note that a similar study was done for depth control in the main article (Table 1). The HDD score is scaled by $10^{-1}$. As we concluded for depth control, we can confirm that Type B is on average the best architecture choice for edge control. In \cref{tab:quantative_analysis_edges}, we evaluate the effect that model size has on the performance of ControlNet-XS with 491M, 55M, 11.7M and 1.7M parameters, respectively. We also compare our models to the standard ControlNet~\cite{Zhang2023_ControlNet} for edge control to emphasise the effect our proposed bidirectional communication. 
Our best model with 55M parameters performs best in terms of quality (FID) and control (LPIPS and HDD). When we continue to decrease the size of the control network, we observe a decrease in performance as the controlled generation approaches the uncontrolled generation of the generative stable diffusion base. Without the bidirectional communication mechanism, ControlNet with 361M parameters performs worse in all metrics, apart from CLIP-Aesthetic. Even the smallest ControlNet-XS variant with 1.7M parameters, outperforms the ControlNet baseline. Overall, these results follow the same trend as observed for depth control in the main article (see Table 2,3).

\begin{table}
\centering
\caption{{\bf Ablation study for four different architectures} illustrated in 
Figure 3 in the main article
with edge control. We see that with additional, immediate corrective connections in Type B and Type C, the performance considerably increases for all metrics. We choose Type B as our ControlNet-XS architecture, since it performs on a par, on average, with Type C but has fewer parameters. Overall, this follows the same trend as observed for depth control in the main article (see. Table 1).}
\begin{tabular}{lccccc}
    \hline
     &  Both & \multicolumn{2}{c}{Control} & \multicolumn{2}{c}{Quality} \\
    \multicolumn{1}{c}{Method} & FID $\downarrow$ & HDD $\downarrow$ & 
             LPIPS $\downarrow$ & CLIP-Sc $\uparrow$ & CLIP-Ae $\uparrow$  \\
    \hline \hline
    CN (361M) 
              & \small \textcolor{black}{21.18}         
              & \small \textcolor{black}{18.52}         
              & \small \textcolor{black}{0.544}         
              & \small \textcolor{black}{29.01}         
              & \small \textbf{6.17}            
              \\

    Our Type A (53M)   & \small \textcolor{black}{17.40}           
                    & \small \textcolor{black}{15.46}           
                    & \small \textcolor{black}{0.452}           
                    &  \small \textcolor{black}{29.04}           
                    &  \small \textcolor{black}{5.85}           
                    \\
    
    Our Type B (55M)   & \small \textbf{15.13}        
                    & \small 15.22          
                    & \small 0.417          
                    &  \small \textbf{29.61}          
                    & \small 5.98           
                    \\
    
    Our Type C (117M)  & \small 15.34          
                    & \small \textbf{15.18}         
                    & \small \textbf{0.405}         
                    & \small 29.41           
                    & \small 5.99          
                    \\
    \hline
\end{tabular}

\label{tab:arch_type_edges}
\end{table}

\begin{table}
\centering

\caption{{\bf Quantitative evaluation} for edge control with respect to change in model size of ControlNet-XS, and the ControlNet baseline model. We observe that our best model, ControlNet-XS (CN-XS) with $55$M parameters, outperforms the baseline controlling network  ControlNet (CN)~\cite{Zhang2023_ControlNet} for every  metric besides the CLIP-Aesthetic score. Furthermore, for ControlNet-XS models with few parameters, \eg 1.7M, we notice that the fidelity of the control reduces, see FID, LPIPS and HDD scores. As discussed in the main text, models with fewer parameters approach the performance of the uncontrolled stable diffusion.
}
\begin{tabular}{lccccc}
    \hline
     &  Both & \multicolumn{2}{c}{Control} & \multicolumn{2}{c}{Quality} \\
    \multicolumn{1}{c}{Method} & FID $\downarrow$ & HDD $\downarrow$ & 
             LPIPS $\downarrow$ & CLIP-Sc $\uparrow$ & CLIP-Ae $\uparrow$  \\
    \hline \hline
    Stable Diffusion 
                     & \small \textcolor{gray}{22.69}           
                     & \small \textcolor{gray}{(18.87)}         
                     & \small \textcolor{gray}{(0.618)}         
                     & \small \textcolor{gray}{28.40}           
                     & \small \textcolor{gray}{6.16}            
                     \\
    \hline
    CN (361M) 
              & \small \textcolor{black}{21.18}         
              & \small \textcolor{black}{18.52}         
              & \small \textcolor{black}{0.544}         
              & \small \textcolor{black}{29.01}         
              & \small \textbf{6.17}            
              \\
    CN-XS (491M)  
                  & \small 15.90            
                  & \small 15.75            
                  & \small \textcolor{black}{0.429}         
                  & \small 29.48            
                  & \small \textcolor{black}{6.06}           
                  \\
    CN-XS (55M)   
                  & \small \textbf{15.13}           
                  & \small \textbf{15.22}           
                  & \small \textbf{0.417}           
                  & \small \textbf{29.61}           
                  & \small \textcolor{black}{5.98}          
                  \\
    CN-XS (11.7M) 
                  & \small \textcolor{black}{16.56}         
                  & \small \textcolor{black}{15.49}         
                  & \small 0.474            
                  & \small \textcolor{black}{29.10}         
                  &\small  6.04         
                  \\
    CN-XS (1.7M)  
                  & \small \textcolor{black}{17.07}         
                  & \small \textcolor{black}{15.57}         
                  & \small \textcolor{black}{0.482}         
                  & \small \textcolor{black}{29.02}         
                  & \small 6.10         
                  \\
    \hline
\end{tabular}
\label{tab:quantative_analysis_edges}
\end{table}

\section{Additional Qualitative Results}
\label{sec:supp_qualitative_results}
\begin{figure*}
    \centering

    \begin{subfigure}[t]{0.24\textwidth}
        \centering
        \includegraphics[width=.99\textwidth]{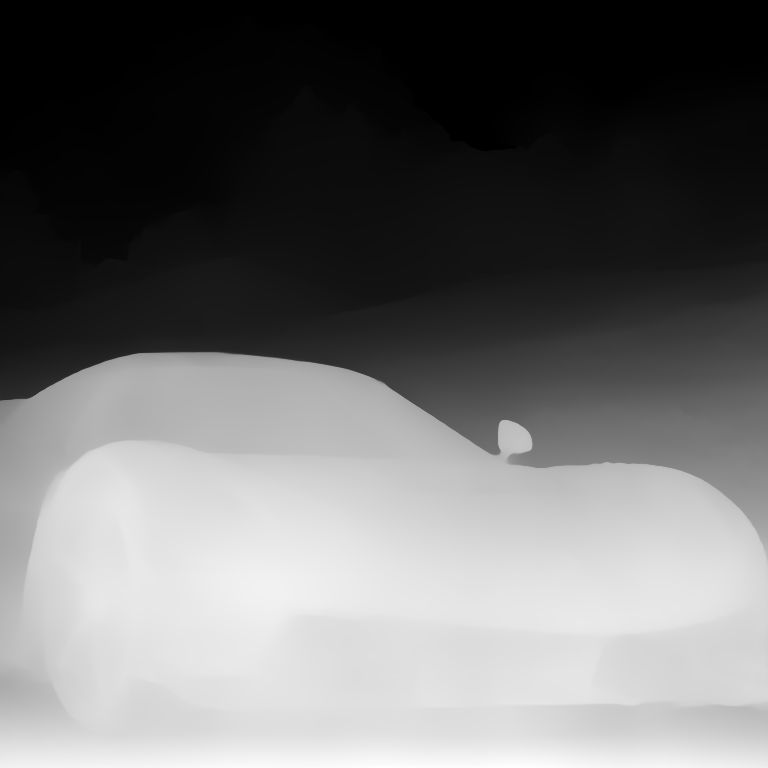}
        \caption{Control}
    \end{subfigure}
    \hfill
    \begin{subfigure}[t]{0.24\textwidth}
        \centering
        \includegraphics[width=.99\textwidth]{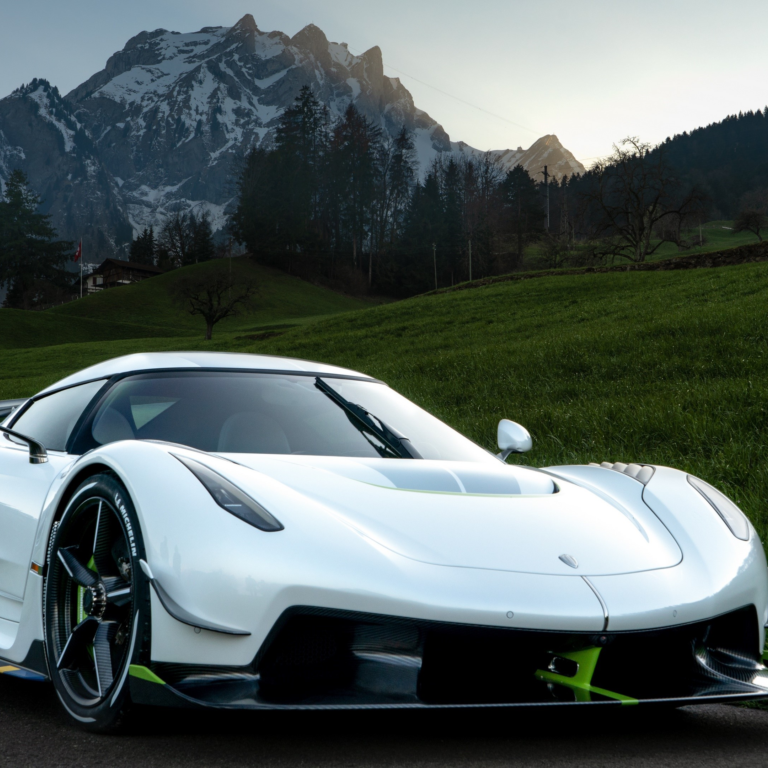}
        \caption{Original Image}
    \end{subfigure}
    \hfill
    \begin{subfigure}[t]{0.24\textwidth}
        \centering
        \includegraphics[width=.99\textwidth]{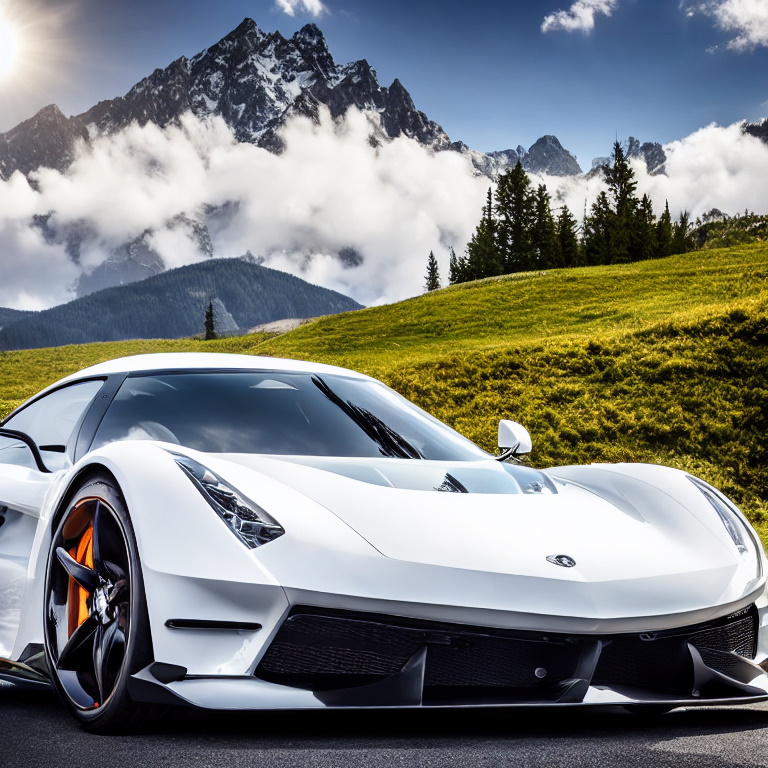}
        \caption{photo of white sports car, mountains, summer, award winning image, photorealistic, 4k}
    \end{subfigure}
    \hfill
    \begin{subfigure}[t]{0.24\textwidth}
        \centering
        \includegraphics[width=.99\textwidth]{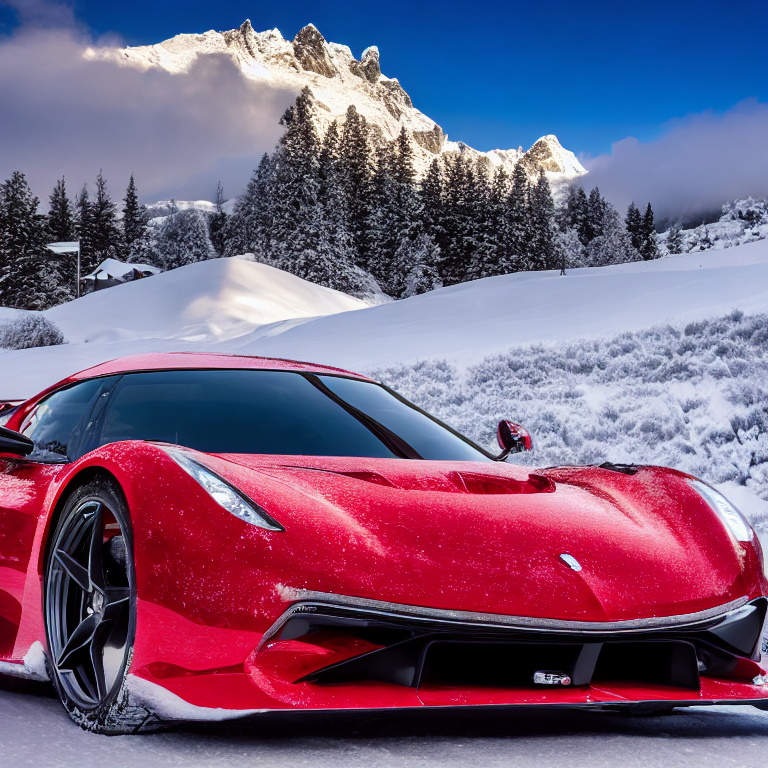}
        \caption{photo of red sports car, mountains, winter, award winning image, photorealistic, 4k, snowing}
    \end{subfigure}
    
    \hfill
    \begin{subfigure}[t]{0.24\textwidth}
        \centering
        \includegraphics[width=.99\textwidth]{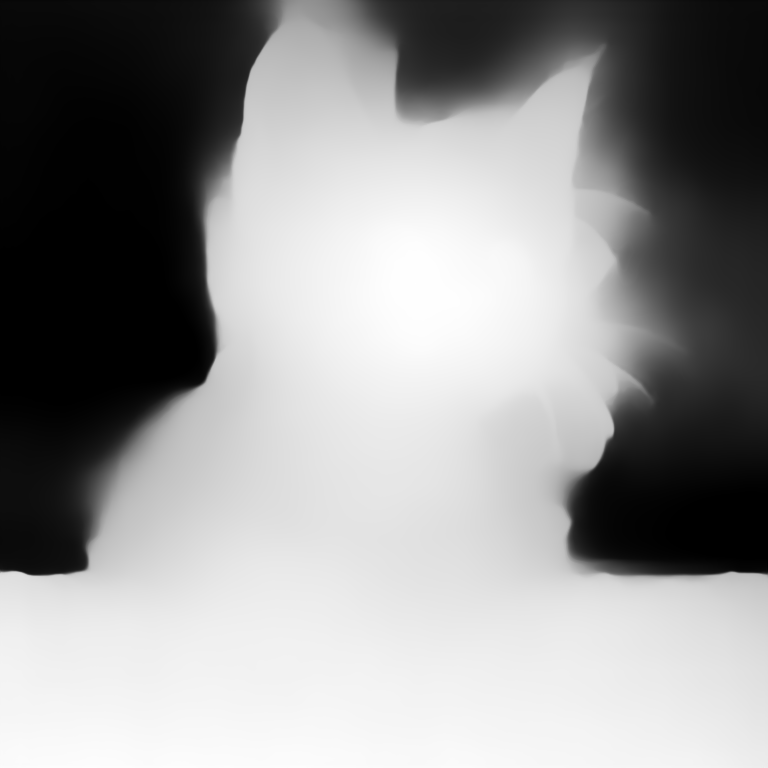}
        \caption{Control}
    \end{subfigure}
    \hfill
    \begin{subfigure}[t]{0.24\textwidth}
        \centering
        \includegraphics[width=.99\textwidth]{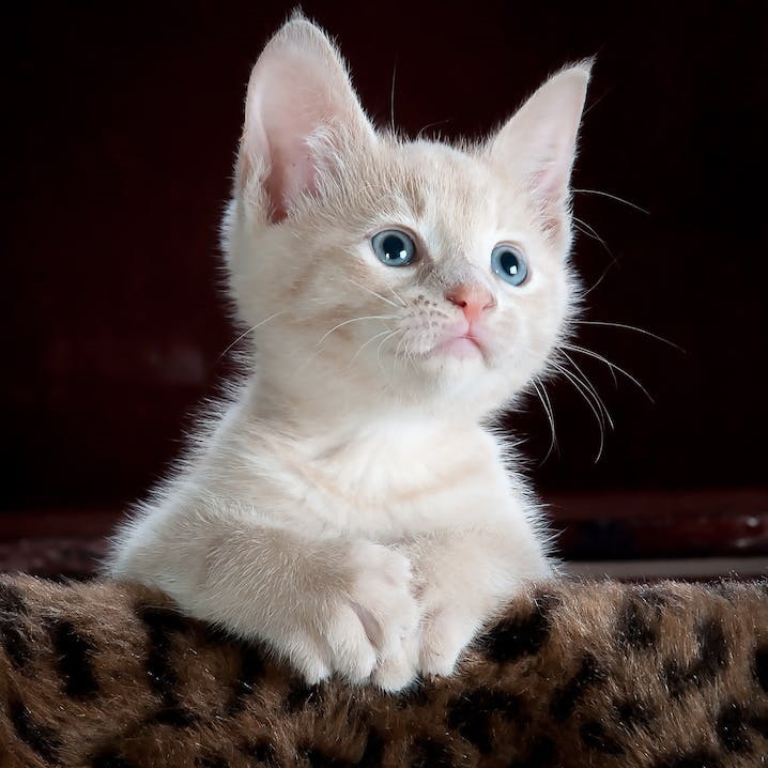}
        \caption{Original Image}
    \end{subfigure}
    \hfill
    \begin{subfigure}[t]{0.24\textwidth}
        \centering
        \includegraphics[width=.99\textwidth]{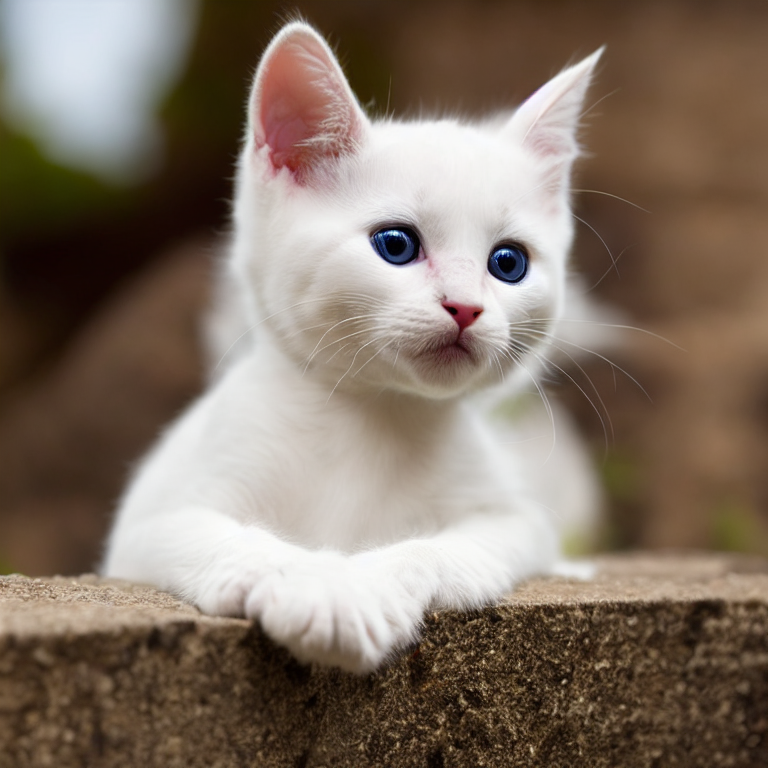}
        \caption{high resolution image of a cute white kitten, high quality, award winning image}
    \end{subfigure}
    \hfill
    \begin{subfigure}[t]{0.24\textwidth}
        \centering
        \includegraphics[width=.99\textwidth]{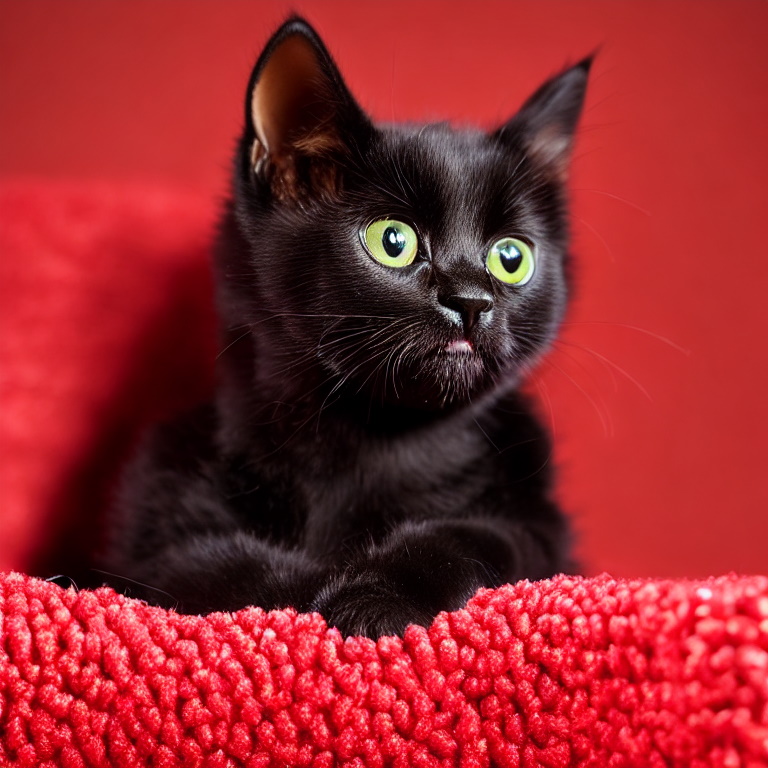}
        \caption{high resolution image of a cute black kitten, high quality, award winning image}
    \end{subfigure}
    \hfill

    \begin{subfigure}[t]{0.24\textwidth}
        \centering
        \includegraphics[width=.99\textwidth]{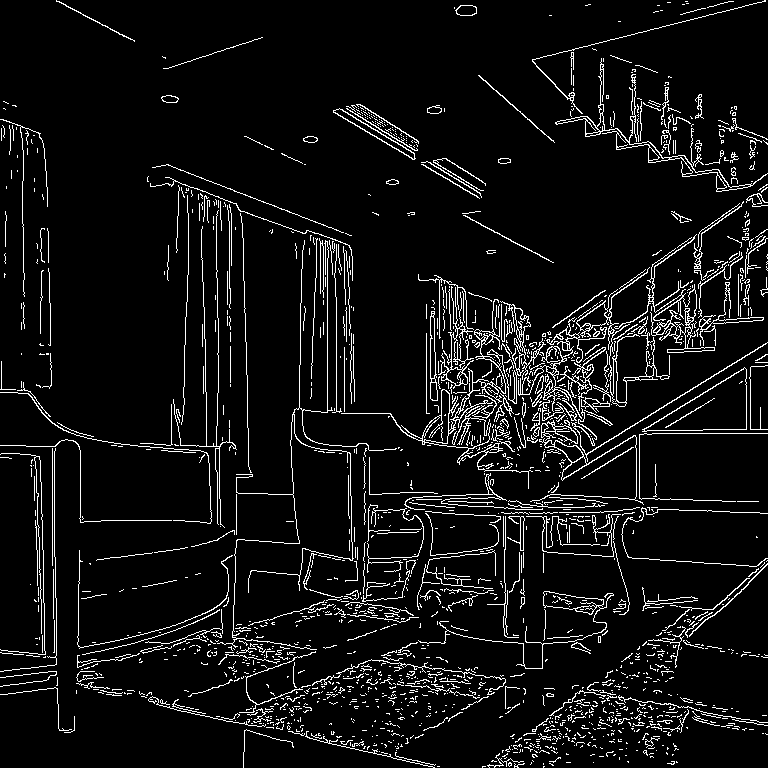}
        \caption{Control}
    \end{subfigure}
    \hfill
    \begin{subfigure}[t]{0.24\textwidth}
        \centering
        \includegraphics[width=.99\textwidth]{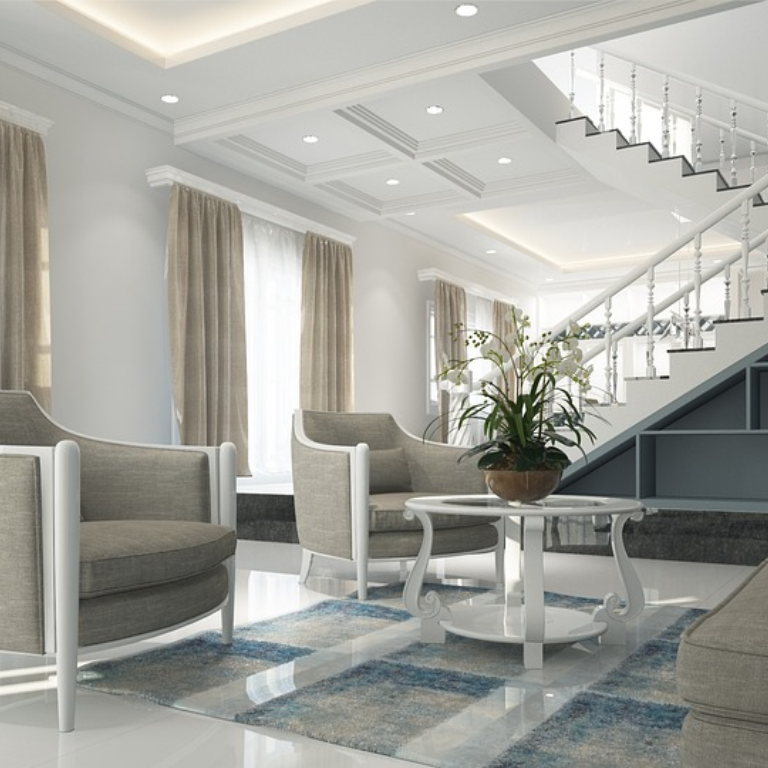}
        \caption{Original Image}
    \end{subfigure}
    \hfill
    \begin{subfigure}[t]{0.24\textwidth}
        \centering
        \includegraphics[width=.99\textwidth]{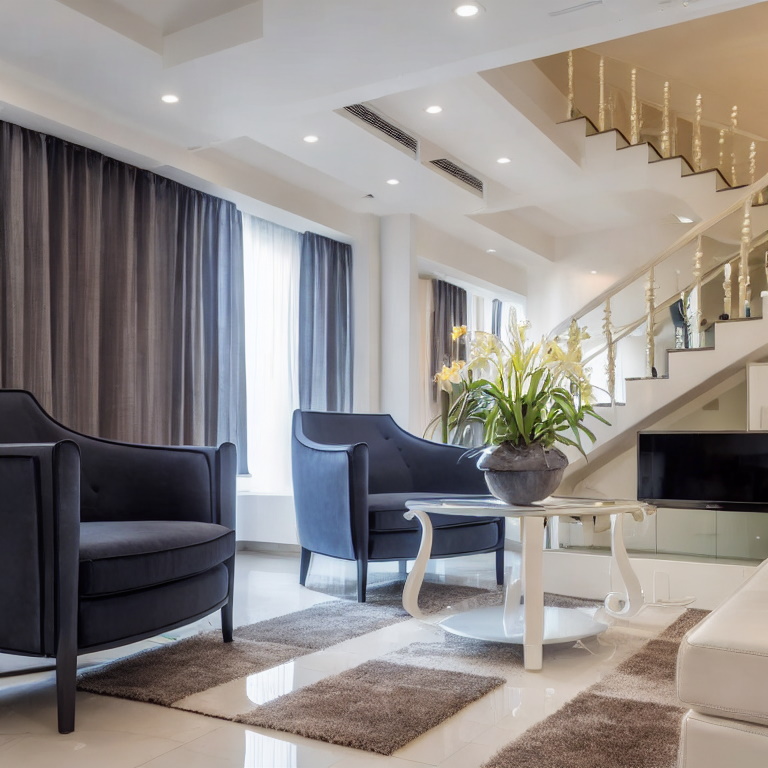}
        \caption{cinematic, luxury apartment, colourful, highly detailed}
    \end{subfigure}
    \hfill
    \begin{subfigure}[t]{0.24\textwidth}
        \centering
        \includegraphics[width=.99\textwidth]{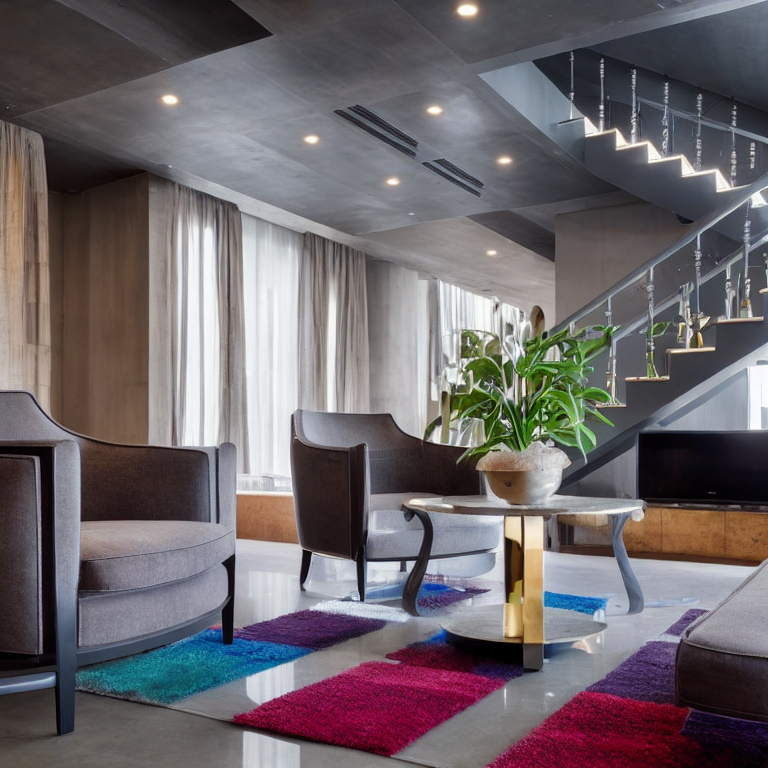}
        \caption{cinematic, cyberpunk apartment out of steel and concrete, colourful, highly detailed}
    \end{subfigure}
    \hfill

    \begin{subfigure}[t]{0.24\textwidth}
        \centering
        \includegraphics[width=.99\textwidth]{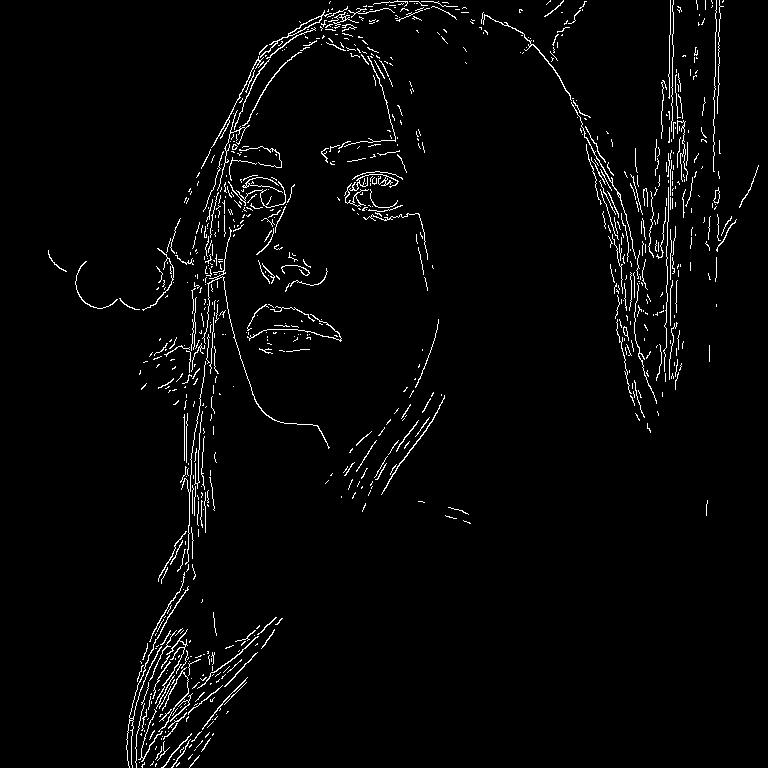}
        \caption{Control}
    \end{subfigure}
    \hfill
    \begin{subfigure}[t]{0.24\textwidth}
        \centering
        \includegraphics[width=.99\textwidth]{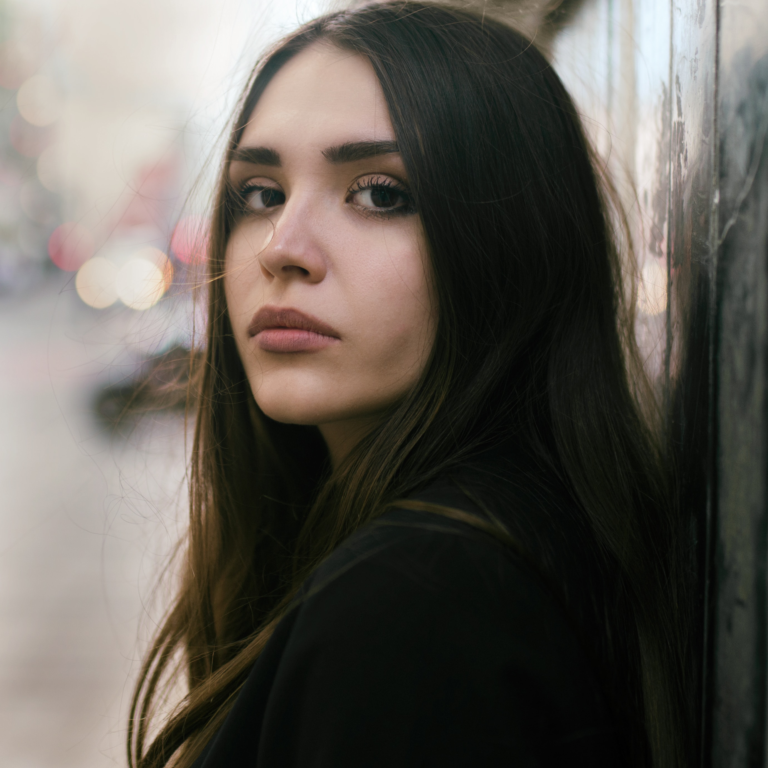}
        \caption{Original Image}
    \end{subfigure}
    \hfill
    \begin{subfigure}[t]{0.24\textwidth}
        \centering
        \includegraphics[width=.99\textwidth]{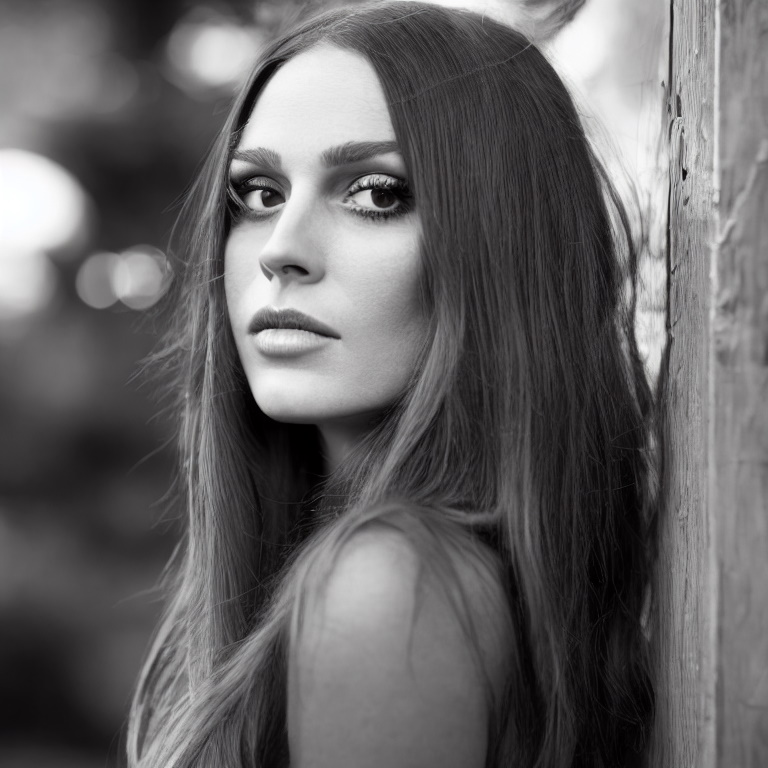}
        \caption{photo of a beautiful young woman, award winning picture, professional photo, black and white}
    \end{subfigure}
    \hfill
    \begin{subfigure}[t]{0.24\textwidth}
        \centering
        \includegraphics[width=.99\textwidth]{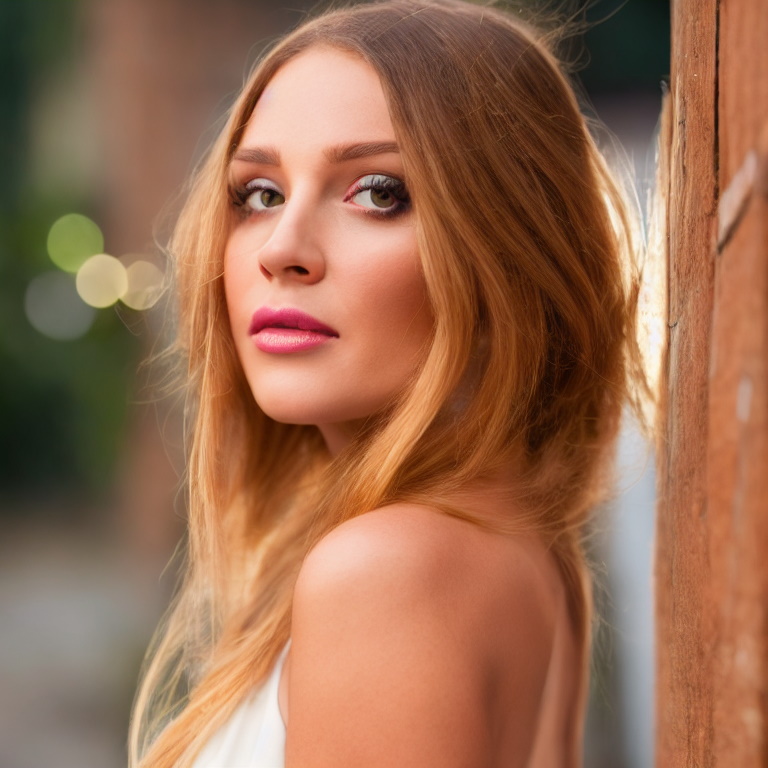}
        \caption{photo of a beautiful young woman, award winning picture, professional photo}
    \end{subfigure}
    \hfill

    \caption{Images generated by ControlNet-XS (55M) and Stable Diffusion 1.5~\cite{Rombach2022_LDM} as generative model with two different text-prompts. The generated images have the resolution of $768 \times 768$.}
    \label{fig:GenerationsSD1.5}
\end{figure*}

\begin{figure*}
    \centering

    \begin{subfigure}[t]{0.24\textwidth}
        \centering
        \includegraphics[width=.99\textwidth]{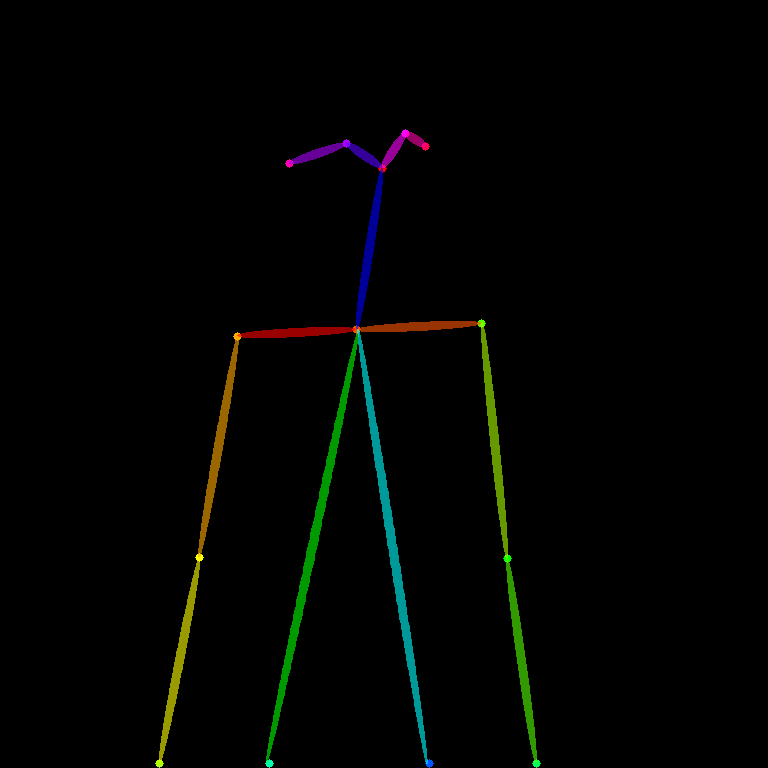}
        \caption{Control}
    \end{subfigure}
    \hfill
    \begin{subfigure}[t]{0.24\textwidth}
        \centering
        \includegraphics[width=.99\textwidth]{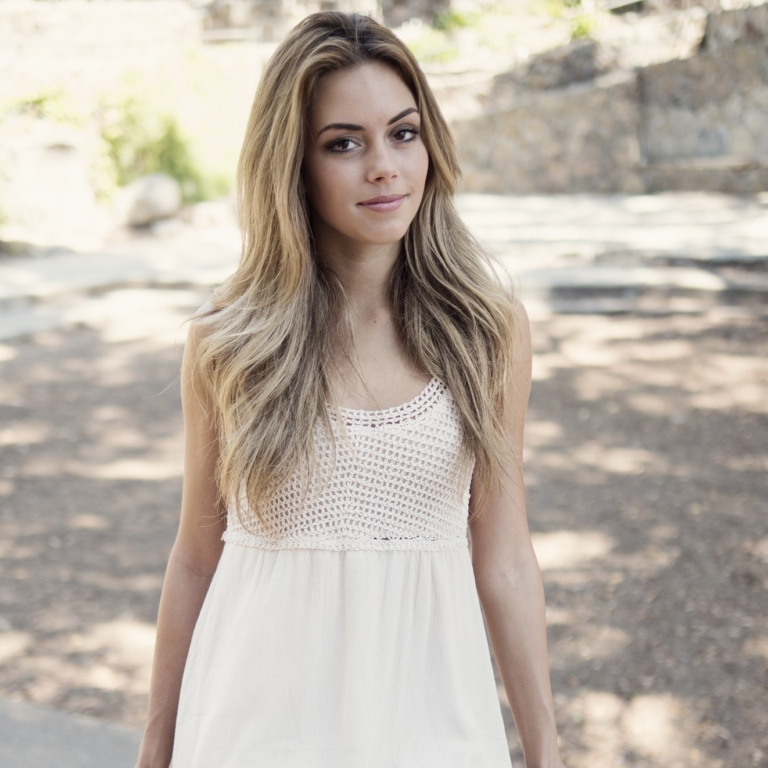}
        \caption{Original Image}
    \end{subfigure}
    \hfill
    \begin{subfigure}[t]{0.24\textwidth}
        \centering
        \includegraphics[width=.99\textwidth]{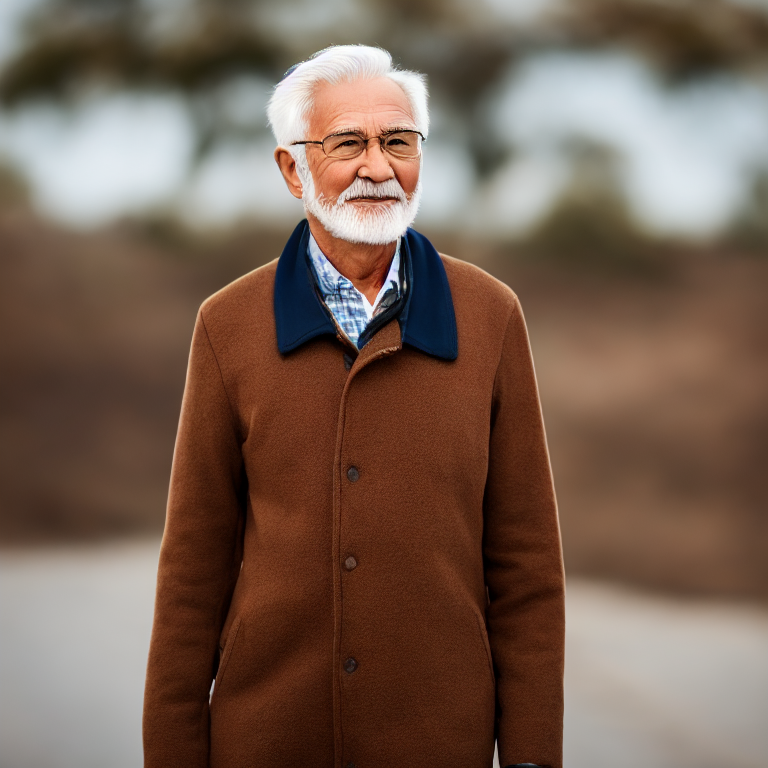}
        \caption{old man. High Quality image, High detail, beautiful, detailed eyes, short hair, blurry background}
    \end{subfigure}
    \hfill
    \begin{subfigure}[t]{0.24\textwidth}
        \centering
        \includegraphics[width=.99\textwidth]{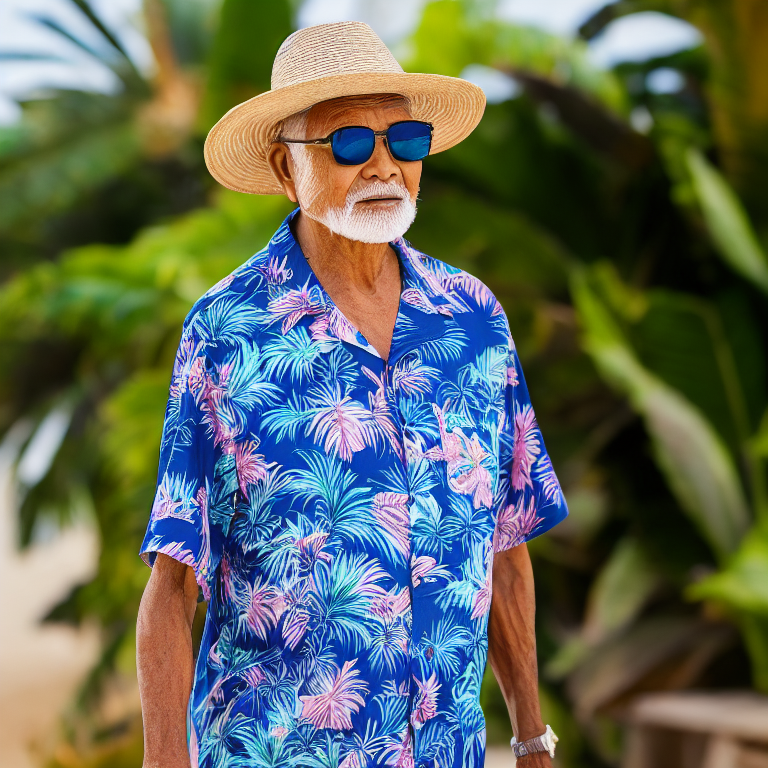}
        \caption{old man in a Hawaii shirt. High Quality image, High detail, beautiful, detailed eyes, short hair, blurry background}
    \end{subfigure}
    
    \hfill
    \begin{subfigure}[t]{0.24\textwidth}
        \centering
        \includegraphics[width=.99\textwidth]{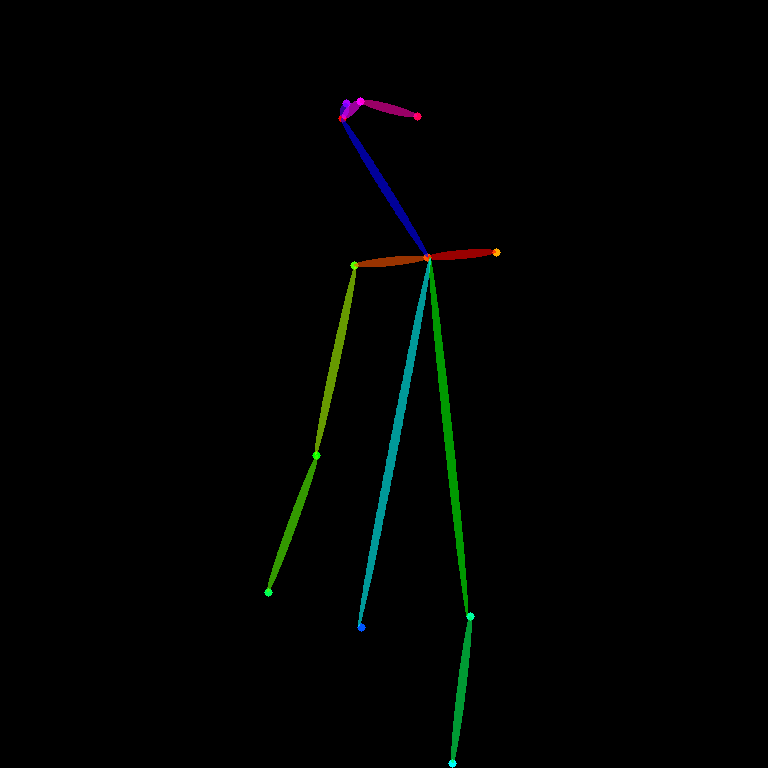}
        \caption{Control}
    \end{subfigure}
    \hfill
    \begin{subfigure}[t]{0.24\textwidth}
        \centering
        \includegraphics[width=.99\textwidth]{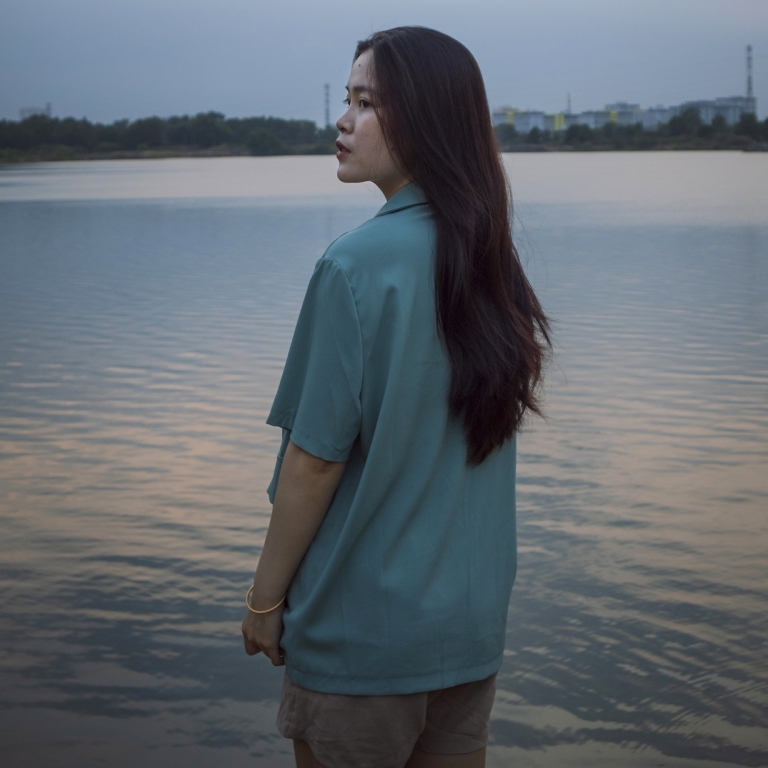}
        \caption{Original Image}
    \end{subfigure}
    \hfill
    \begin{subfigure}[t]{0.24\textwidth}
        \centering
        \includegraphics[width=.99\textwidth]{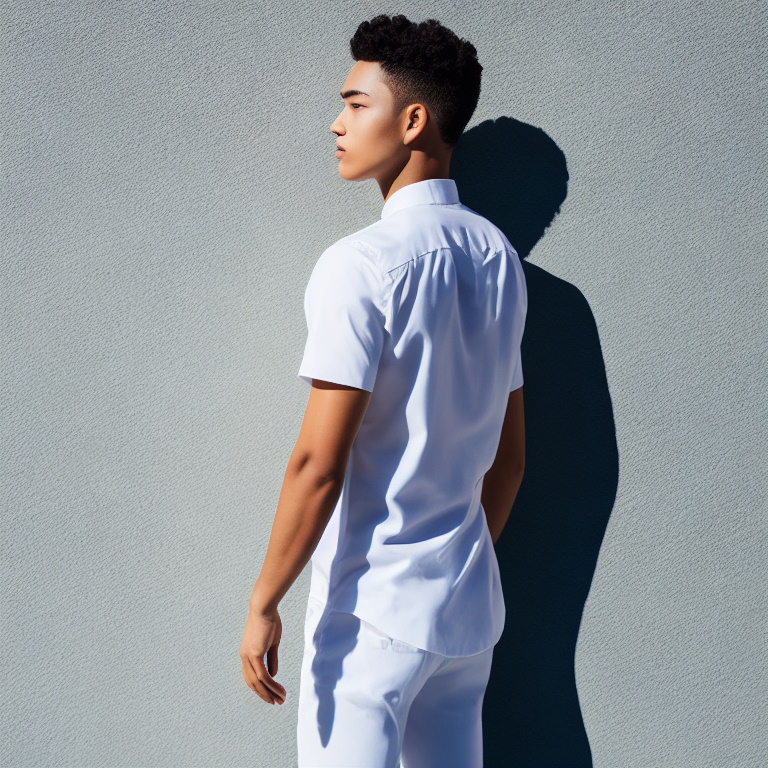}
        \caption{beautiful young man. High Quality, normal skin, High detail, obscure shadow, short hair, blurry background}
    \end{subfigure}
    \hfill
    \begin{subfigure}[t]{0.24\textwidth}
        \centering
        \includegraphics[width=.99\textwidth]{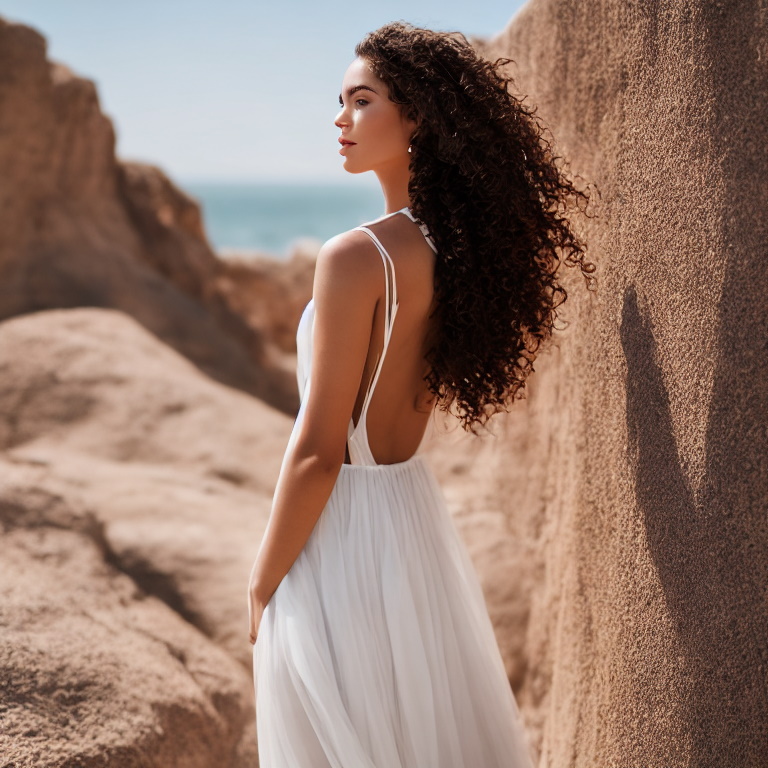}
        \caption{beautiful young woman. High Quality, High resolution, High detail, obscure shadow, long hair, blurry background}
    \end{subfigure}
    \hfill

    \begin{subfigure}[t]{0.24\textwidth}
        \centering
        \includegraphics[width=.99\textwidth]{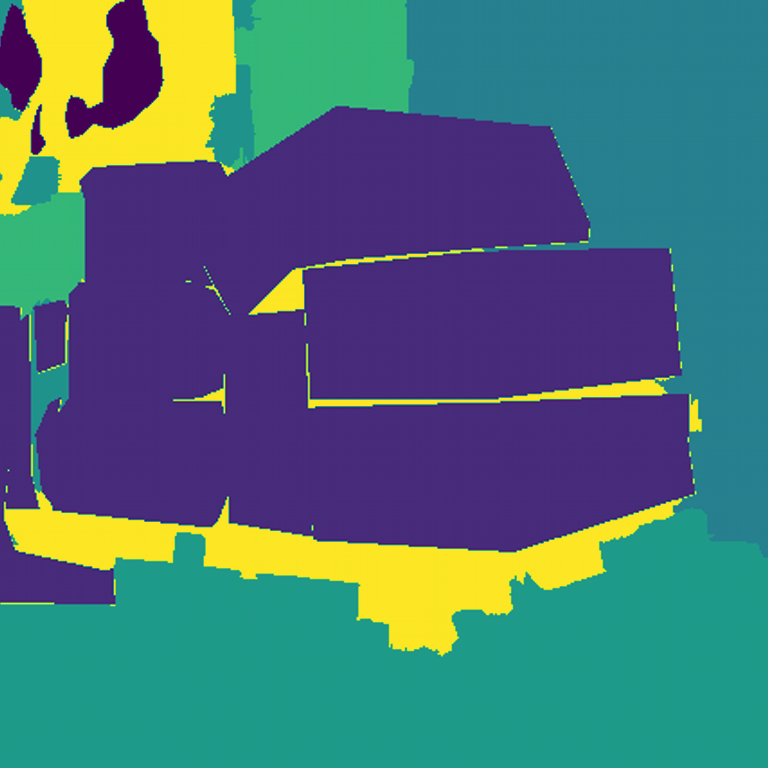}
        \caption{Control}
    \end{subfigure}
    \hfill
    \begin{subfigure}[t]{0.24\textwidth}
        \centering
        \includegraphics[width=.99\textwidth]{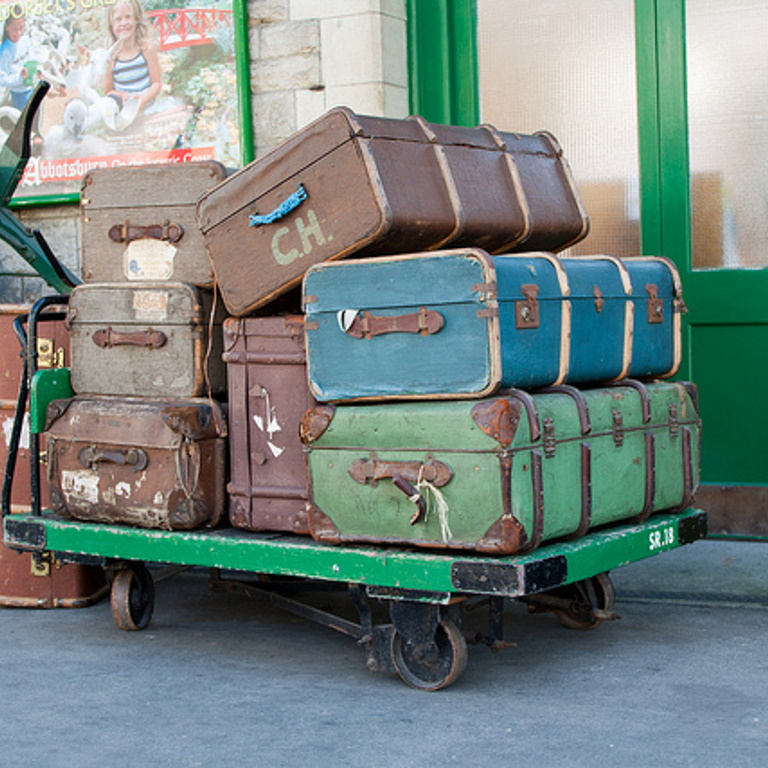}
        \caption{Original Image}
    \end{subfigure}
    \hfill
    \begin{subfigure}[t]{0.24\textwidth}
        \centering
        \includegraphics[width=.99\textwidth]{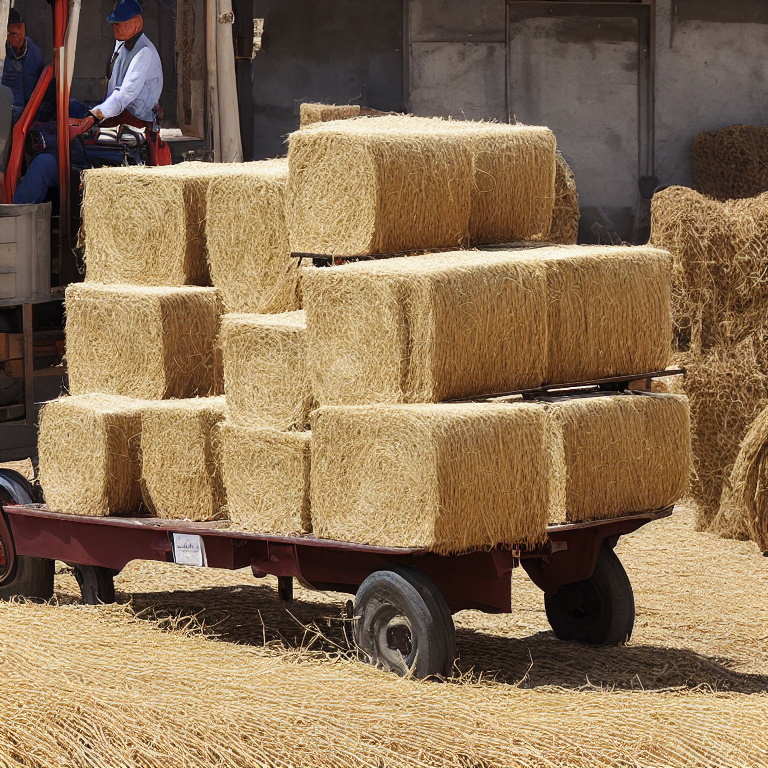}
        \caption{a cart with bales of hay on it. High Quality, High detail, obscure shadow, blurry background}
    \end{subfigure}
    \hfill
    \begin{subfigure}[t]{0.24\textwidth}
        \centering
        \includegraphics[width=.99\textwidth]{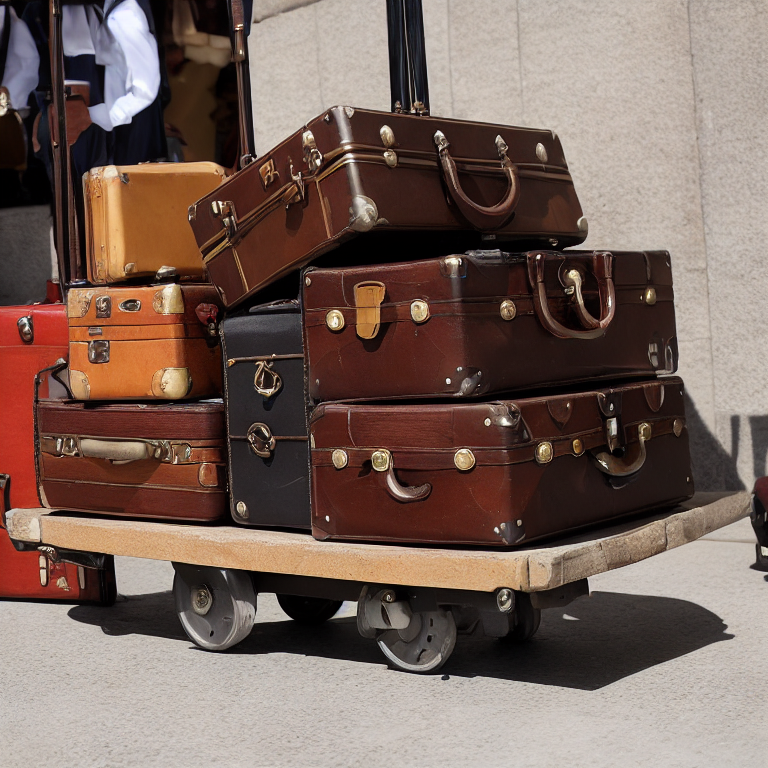}
        \caption{a cart with old suitcases on it. High Quality, High detail, obscure shadow, blurry background}
    \end{subfigure}
    \hfill

    \begin{subfigure}[t]{0.24\textwidth}
        \centering
        \includegraphics[width=.99\textwidth]{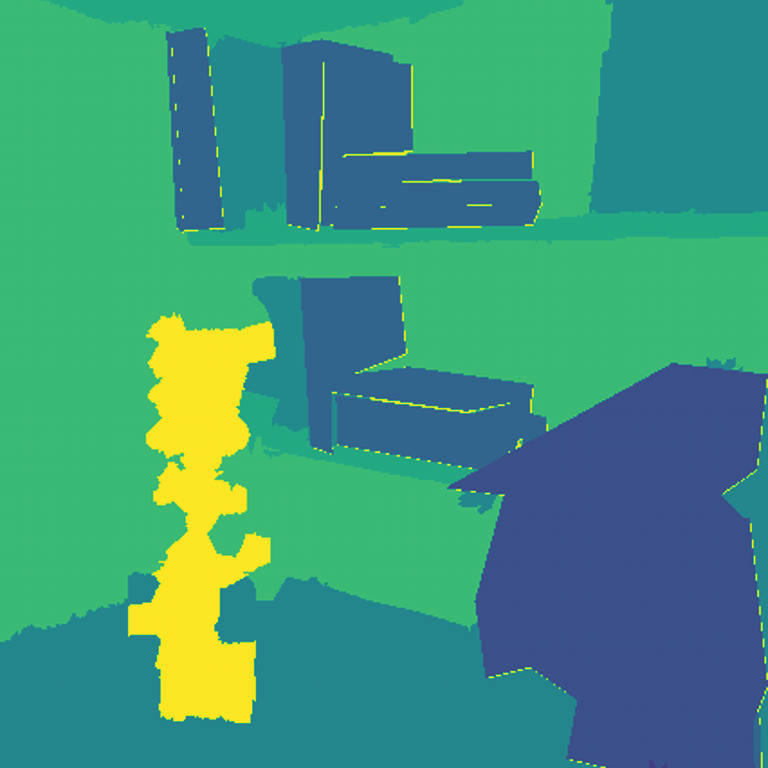}
        \caption{Control}
    \end{subfigure}
    \hfill
    \begin{subfigure}[t]{0.24\textwidth}
        \centering
        \includegraphics[width=.99\textwidth]{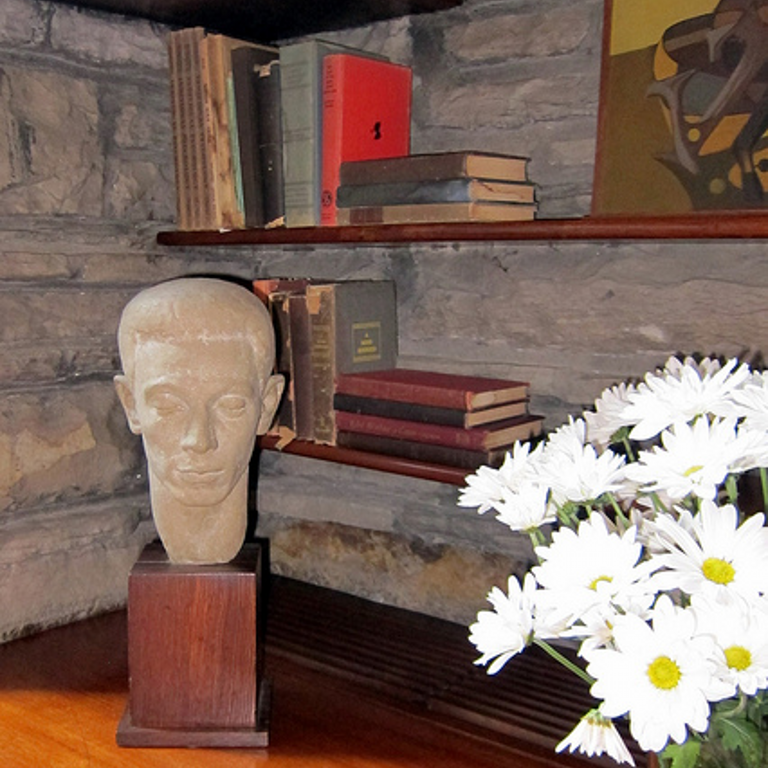}
        \caption{Original Image}
    \end{subfigure}
    \hfill
    \begin{subfigure}[t]{0.24\textwidth}
        \centering
        \includegraphics[width=.99\textwidth]{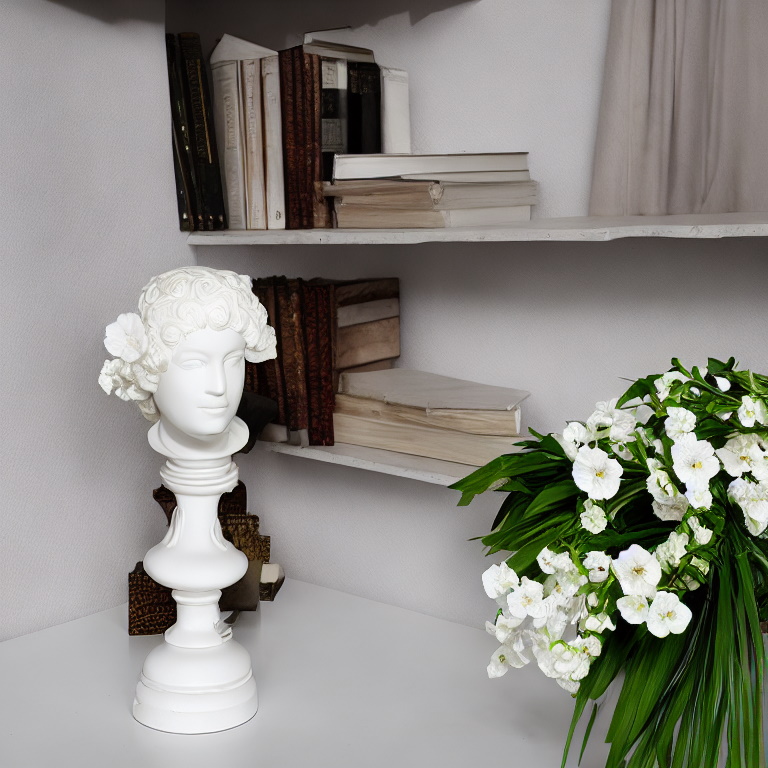}
        \caption{Still life with a bust and white flowers. High Quality, High detail}
    \end{subfigure}
    \hfill
    \begin{subfigure}[t]{0.24\textwidth}
        \centering
        \includegraphics[width=.99\textwidth]{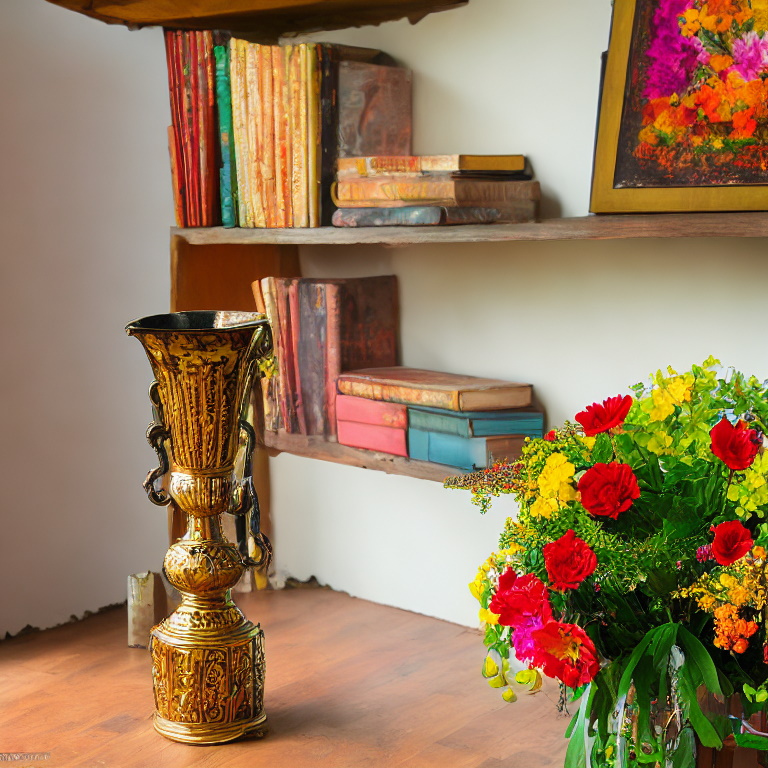}
        \caption{Still life with a vase and colorful flowers. High Quality, High detail}
    \end{subfigure}
    \hfill

    \caption{Images generated by ControlNet-XS (55M) and Stable Diffusion 1.5~\cite{Rombach2022_LDM} as generative model with two different text-prompts. The generated images have the resolution of $768 \times 768$.}
    \label{fig:GenerationsSD15_pose_seg}
\end{figure*}

\begin{figure*}
    \centering

    \begin{subfigure}[t]{0.24\textwidth}
        \centering
        \includegraphics[width=.99\textwidth]{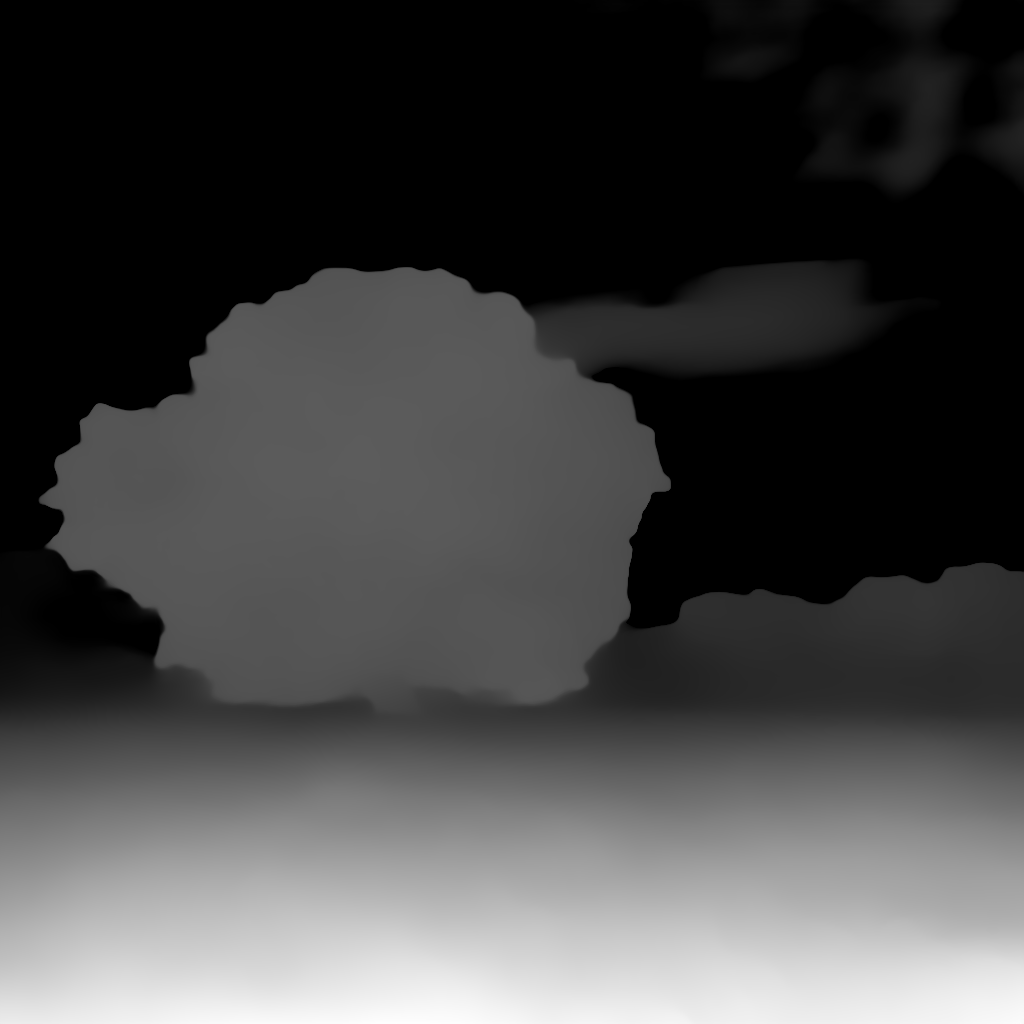}
        \caption{Control}
    \end{subfigure}
    \hfill
    \begin{subfigure}[t]{0.24\textwidth}
        \centering
        \includegraphics[width=.99\textwidth]{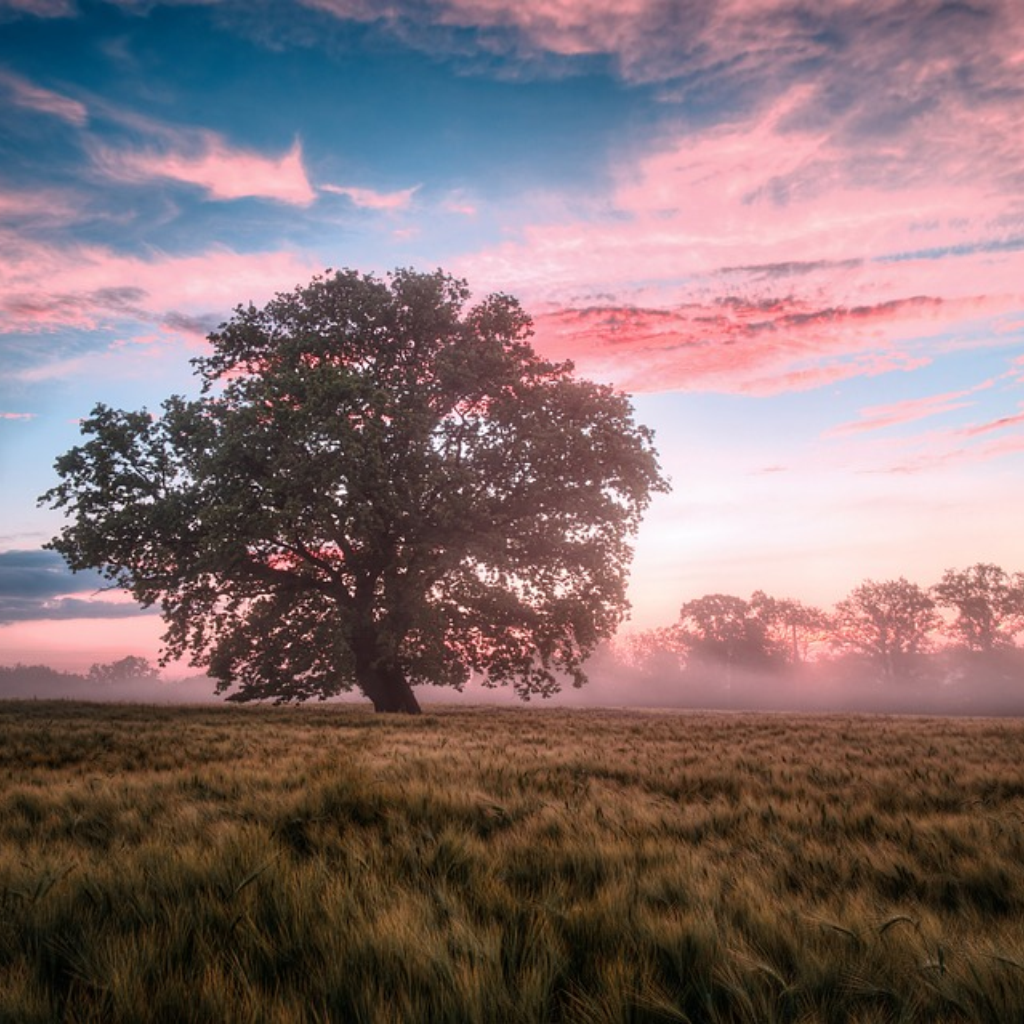}
        \caption{Original Image}
    \end{subfigure}
    \hfill
    \begin{subfigure}[t]{0.24\textwidth}
        \centering
        \includegraphics[width=.99\textwidth]{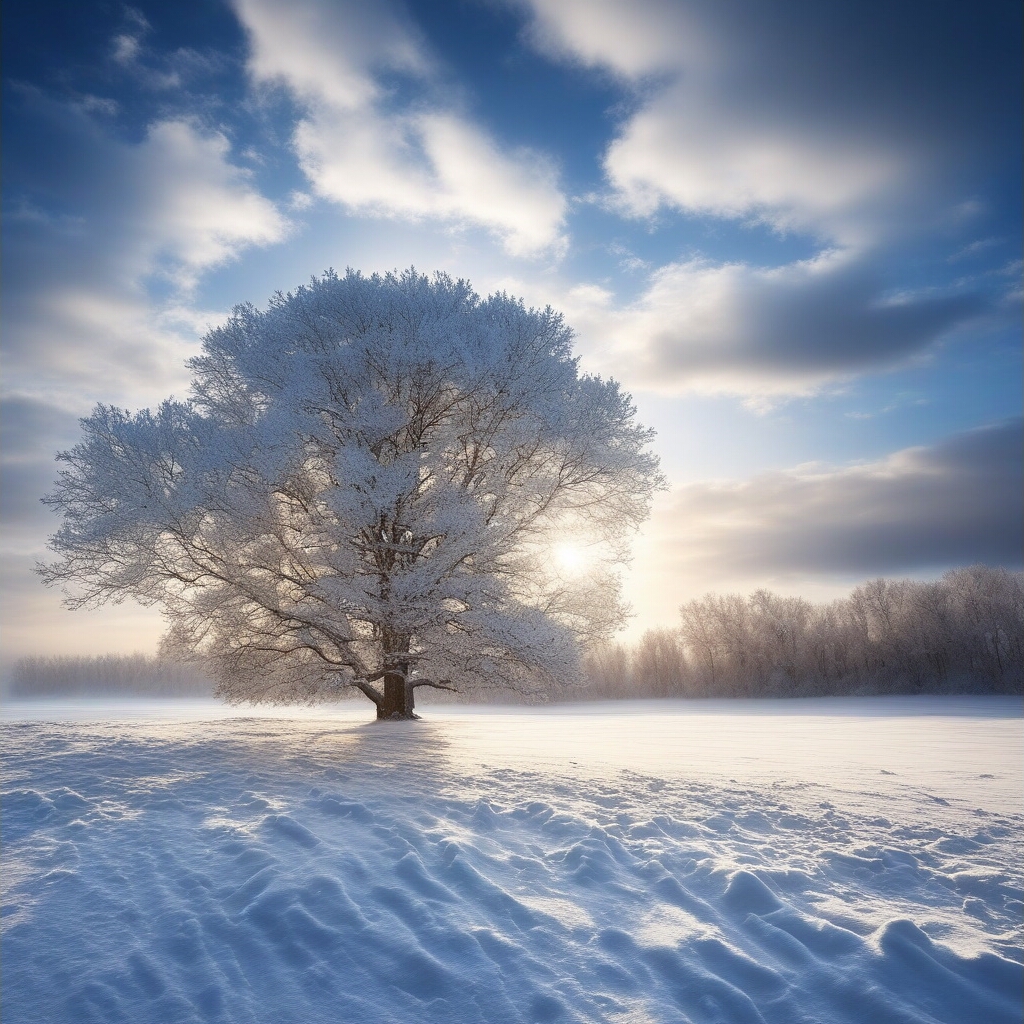}
        \caption{cinematic, winter, tree on a field, dramatic sky, highly detailed, photorealistic}
    \end{subfigure}
    \hfill
    \begin{subfigure}[t]{0.24\textwidth}
        \centering
        \includegraphics[width=.99\textwidth]{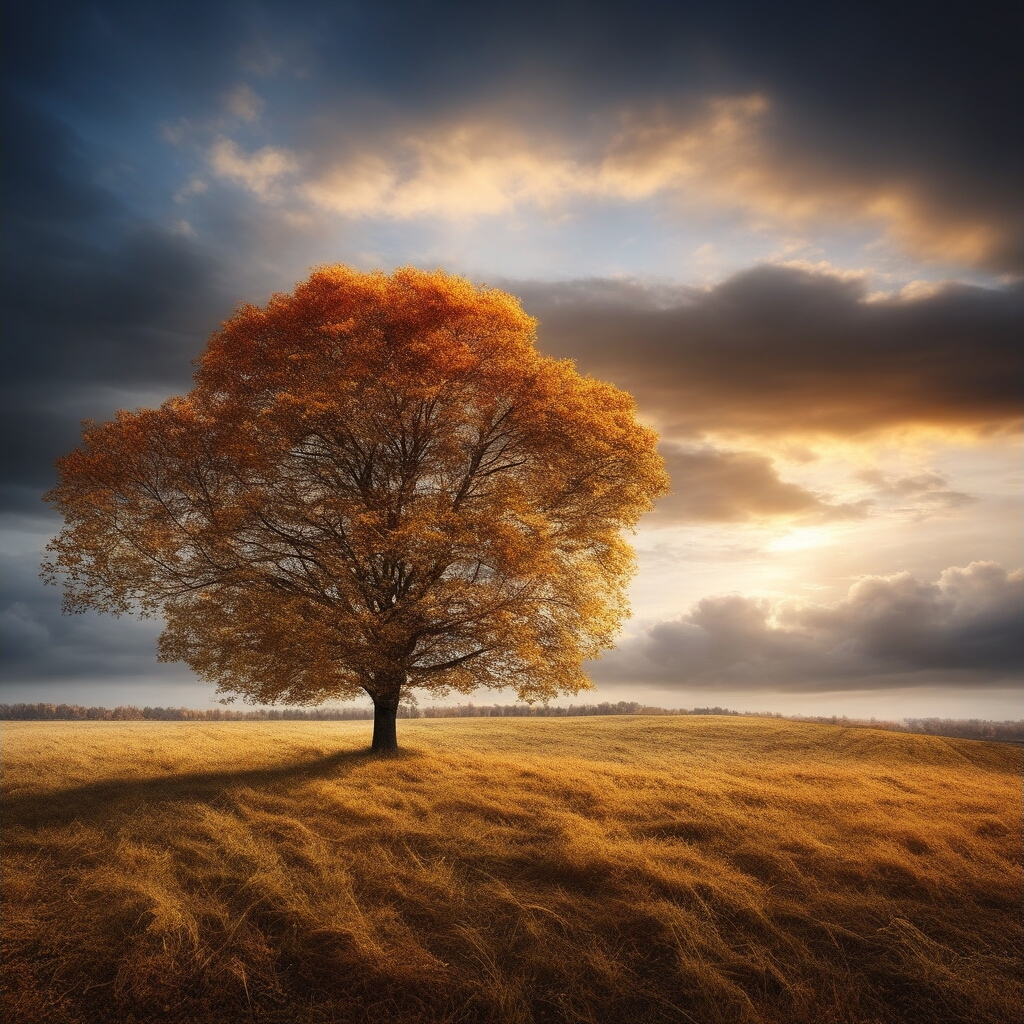}
        \caption{cinematic, autumn, tree on a field, dramatic sky, highly detailed, photorealistic}
    \end{subfigure}
    \hfill

    \begin{subfigure}[t]{0.24\textwidth}
        \centering
        \includegraphics[width=.99\textwidth]{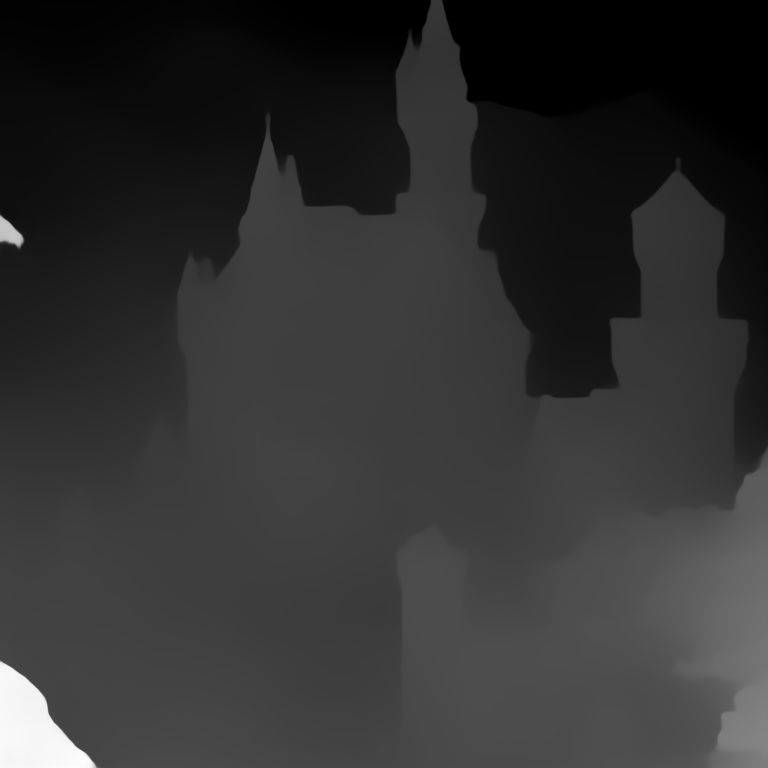}
        \caption{Control}
    \end{subfigure}
    \hfill
    \begin{subfigure}[t]{0.24\textwidth}
        \centering
        \includegraphics[width=.99\textwidth]{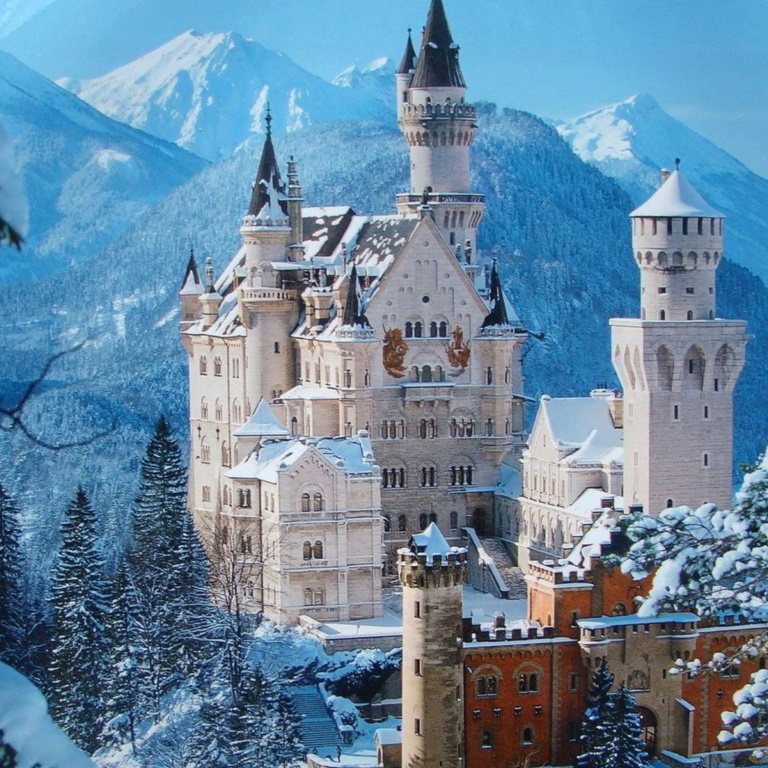}
        \caption{Original Image}
    \end{subfigure}
    \hfill
    \begin{subfigure}[t]{0.24\textwidth}
        \centering
        \includegraphics[width=.99\textwidth]{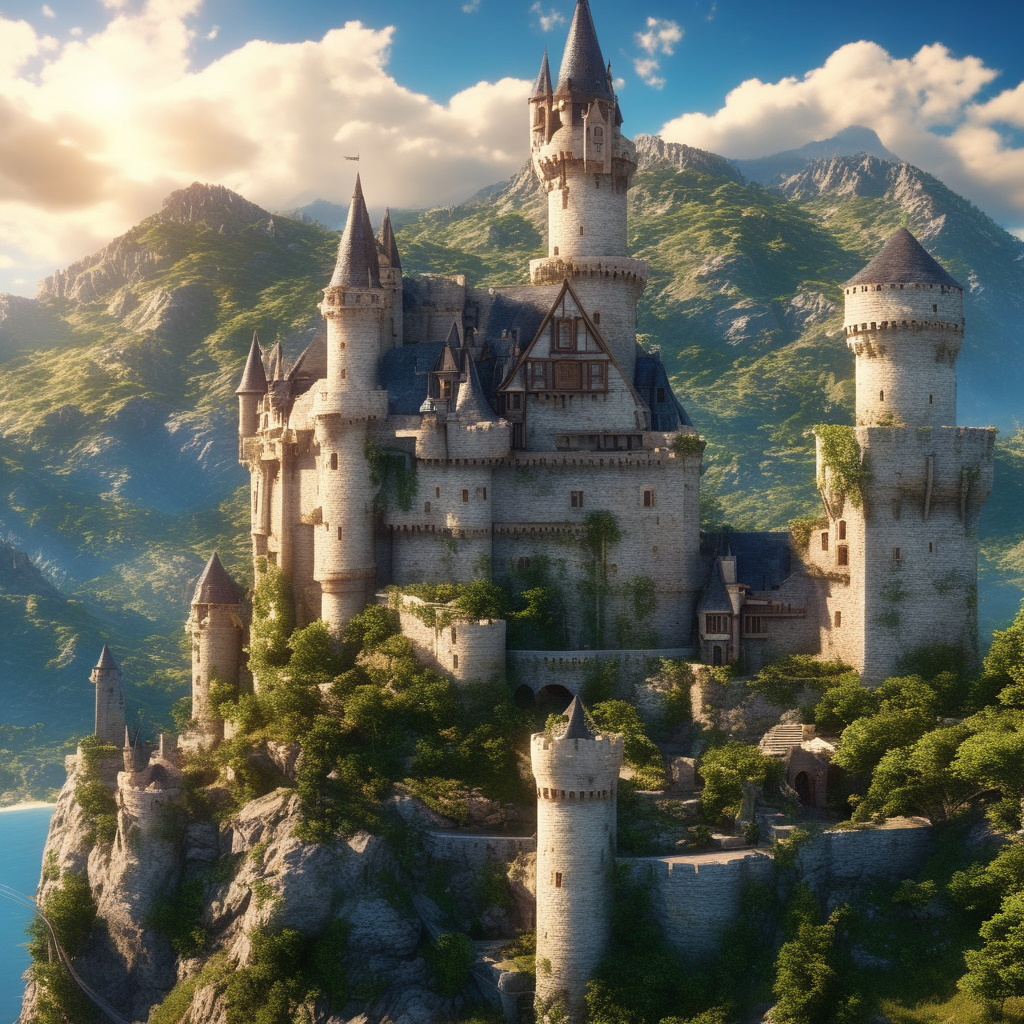}
        \caption{cinematic, highly detailed, castle, beautiful sky, summer, photorealistic, 4k}
    \end{subfigure}
    \hfill
    \begin{subfigure}[t]{0.24\textwidth}
        \centering
        \includegraphics[width=.99\textwidth]{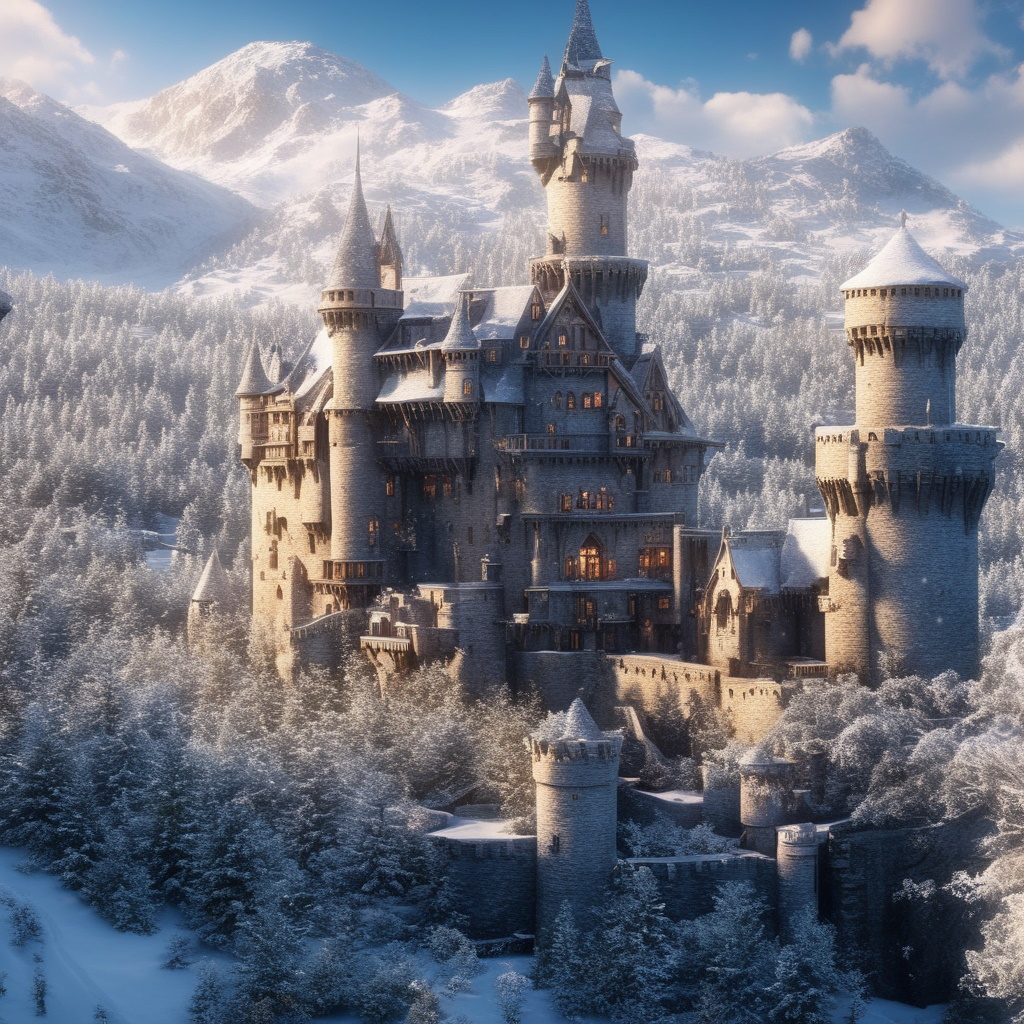}
        \caption{cinematic, highly detailed, snowy castle, beautiful sky, snowy winter, photorealistic, 4k}
    \end{subfigure}
    \hfill

    \begin{subfigure}[t]{0.24\textwidth}
        \centering
        \includegraphics[width=.99\textwidth]{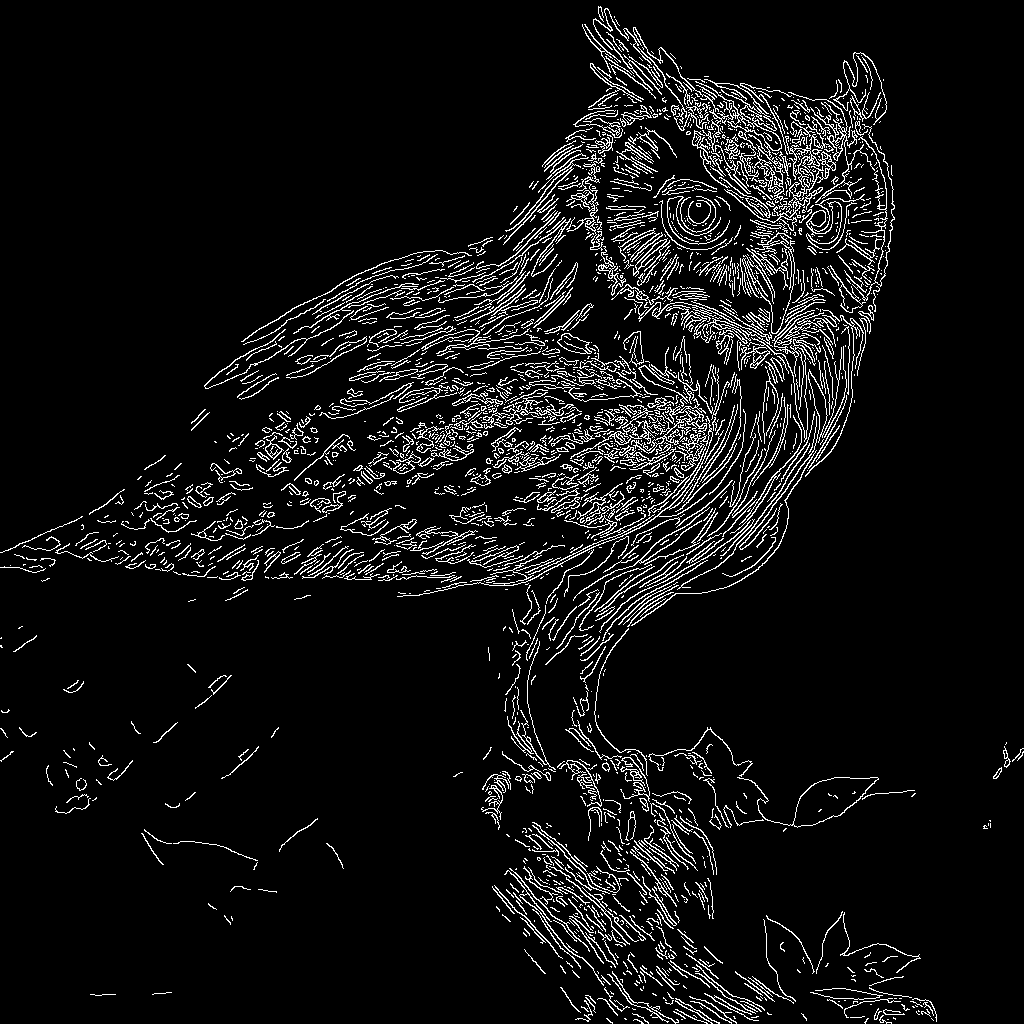}
        \caption{Control}
    \end{subfigure}
    \hfill
    \begin{subfigure}[t]{0.24\textwidth}
        \centering
        \includegraphics[width=.99\textwidth]{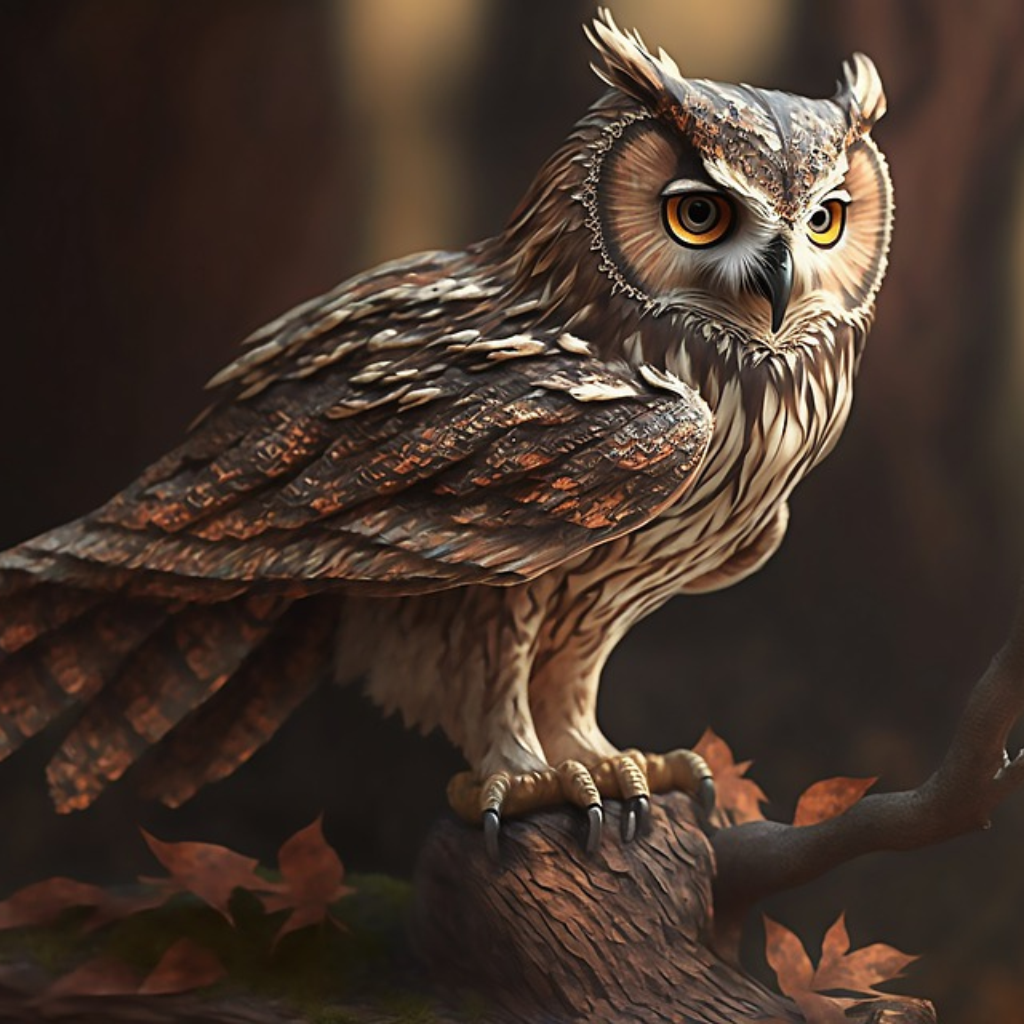}
        \caption{Original Image}
    \end{subfigure}
    \hfill
    \begin{subfigure}[t]{0.24\textwidth}
        \centering
        \includegraphics[width=.99\textwidth]{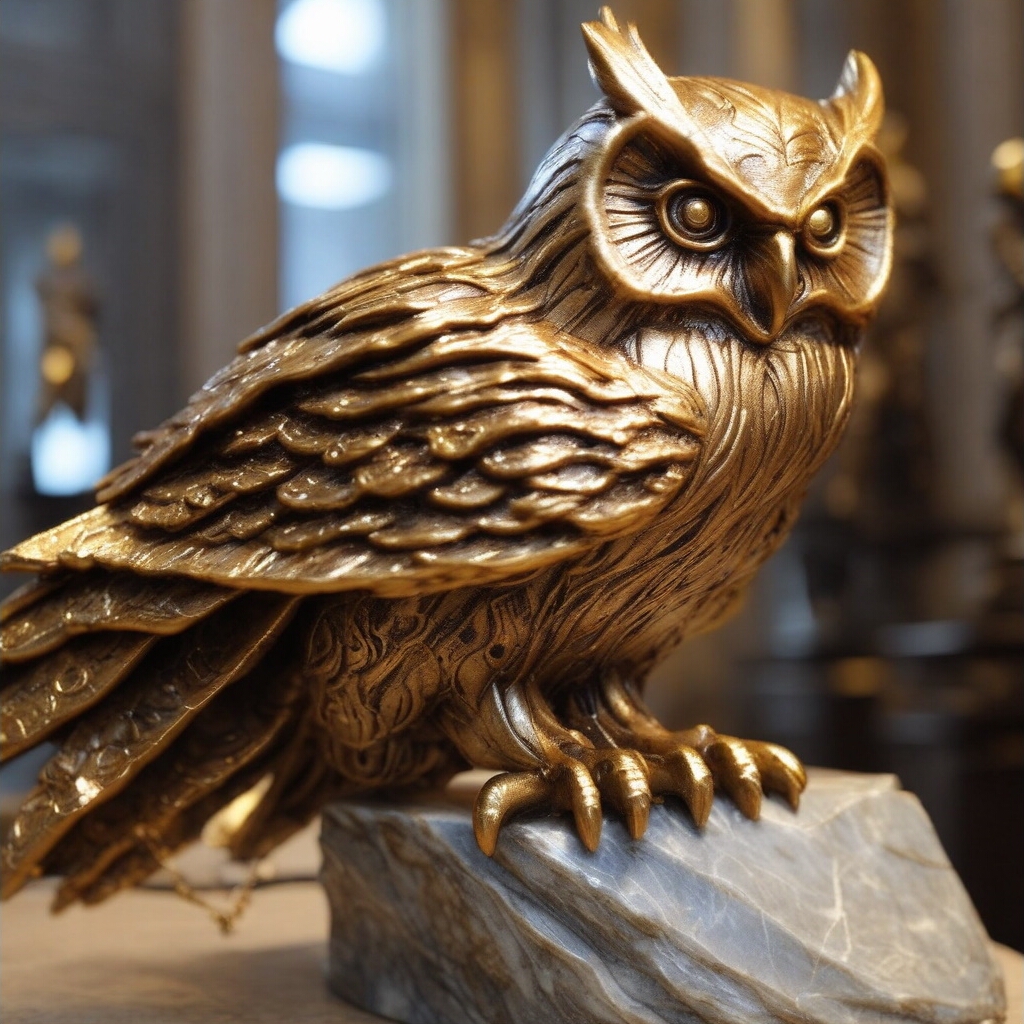}
        \caption{cinematic statue of an owl on a rock, highly detailed, photorealistic}
    \end{subfigure}
    \hfill
    \begin{subfigure}[t]{0.24\textwidth}
        \centering
        \includegraphics[width=.99\textwidth]{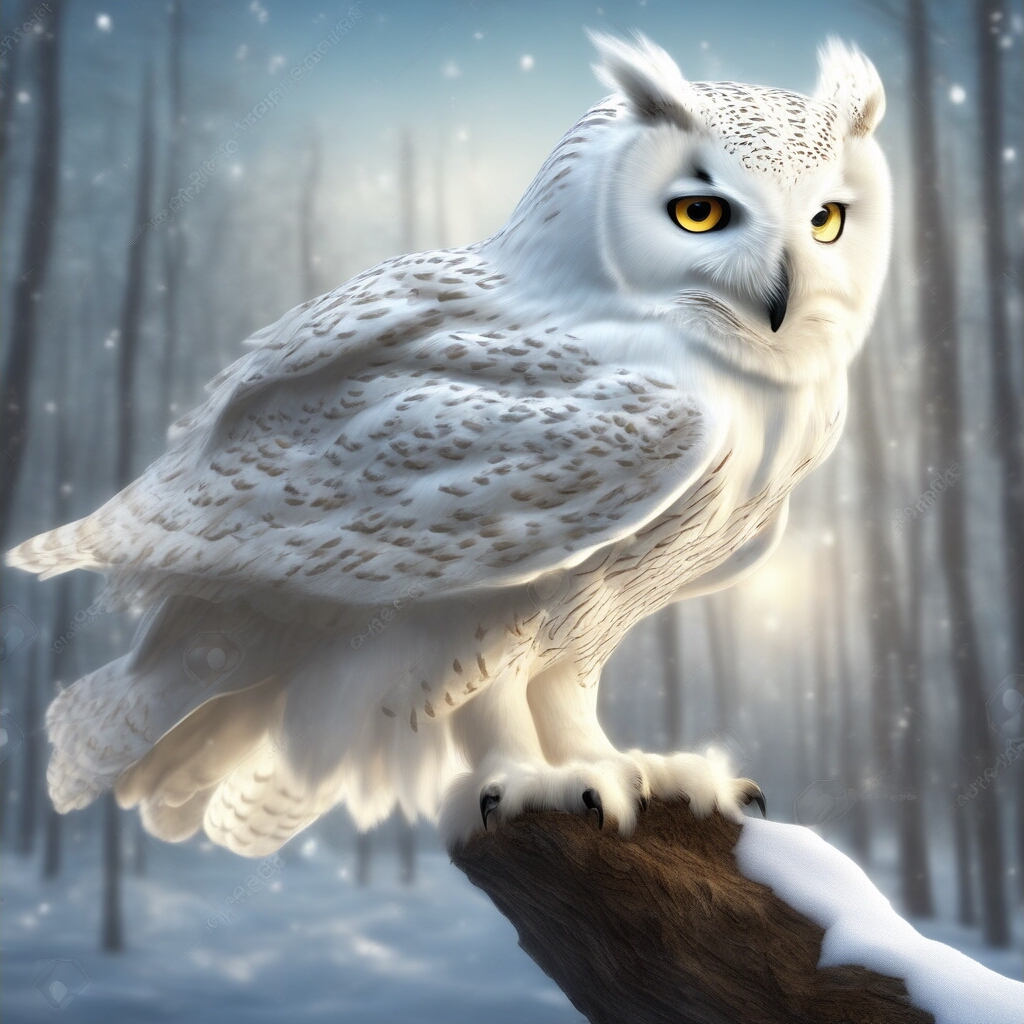}
        \caption{cinematic white snow owl on a rock, highly detailed, photorealistic}
    \end{subfigure}

    \begin{subfigure}[t]{0.24\textwidth}
        \centering
        \includegraphics[width=.99\textwidth]{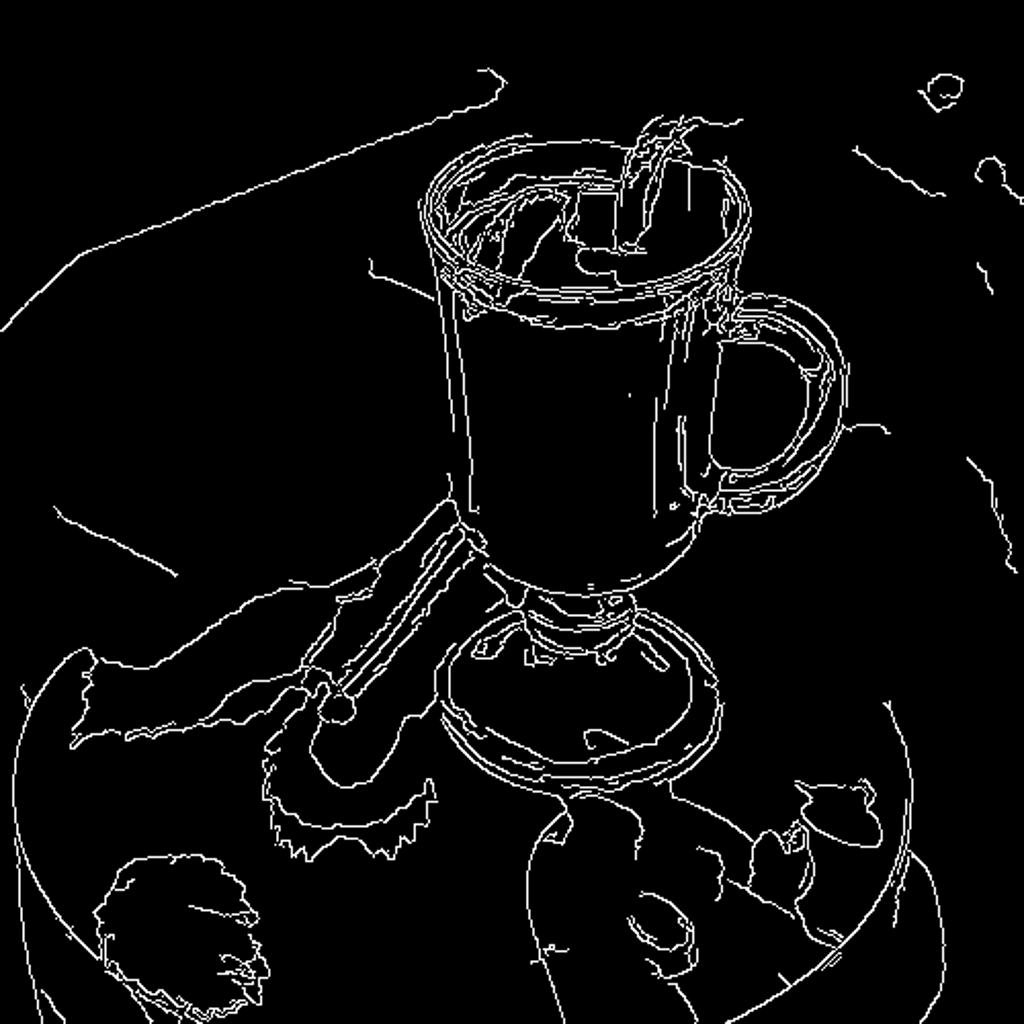}
        \caption{Control}
    \end{subfigure}
    \hfill
    \begin{subfigure}[t]{0.24\textwidth}
        \centering
        \includegraphics[width=.99\textwidth]{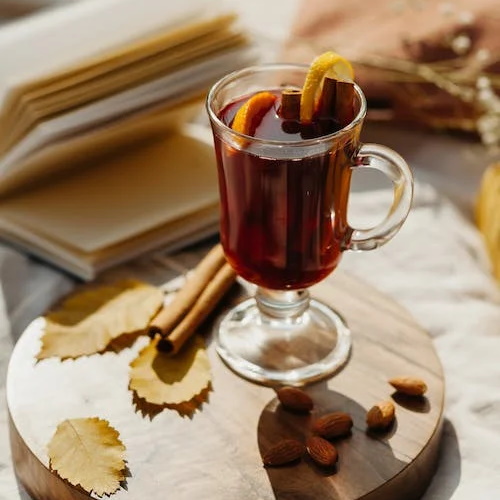}
        \caption{Original Image}
    \end{subfigure}
    \hfill
    \begin{subfigure}[t]{0.24\textwidth}
        \centering
        \includegraphics[width=.99\textwidth]{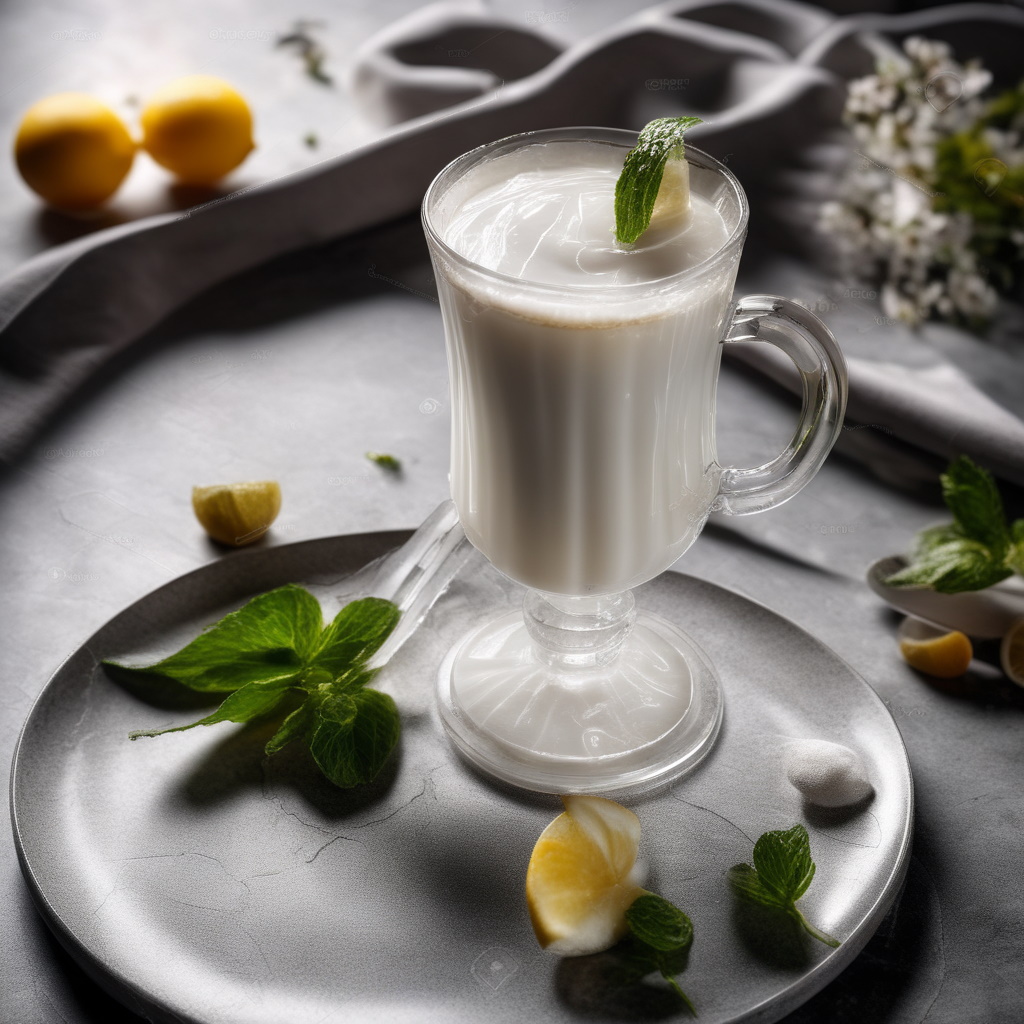}
        \caption{cinematic, highly detailed, milky cocktail on a plate, photorealistic}
    \end{subfigure}
    \hfill
    \begin{subfigure}[t]{0.24\textwidth}
        \centering
        \includegraphics[width=.99\textwidth]{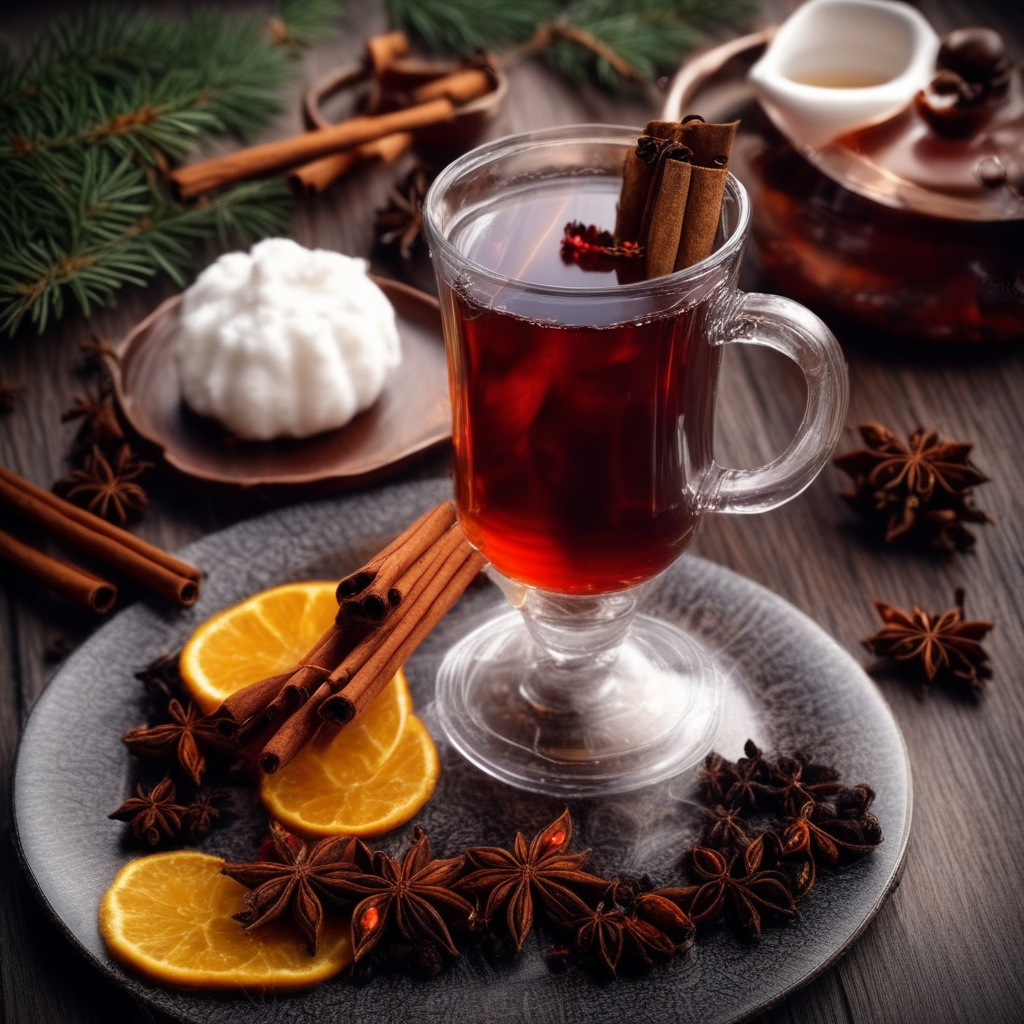}
        \caption{cinematic, highly detailed, winter tea with spices on a plate, photorealistic}
    \end{subfigure}
    \hfill
    \hfill
    \caption{Images generated by ControlNet-XS (20M) and Stable Diffusion XL~\cite{Podell2023_SDXL} as generative model with two different text-prompts. The generated images have the resolution of $1024 \times 1024$.}
    \label{fig:GenerationsXL}
\end{figure*}

We provide additional qualitative results for controlled image generation. We show results for ControlNet-XS applied to Stable Diffusion~\cite{Rombach2022_LDM} with the guidance of Canny-edges, depth maps, semantic maps and human poses in \cref{fig:GenerationsSD1.5,fig:GenerationsSD15_pose_seg}. We also show results with the guidance of Canny-edges and depth maps applied to Stable Diffusion XL~\cite{Podell2023_SDXL} in \cref{fig:GenerationsXL}.

\section{Information about the Prompts for individual Figures}
\label{sec:supp_prompts}
\cref{tab:prompt_info} shows the text-prompts which were used to generate the images provided in the main article and the supplementary material, and are not stated in the respective figures.
If not stated differently, all generated images were sampled with 50 DDIM-steps and with a classifier-free-guidance scale of 9.5.

\begin{table*}
\centering
\caption{Text-prompts that were used to generate the images in the main article as well as in the supplementary material.}
\begin{tabular}{ll}
    \hline
    Figure & Text-Prompt\\
    \hline
    \vspace{+0.1cm}
    Figure 1 (1st) & \makecell[l]{cinematic, beautiful, photo of a guy, street photography, colourful,\\highly detailed, photorealistic}\\
    \vspace{+0.1cm}
    Figure 1 (2nd) & \makecell[l]{cinematic cupcake, blueberry flavoured cupcake, delicious,\\highly detailed, photorealistic}\\
    \vspace{+0.1cm}
    Figure 1 (3rd)& \makecell[l]{Still life of a teddy bear, flowers and an old picture.\\ Everything is on a shelf. High Quality image. Best Quality}\\
    \vspace{+0.1cm}
    Figure 1 (4th) & \makecell[l]{Photo of a woman wearing a fine coat walking through the rain.\\People in the background are blurred out. Colorful lights reflect in\\ the water. depressing mood. High Quality image. Award winning}\\
    \vspace{+0.1cm}
    Figure 4 & aerial image of a city with a big highway intersection\\
    \vspace{+0.1cm}
    Figure 5 & high quality photo of a delicious cake, 4k image\\
    \vspace{+0.1cm}
    \Cref{fig:ControlStrenght} (c-e) & Photo of a big house with stores at the first floor, cars parked, 4k\\
    \vspace{+0.1cm}
    \Cref{fig:ControlStrenght} (h-j) & \makecell[l]{close-up of a young woman, detailed, beautiful, street photography,\\photorealistic, detailed, Kodak ektar 100, natural, candid shot} \\
    \Cref{fig:full_semantic_cn,fig:full_semantic_cnxs-11,fig:full_semantic_cnxs-55,fig:full_semantic_cnxs-491} & \small high quality photo of a delicious cake, 4k image\\
    \hline
\end{tabular}
\label{tab:prompt_info}
\end{table*}

\section{Fidelity of the Control}
\label{sec:supp_control_fidelity}
\begin{figure*}
    \centering

    \begin{subfigure}[t]{0.19\textwidth}
        \centering
        \includegraphics[width=.99\textwidth]{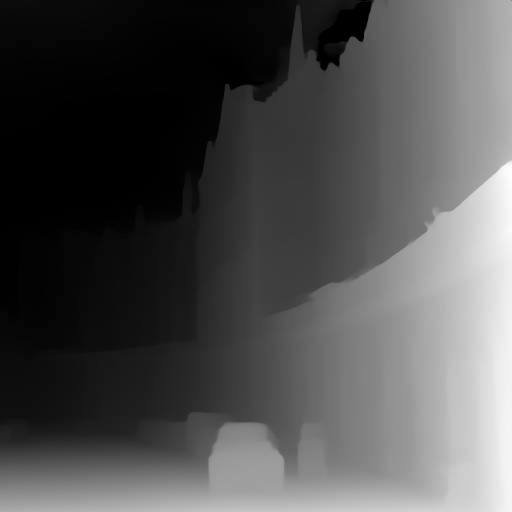}
        \caption{Control}
    \end{subfigure}
    \hfill
    \begin{subfigure}[t]{0.19\textwidth}
        \centering
        \includegraphics[width=.99\textwidth]{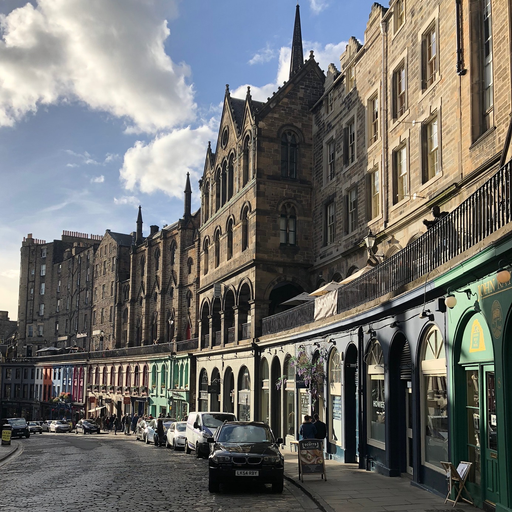}
        \caption{Original image}
    \end{subfigure}
    \hfill
    \begin{subfigure}[t]{0.19\textwidth}
        \centering
        \includegraphics[width=.99\textwidth]{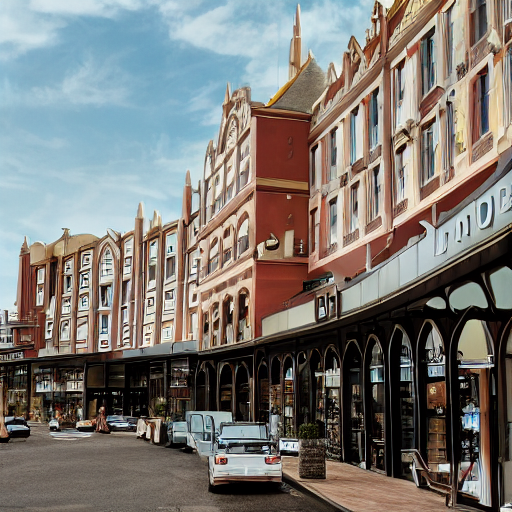}
        \caption{ControlNet-XS 55M}
        \label{subfig:ControlStrenght_street55m}
    \end{subfigure}
    \hfill
    \begin{subfigure}[t]{0.19\textwidth}
        \centering
        \includegraphics[width=.99\textwidth]{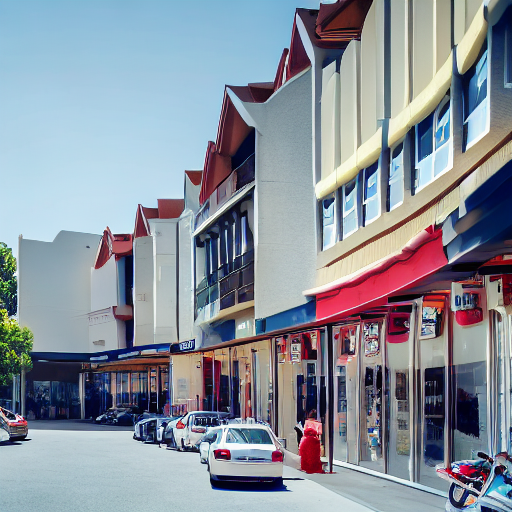}
        \caption{ControlNet-XS 11.7M}
        \label{subfig:ControlStrenght_street11m}
    \end{subfigure}
    \hfill
    \begin{subfigure}[t]{0.19\textwidth}
        \centering
        \includegraphics[width=.99\textwidth]{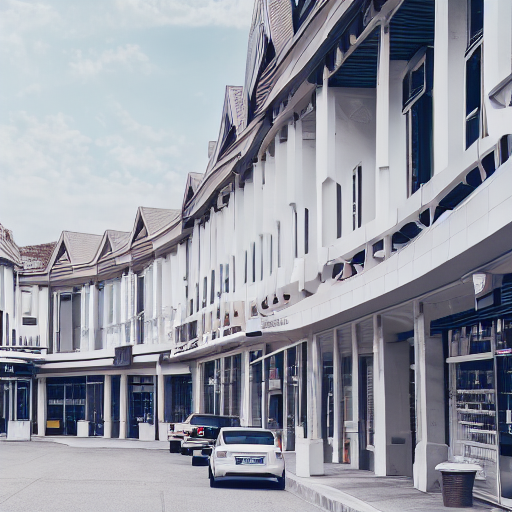}
        \caption{ControlNet-XS 1.7M}
        \label{subfig:ControlStrenght_street1m}
    \end{subfigure}
    \hfill
    \begin{subfigure}[t]{0.19\textwidth}
        \centering
        \includegraphics[width=.99\textwidth]{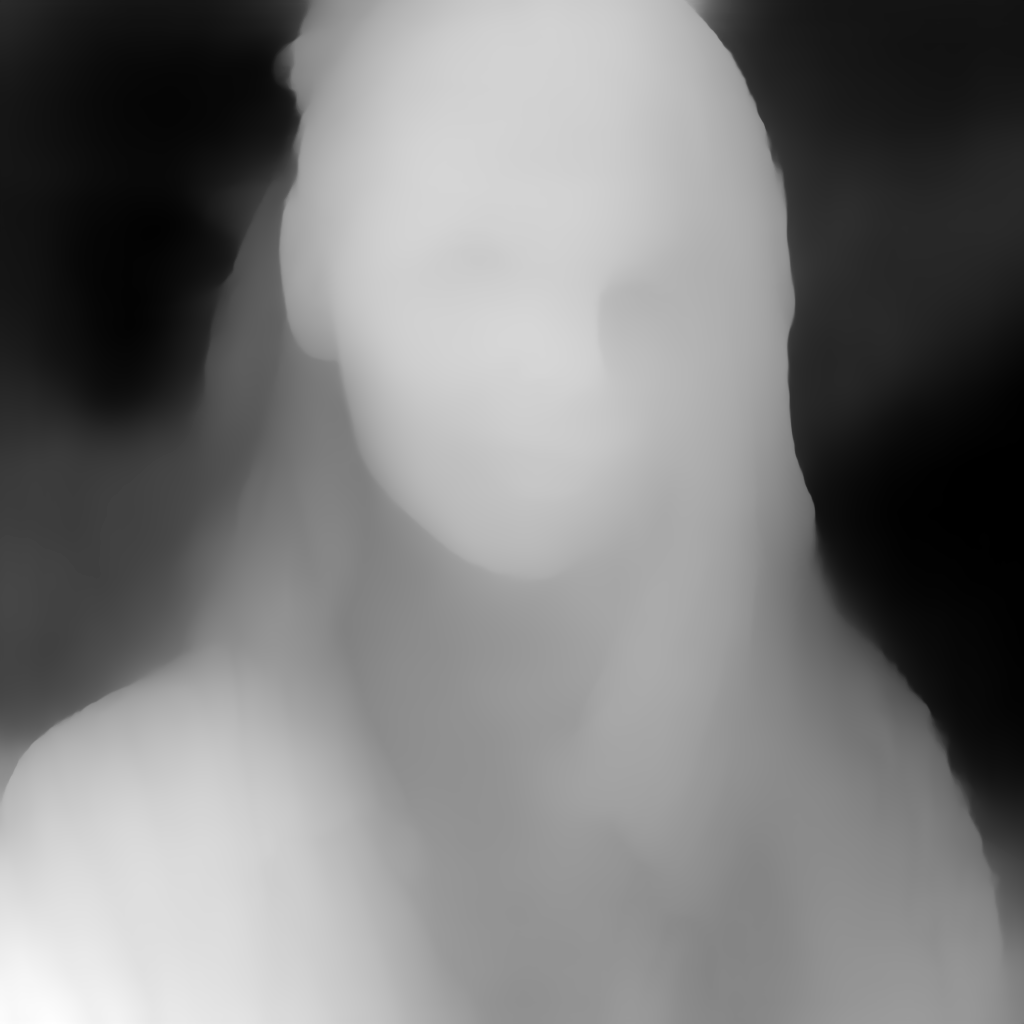}
        \caption{Control}
    \end{subfigure}
    \hfill
    \begin{subfigure}[t]{0.19\textwidth}
        \centering
        \includegraphics[width=.99\textwidth]{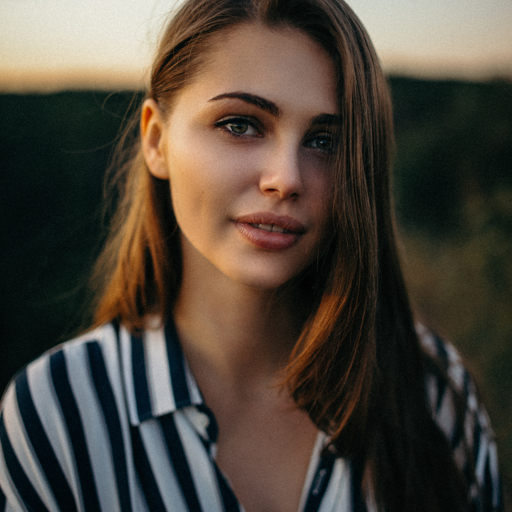}
        \caption{Original Image}
    \end{subfigure}
    \hfill
    \begin{subfigure}[t]{0.19\textwidth}
        \centering
        \includegraphics[width=.99\textwidth]{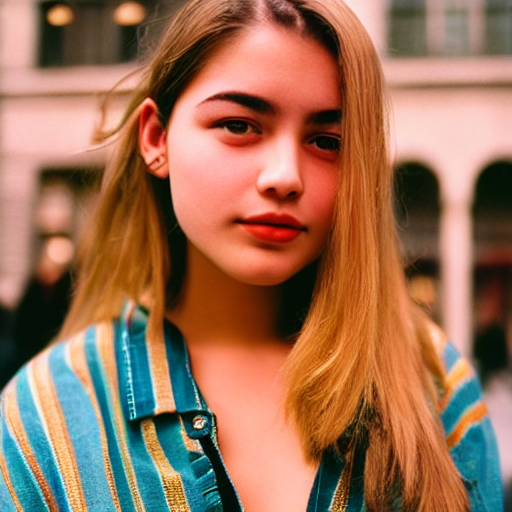}
        \caption{ControlNet-XS 55M}
        \label{subfig:ControlStrenght_face55m}
    \end{subfigure}
    \hfill
    \begin{subfigure}[t]{0.19\textwidth}
        \centering
        \includegraphics[width=.99\textwidth]{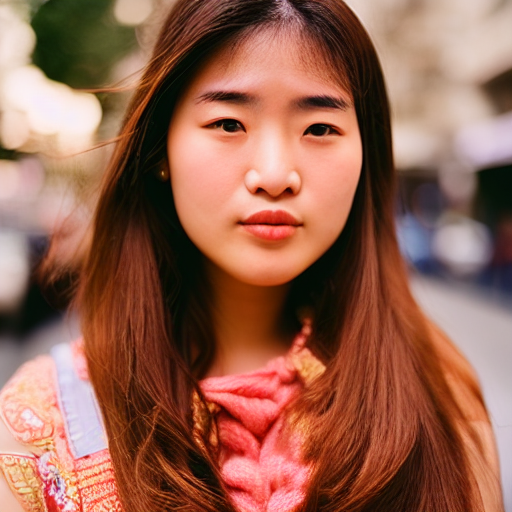}
        \caption{ControlNet-XS 11.7M}
        \label{subfig:ControlStrenght_face11m}
    \end{subfigure}
    \hfill
    \begin{subfigure}[t]{0.19\textwidth}
        \centering
        \includegraphics[width=.99\textwidth]{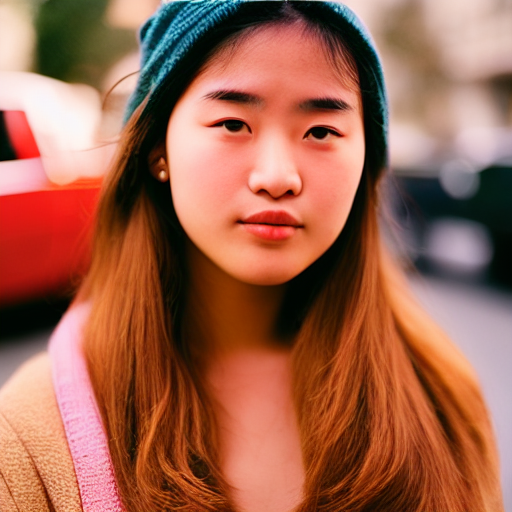}
        \caption{ControlNet-XS 1.7M}
        \label{subfig:ControlStrenght_face1m}
    \end{subfigure}
    \hfill
    \caption{
    {\bf The fidelity of the control} reduces with smaller model sizes of ControlNet-XS for complex, detailed control maps. The generated street scenes (c - e) are with respect to a detailed control depth map (a). We see that in the outputs controlled by the  $55$M parameter model, the complex structures are identical to the original image. Outputs controlled by smaller models with $11.7$M and $1.7$M parameters, respectively, are still guided by the control but less rigorously (\eg roofs of the buildings). Depth maps with fewer details such as the faces in (f) are processed with a similar fidelity of the control by all model sizes (h-j).
    }
    \label{fig:ControlStrenght}
\end{figure*}

Here, we provide additional examples with respect to fidelity of the control, see also main article (Figure 4). In \cref{fig:ControlStrenght}, we qualitatively evaluate the effect of decreased fidelity of the control with smaller ControlNet-XS models. We additionally illustrate the effect which the complexity of control images can have on the fidelity of the control.

\newpage
\section{Semantic Bias of Large Control Models}
\label{sec:supp_semantic_bias}
\begin{figure*}
    \centering
    \includegraphics[width=.99\textwidth]{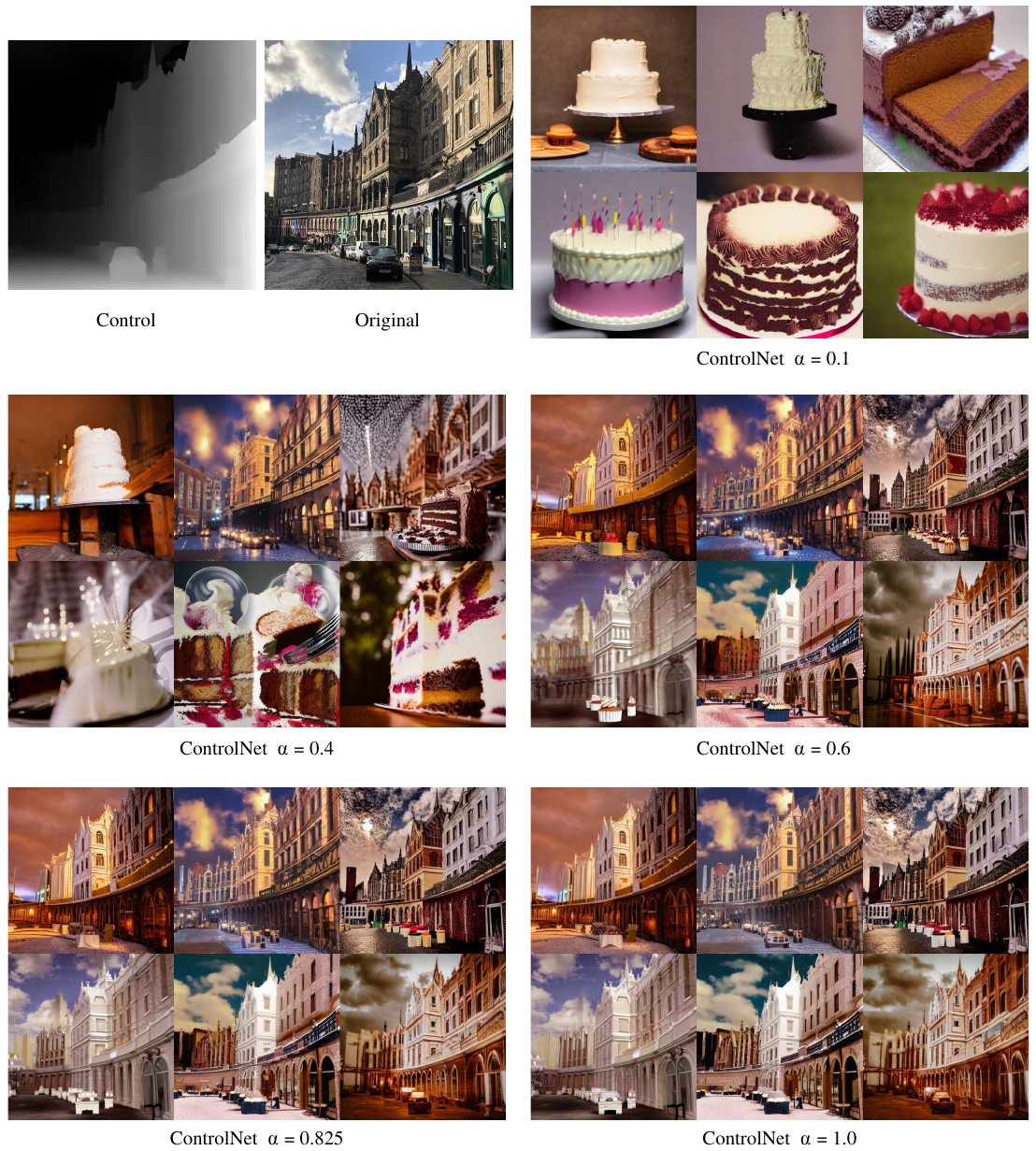}
    \caption{{\bf Semantic bias for depth control.} All images are generated with a control depth map of a street scene and an unrelated text-prompt: ``high quality photo of a delicious cake, 4k image''. Note that these are not contradicting control inputs, but the inputs rather challenge the generative process to produce a creative solution with a cake in form of a street scene. We see that ControlNet~\cite{Zhang2023_ControlNet} is not able to produce satisfying results, 
    even when adjusting the control strength $\alpha$. Note that $\alpha=0.825$ is the default for ControlNet. With this default value, 
    ControlNet shows proper house facade textures, while ControlNet-XS shows typical cake textures such as ``sponge'', ``marzipan'' or ``icing''.   
    (Note that the output signals of the controlling network are added with a global weighting $\alpha$ to the output signals of the generation network at the respective neural blocks. This weighting can be adjusted at test time). Here, $\alpha=0.4$ was the ``sweet spot'' where ControlNet suddenly transitions from producing images of a cake to images of a street scene.}
    \label{fig:full_semantic_cn}
\end{figure*}

\begin{figure*}
    \centering
    \includegraphics[width=.99\textwidth]{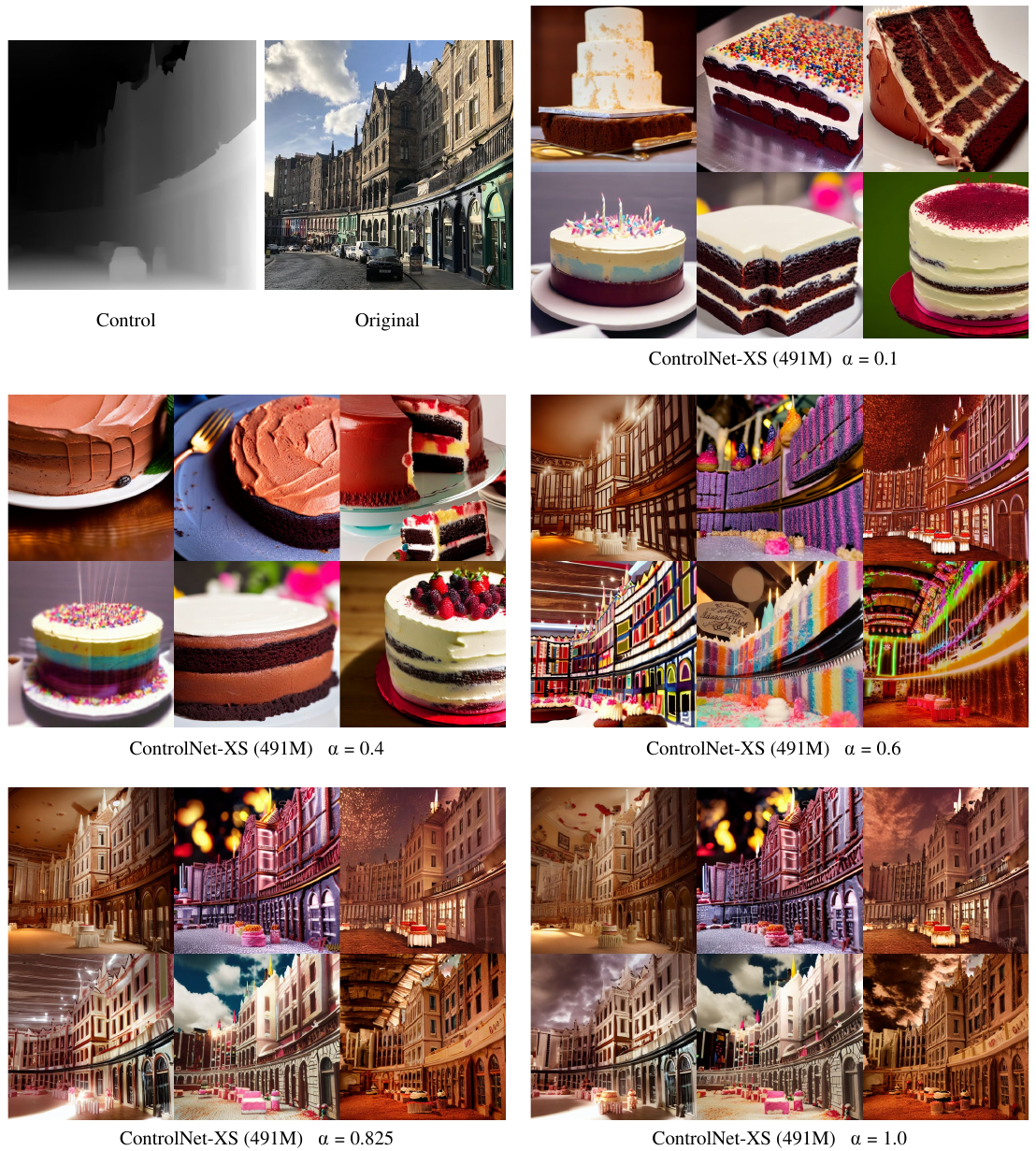}
    \caption{{\bf Semantic bias for depth control with ControlNet-XS (491M)}. Please find a detailed explanation in \cref{fig:full_semantic_cn}. We see that for $\alpha=0.6$ the control is picked up. For $\alpha = 0.6$ there are examples with cake like textures, but in the layout of the guidance image. However, as expected, for $\alpha \in [0.825, 1]$ the semantic bias becomes more evident. The overall image is still a street scene, while there are only a few cake-like textures.}
    \label{fig:full_semantic_cnxs-491}
\end{figure*}

\begin{figure*}
    \centering
    \includegraphics[width=.99\textwidth]{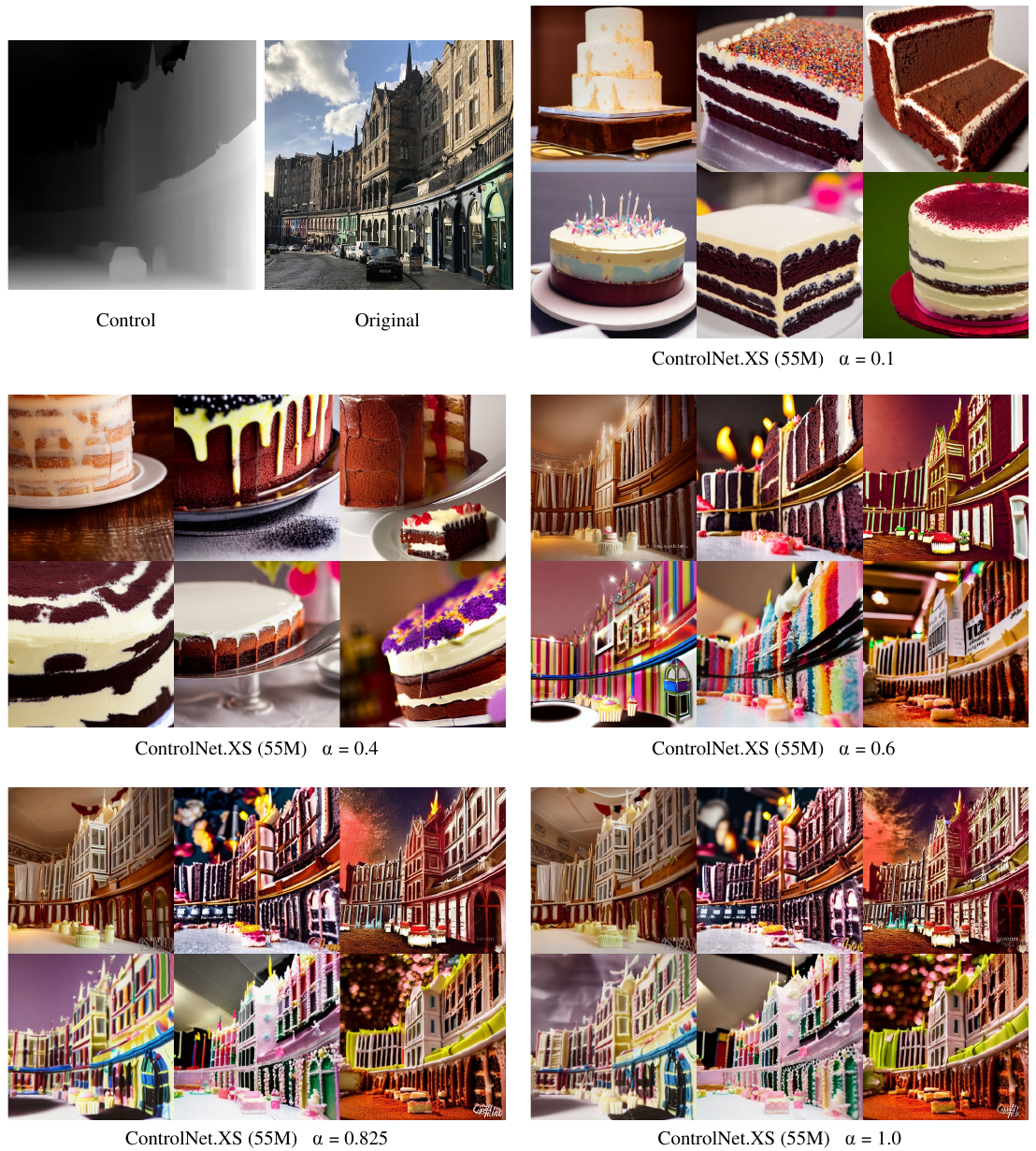}
    \caption{{\bf Semantic bias for depth control with ControlNet-XS (55M)}. Please find a detailed explanation in \cref{fig:full_semantic_cn}. We see that for $\alpha=0.6$ the control is picked up. For all $\alpha\in[0.6, 1]$ a street scene with a cake texture is shown. The cake texture is more detailed than with ControlNet-XS (11.7M) (see \cref{fig:full_semantic_cnxs-11}).}
    \label{fig:full_semantic_cnxs-55}
\end{figure*}

\begin{figure*}
    \centering
    \includegraphics[width=.99\textwidth]{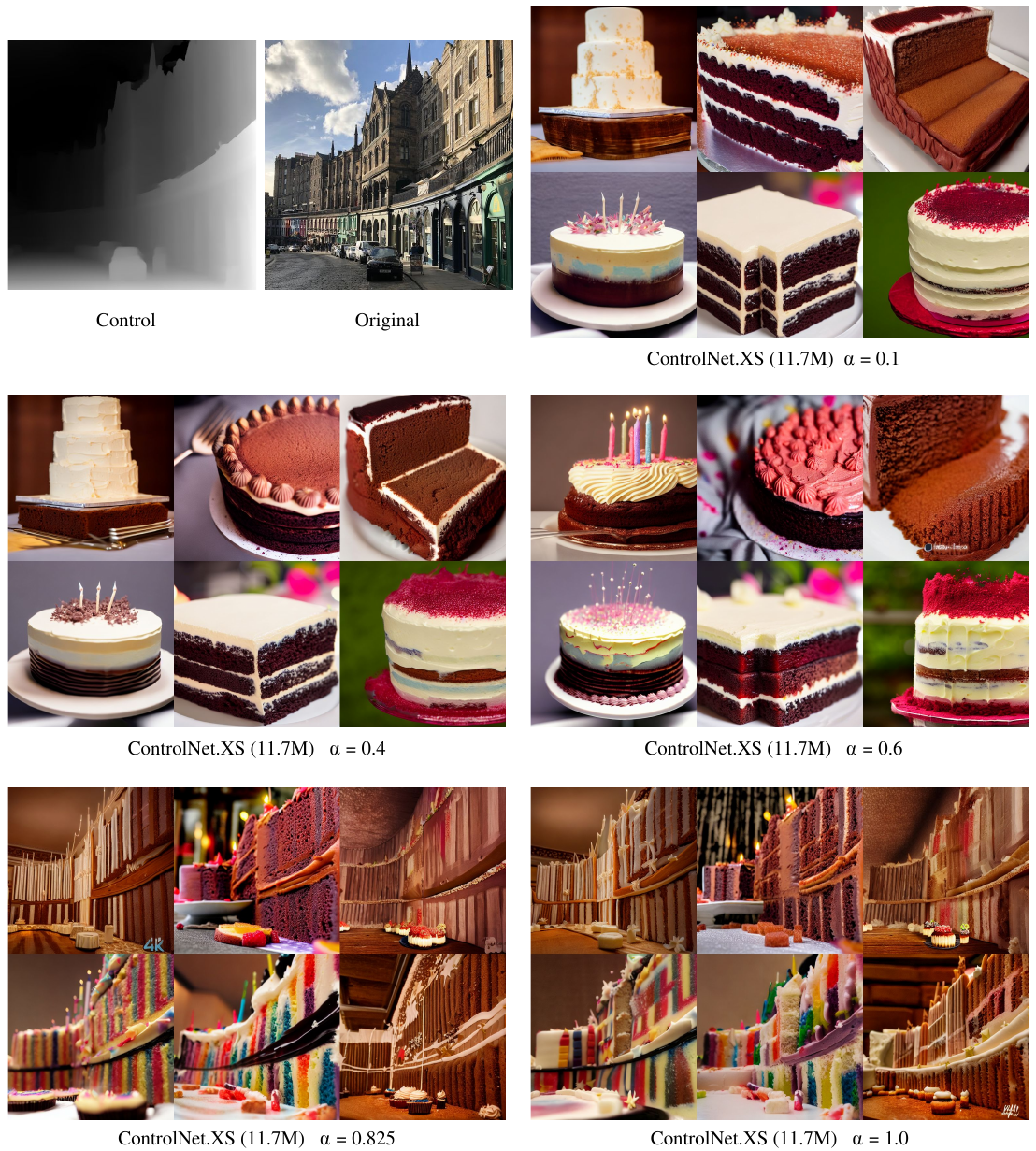}
    \caption{{\bf Semantic bias for depth control with ControlNet-XS (11.7M)}. Please find a detailed explanation in \cref{fig:full_semantic_cn}. We see that for $\alpha=0.825$ the control is picked up. For both $\alpha=0.825$ and $\alpha=1$, a street since with a cake texture is shown. The cake texture is less detailed then with ControlNet-XS (55M) (see \cref{fig:full_semantic_cnxs-55}). This can be expected since we have seen above (e.g. \cref{fig:ControlStrenght}) that 
    smaller models have reduced fidelity of the control. 
    }
    \label{fig:full_semantic_cnxs-11}
\end{figure*}

In the following, we provide additional results for Figure 5 from the main article regarding the semantic bias of large control models. We show results for different control strengths $\alpha$ for ControlNet~(\cref{fig:full_semantic_cn}), ControlNet-XS with 491M parameters (\cref{fig:full_semantic_cnxs-491}), ControlNet-XS with 55M parameters (\cref{fig:full_semantic_cnxs-55}) and ControlNet-XS with 11.7M parameters (\cref{fig:full_semantic_cnxs-11}).
Overall, we notice that smaller ControlNet-XS networks (55M, 11.7M) have less semantic bias than larger models like ControlNet (361M) and the over-sized ControlNet-XS (491M). We also observe that the semantic bias cannot be removed by adapting the control-strength $\alpha$ for larger control networks.
We conjecture that the reason for the semantic bias, induced by large control models, is that larger models can use redundant parameters to add semantic meaning to the control.
With controlling networks that utilise an improved communication between generator and controller, like ControlNet-XS, the semantic bias can be slightly reduced by optimising $\alpha$ even in over-sized controlling networks. Hence we start to see in \cref{fig:full_semantic_cnxs-491} results which roughly show a cake texture for $\alpha=0.6$.

\end{document}